\documentclass[journal]{IEEEtran}
\usepackage{times}
\usepackage{epsfig}
\usepackage{graphicx}
\usepackage{threeparttable}
\usepackage{multirow}
\usepackage{booktabs}
\usepackage{subfigure}
\usepackage{cite}
\usepackage{graphicx}
\usepackage{float}
\usepackage{amsmath,amssymb}
\usepackage{color}
\usepackage{enumerate}
\usepackage{algorithm}
\usepackage{algorithmic}
\usepackage{stfloats}
\usepackage{paralist}
\usepackage[pagebackref=true,breaklinks=true,letterpaper=true,colorlinks,bookmarks=false]{hyperref}

\def\Fix#1{{\color{black}{}{#1}}}
\def\Fixvv#1{{\color{black}{}{#1}}}

\newcommand\ie{\emph{i.e. }}
\newcommand\ien{\emph{i.e.}}

\newcommand\cf{\emph{c.f. }}

\newcommand\wrt{w.r.t. }

\newcommand\etal{\emph{et al.}}

\begin{document}
\title{Deep Idempotent Network for Efficient Single Image Blind Deblurring}

\author{Yuxin~Mao,
        Zhexiong~Wan,
        Yuchao~Dai,~\IEEEmembership{Member,~IEEE}
        and~Xin~Yu,~\IEEEmembership{Member,~IEEE}
\thanks{Yuxin Mao, Zhexiong Wan, and Yuchao Dai are with School of Electronics and Information, Northwestern Polytechnical University, Xi'an, Shaanxi, 710129, China. Yuxin Mao and Zhexiong Wan contributed equally. Yuchao Dai (daiyuchao@gmail.com) is the corresponding author.}
\thanks{Xin Yu is with Faculty of Engineering and Information Technology, the University of Technology Sydney, Sydney, NSW 2007, Australia.}
\thanks{Digital version are available at https://doi.org/10.1109/TCSVT.2022.3202361}
}

\markboth{IEEE Transactions on Circuits and Systems for Video Technology,~Vol.~xx, No.~xx, xx~xx}
{Shell \MakeLowercase{\textit{et al.}}: Bare Demo of IEEEtran.cls for IEEE Journals}
\maketitle

\begin{abstract}
  Single image blind deblurring is highly ill-posed as neither the latent sharp image nor the blur kernel is known. 
  Even though considerable progress has been made, several major difficulties remain for blind deblurring, including the trade-off between high-performance deblurring and real-time processing.
  Besides, we observe that current single image blind deblurring networks cannot further improve or stabilize the performance but significantly degrades the performance when re-deblurring is repeatedly applied. This implies the limitation of these networks in modeling an ideal deblurring process.
   In this work, we make two contributions to tackle the above difficulties: (1) We introduce the idempotent constraint into the deblurring framework and present a deep idempotent network to achieve improved blind non-uniform deblurring performance with stable re-deblurring. 
  (2) We propose a simple yet efficient deblurring network with lightweight encoder-decoder units and a recurrent structure that can deblur images in a progressive residual fashion.
   Extensive experiments on synthetic and realistic datasets prove the superiority of our proposed framework. Remarkably, our proposed network is nearly 6.5$\times$ smaller and 6.4$\times$ faster than the state-of-the-art while achieving comparable high performance.
\end{abstract}


\IEEEpeerreviewmaketitle

\section{Introduction}
\begin{figure}[tbp]
\centering
\includegraphics[width=0.47\textwidth]{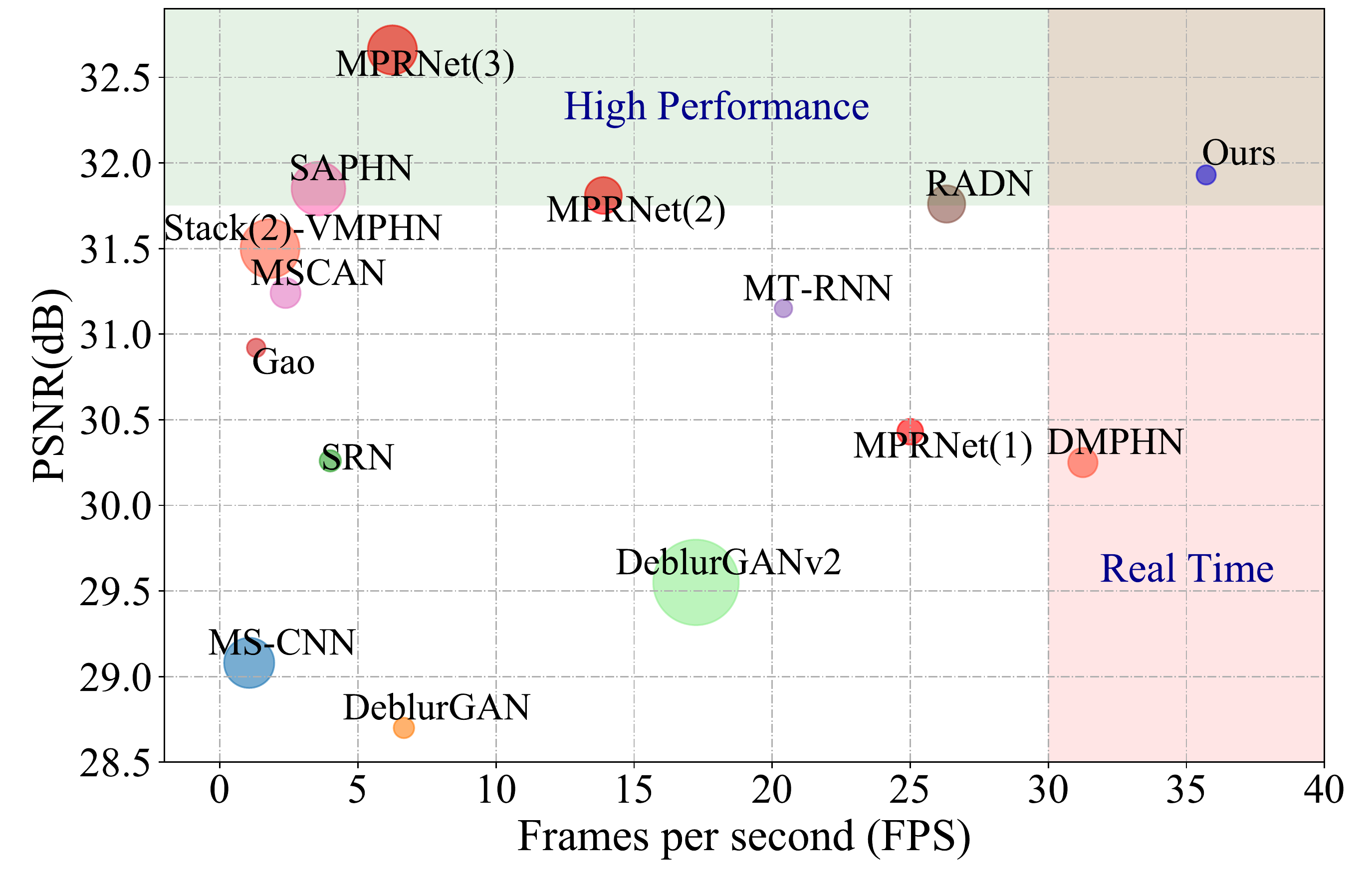}
\caption{\textbf{Speed vs Performance.} Each circle represents the performance of a model in terms of FPS and PSNR on the GoPro~\cite{nah_DeepDeblur_CVPR_2017} dataset with 1280 $\times$ 720 images using an RTX 2080Ti GPU. The radius of each circle denotes the model’s number of parameters. Our method achieves high performance with real-time runtime and small parameters compared with state-of-the-art blind deblurring methods including MS-CNN~\cite{nah_DeepDeblur_CVPR_2017}, SRN~\cite{tao_SRN_CVPR_2018}, DMPHN~\cite{zhang_DMPHN_CVPR_2019}, DeblurGAN~\cite{kupyn_DeblurGAN_CVPR_2018}, DeblurGANv2~\cite{ kupyn_DeblurGANv2_ICCV_2019}, Gao\etal~\cite{gao_NestedSkip_CVPR_2019}, MT-RNN~\cite{park_MTRNN_ECCV_2020}, SAPHN~\cite{suin_AttentionDMPHN_CVPR_2020}, RADN~\cite{purohit_RADN_AAAI_2020}, MSCAN~\cite{wan_channel_attention_TCSVT_2021} and MPRNet~\cite{waqas_MPRNet_CVPR_2021}.}
\label{fig:scatter}
\end{figure}


\IEEEPARstart{S}{ingle} image blind deblurring aims at estimating a sharp image from a blurry input. It is a challenging ill-posed problem as both the sharp image and the kernel need to be estimated~\cite{michaeli_PatchRecurrence_ECCV_2014, pan_DarkChannel_CVPR_2016, pan_PhaseOnlyDeblur_CVPR_2019, kaufman_DeblurAnalysisSynthesis_CVPR_2020}. 
This problem becomes even more challenging for real-world complex blurry images when different exposure times, camera motion, and multiple moving objects exist. 
To tackle these difficulties, traditional optimization-based approaches estimate the sharp image and the kernel alternatively, where various priors have been exploited to regularize this procedure~\cite{xu_L0Prior_CVPR_2013, michaeli_PatchRecurrence_ECCV_2014, pan_DarkChannel_CVPR_2016, yan_ExtremePrior_CVPR_2017, pan_PhaseOnlyDeblur_CVPR_2019}. 
Recently, remarkable progress has been made by designing different network architectures to learn the mapping from a blurred image to its corresponding sharp version in an end-to-end manner without estimating the blur kernels~\cite{nah_DeepDeblur_CVPR_2017, kupyn_DeblurGAN_CVPR_2018, zhang_DMPHN_CVPR_2019, suin_AttentionDMPHN_CVPR_2020}. 
These approaches have achieved profound success on benchmark datasets such as GoPro~\cite{nah_DeepDeblur_CVPR_2017}. 
Despite the advance, a very recent study~\cite{rim_RealBlurDataset_ECCV_2020} reports that state-of-the-art models trained on the synthetic dataset do not generalize well to real-world blurry images. Therefore, it is still challenging for existing methods to address complex non-uniform motion blurs in dynamic scenes.

\begin{figure}[tbp]
\centering
\includegraphics[width=0.45\textwidth]{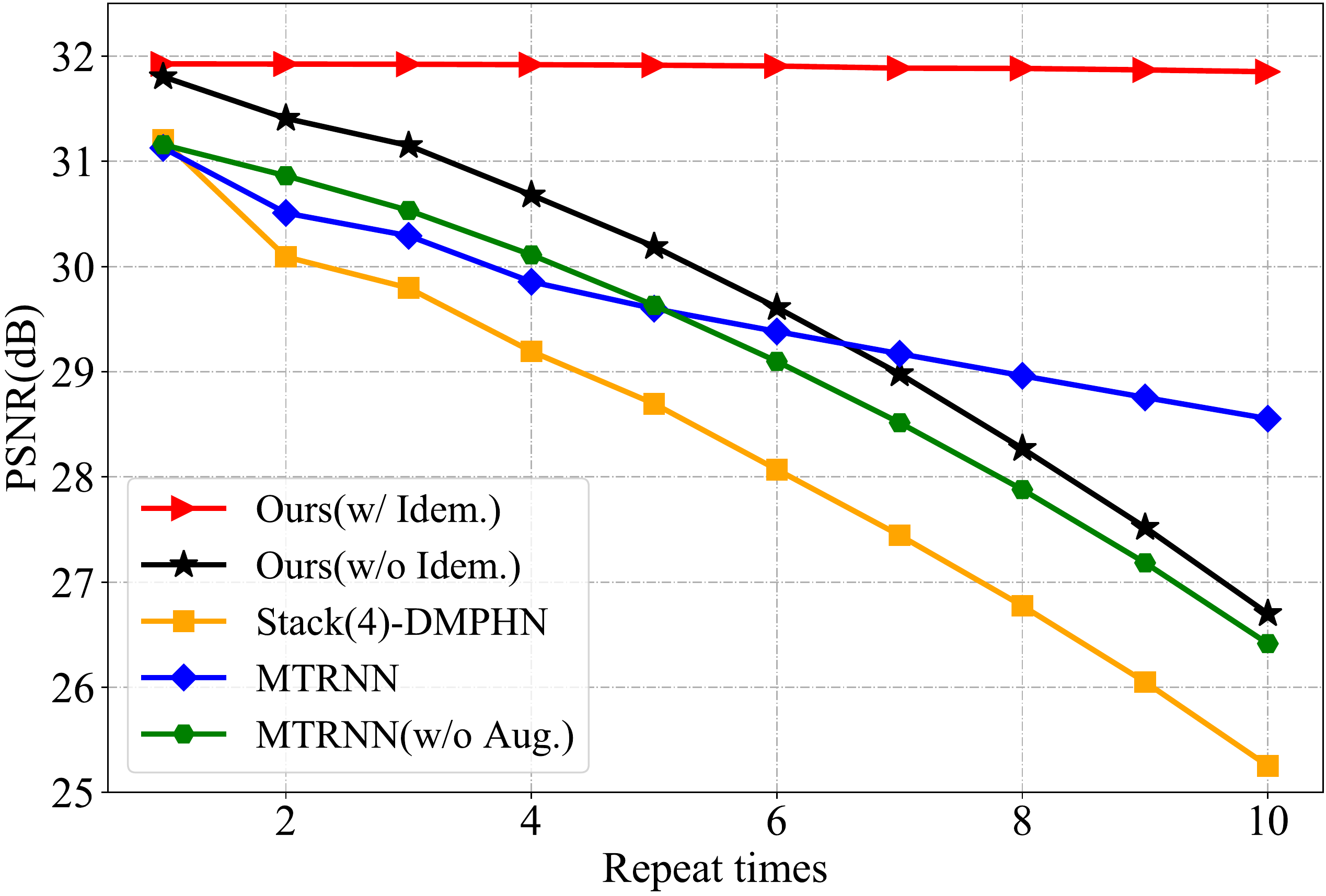}
\caption{\textbf{The performance curve of repeatedly re-deblurring.} We repeatedly input the deblurred image to the network by multiple times and report the deblurring results on the GoPro dataset. Our proposed deep idempotent network achieves very stable deblurring results while the performance of all other state-of-the-art methods decreases as the repeating times increase. 
Note that, to keep the training settings consistent with our results without idempotent constraint, we re-trained MT-RNN~\cite{park_MTRNN_ECCV_2020} without their multi-temporal data augmentation.} 
\label{fig:idem_blur}
\end{figure}

\Fix{In practical applications, it is intuitive to determine whether any obtained image is blurred or not, but it is difficult to assert whether the image has been deblurred. Therefore, for an image deblurring system, it is inevitable to repeatedly input an image that has been deblurred. We expect the performance could be improved or maintained when a deblurred image is repeatedly input to an image deblurring model.}
To verify this hypothesis, we repeatedly implement the state-of-the-art methods with their pre-trained models~\cite{zhang_DMPHN_CVPR_2019, park_MTRNN_ECCV_2020} and report their re-deblurring performance in terms of PSNR in Fig.~\ref{fig:idem_blur}. Surprisingly, we observe a significant performance decrease after re-deblurring. This demonstrates that existing deblurring models cannot further improve their performance by repetitively applying themselves to blurry inputs. 



\Fix{Although we do not expect that re-deblur a deblurred image once again would lead to a sharper image, at least it should not significantly degrade the previously deblurred result.}
To remedy this issue, we resort to the concept of idempotence in mathematics, \ien, an operator can be applied many times and still maintain the primary results.
An ideal deblurring model should also own this property, because there is a theoretical upper limit (\ien, output the completely sharp image) in deblurring.
Therefore, the ideal situation is that the results are consistent when we implement the algorithm repeatedly, and we call it \textit{Idempotent Property}.
To achieve this goal, we introduce an idempotent constraint into the network design and propose our \emph{deep idempotent network} for single image blind deblurring. \Fix{The idempotent constraint aims to maintain the consistency between the deblurred image and the re-deblurred image.}
With the introduced constraint, our network outputs stable deblurred results even after deblurring multiple times, as shown in Fig.~\ref{fig:idem_blur}.

Moreover, many state-of-the-art blind deblurring methods have large model sizes and take long inference time, as shown in Fig.~\ref{fig:scatter}.
To satisfy the requirements of real-time (at least 30 FPS) applications, we design a novel, efficient, and lightweight single image blind deblurring network. Our idempotent network is composed of a lightweight encoder-decode module within a progressively recurrent iteration structure.
Here, we can control the number of recurrent structure to balance the deblurring efficiency and performance. 
The whole network is trained by applying our idempotent constraint to the outputs of the deblurring and re-deblurring processes.
Once our network is trained, it only runs in a single feed-forward fashion without re-deblurring in testing.

Our proposed simple yet efficient idempotent network achieves state-of-the-art deblurring performance on the GoPro dataset with $1280 \times 720$ images and runs in real-time.
Our idempotent constraint
is quite generic, and we further demonstrate its superiority on the tasks of image dehazing and deraining.
Moreover, we apply the idempotent constraint to the open-source code of the existing state-of-the-art models, and further improve their results.


Our main contributions are summarized as follows:
\begin{enumerate}[1)]
    \item We introduce an idempotent property to single image blind deblurring, and propose an idempotent constraint to improve non-uniform deblurring performance.
    \item We design a simple yet efficient deblurring network that achieves real-time and high performance by progressive residual deblurring with recurrent structure.
    \item Our model, while achieving comparable high performance on the GoPro benchmark, is nearly 6.5$\times$ smaller and 6.4$\times$ faster than the existing state-of-the-art approach, \ien, MPRNet~\cite{waqas_MPRNet_CVPR_2021}.
    \item Our proposed model achieves superior generalization performance on the real captured RealBlur benchmark. The proposed idempotent network architecture and constraint can be easily generalized to dehazing and deraining and improve their performance.
\end{enumerate}

\section{Related Work}
In this section, we briefly review both optimization-based and learning-based blind image deblurring methods, and idempotence in deep learning.

\noindent\textbf{Optimization based image deblurring.} 
Existing methods based on optimization mainly focus on exploiting different image priors to recover sharp images from blurry images. These valid priors can be enumerated as sparse gradients~\cite{fergus_SparseGradients_SIGGRAPH_2006}, $l_0$ norm prior~\cite{xu_L0Prior_CVPR_2013}, patch recurrence prior~\cite{michaeli_PatchRecurrence_ECCV_2014}, dark channel prior~\cite{pan_DarkChannel_CVPR_2016}, bright channel prior~\cite{yan_ExtremePrior_CVPR_2017}, latent structure prior~\cite{bai_ms_structure_prior_TCSVT_2020}, minimal pixels prior~\cite{wen_local_intensity_prior_TCSVT_2021} and super-pixel prior~\cite{luo_superpixel_seg_prior_TCSVT_2021}. 
Benefiting from the hand-crafted priors, optimization-based algorithms achieve competitive deblurring results for blurry images. However, many priors are only designed for specific blurry scenes and cannot generalize to cross-domain images. 
\Fix{Besides, Srinivasan \etal~\cite{srinivasan_lightmotiondeblur_cvpr_2017} introduce a general model for light field camera motion estimation and image deblurring, and Mohan \etal~\cite{mohan_dividedeblur_cvpr_2018} decompose this model to achieve full-resolution motion deblurring.} 
Meanwhile, optimization-based methods are often time-consuming and need a complex parameter-tuning strategy for different datasets, which restricts their real-world applications.

\noindent\textbf{Deep image deblurring.} Image deblurring greatly benefits from the progress of deep learning. Deep neural networks learn the nonlinear mapping between the blurred and sharp image pairs to deal with complex motion blur. Previous deblurring methods apply convolutional neural networks (CNN) in the process of non-blind deblurring~\cite{xu_DeepDconv_NIPS_2014, zhang_iterative_non_blind_CVPR_2017}.

Recently, researchers shift their attention to blind deblurring~\cite{sun_CVPR_2015, nah_DeepDeblur_CVPR_2017, gao_NestedSkip_CVPR_2019, tao_SRN_CVPR_2018, zhang_DMPHN_CVPR_2019, ren_DeepPriorDeblur_CVPR_2020, suin_AttentionDMPHN_CVPR_2020, park_MTRNN_ECCV_2020, shen_HumanDeblur_CVPR_2019, cai_DBCPeNet_TIP_2020, yuan_OptFlowDeconv_CVPR_2020}. 
Following a coarse-to-fine scheme, MS-CNN~\cite{nah_DeepDeblur_CVPR_2017} and SRN-Deblur~\cite{tao_SRN_CVPR_2018} introduce multi-scale deep networks to restore sharp images in an end-to-end manner. 
DeblurGAN~\cite{kupyn_DeblurGAN_CVPR_2018, kupyn_DeblurGANv2_ICCV_2019} and DBGAN~\cite{zhang_RealisticBlur_CVPR_2020} regard image deblurring as an image generation problem and use a Generative Adversarial Network (GAN)~\cite{goodfellow_GAN_NIPS_2014} for deblurring. 
Gao \etal~\cite{gao_NestedSkip_CVPR_2019} add nested skip connections and a multi-scale parameter selective sharing strategy on a network. 
\Fix{Lumentut \etal~\cite{lumentut_deeprecurrentlightdeblur_spl_2019} propose a recurrent network for full-resolution light field image deblurring.}
Shen \etal~\cite{shen_HumanDeblur_CVPR_2019} propose a human-aware deblurring approach to remove the blur of foreground humans.
DMPHN~\cite{zhang_DMPHN_CVPR_2019} proposes a multi-patch hierarchical network to better deal with spatially non-uniform motion blur.
SAPHN~\cite{suin_AttentionDMPHN_CVPR_2020} combines the multi-patch hierarchical structure with global attention and adaptive local filters to learn the transformation of features in the deblurring process.
MSCAN~\cite{wan_channel_attention_TCSVT_2021} proposes a channel-attention convolutional neural network for single image denblurring.
Wang \etal~\cite{zhang_recursive_video_deblur_TCSVT_2021} propose a recursive video deblurring network, and MACNN~\cite{wang_multi_attention_video_deblur_TCSVT_2021} introduces the multi-attention mechanism to video deblurring.
MPRNet~\cite{waqas_MPRNet_CVPR_2021} proposes an effective multi-stage architecture with a cross-stage feature fusion module and supervised attention module, that progressively learns restoration functions for the degraded inputs. 
MT-RNN~\cite{park_MTRNN_ECCV_2020} designs a shared weight neural network with recurrent feature maps and proposed an incremental temporal training strategy with additional temporal data augmentation. In the training process, they used synthetic images with different levels of blurriness as their supervision, thus their network learned the ability to progressively deblur. 
On the contrary, as shown in Table~\ref{tab:GoPro_benchmark} and Fig.~\ref{fig:fig_progress}, our proposed deep idempotent network achieves better deblurring performance and still owns the ability of progressive deblurring without using such temporal data augmentation strategy.


\noindent\textbf{Idempotence in deep learning.}
There are few works in exploring idempotence in deep learning. 
Zhao \etal~\cite{zhao_MergeAndRun_IJCAI_2018} propose a Merge-and-Run mapping with parallel residual branches to keep the information flow linearly idempotent, which assembles the residual branches in parallel. 
Xing \etal~\cite{Xing_CouplingRGBDSeg_ICIP_2019} extend the Merge-and-Run block into a semantic RGB-D segmentation task to effectively fuse two modality inputs.
Different from these works, we enforce our deblurring network to achieve this idempotent property, thus keeping the deblurring and re-deblurring results consistent.
They try to enforce the idempotency between the convolutional layers but do not guarantee the idempotent property for the whole network outputs, which do not achieve our goal of idempotent deblurring.
Another perspective to understand the idempotent property of the model is from the fixed point theory. 
It can be considered as finding a fixed point in the network output space that can ensure the stability of the deblurring performance. 
To achieve this goal, the deep equilibrium model~\cite{Bai_DEQ_NIPS_2019} solves the fixed point directly by a few convolutional layers.
Instead of solving the fixed point analytically, we introduce a novel idempotent constraint to approximate realization of the fixed point constraints and achieve stable re-deblurring.

\section{Deep Idempotent Deblurring Framework}
In this section, we propose a deep idempotent deblurring framework that embeds the idempotent property into a deep network for efficient single image blind deblurring.
To begin with, we first explain why idempotency is needed for an image deblurring model. Then, we present the definition of the idempotent network and introduce our idempotent network architecture in detail.
Finally, we describe the proposed idempotent constraint as the training loss function.

\subsection{Why idempotent property is needed?}

Images captured in the real world may have different blurring levels from the extreme case of completely sharp (without any blur) to seriously blurred.
The image deblurring algorithms deployed in a real-world system should be able to handle these situations.
A straightforward idea is to design a classifier to select different deblurring models based on the blurring levels.
However, this classifier requires additional computational effort and will lead to worse deblurring performance when the classification results are incorrect.
Therefore, we would prefer an end-to-end model that can better handle the complex non-uniform motion blur in practice.

\Fix{In addition, the output of an ideal deblurring model should be a sharp image. If we apply the same model again on the deblurred image, it should stably output the same sharp image. We believe that an ideal deblurring model should have the idempotent property. 
From another perspective, the re-deblurring process can be regarded as deblurring an image with a blur level of zero. Previous work \cite{park_MTRNN_ECCV_2020} has shown that a fully convolutional network can deal with different levels of motion blur at different spatial locations. 
Based on this, we use a uniform regularization term for each pixel with non-uniform blurring levels by introducing the idempotent constraint.
We believe that the idempotent property also helps improve convolution-based models' ability to handle non-uniform motion blur.}
In the subsequent sections, we will present our idempotent deblurring framework in detail, which makes our network output idempotent deblurred results.
Experiments show that idempotent constraint enables the model to better deal with non-uniform motion blur.

\subsection{Definition of Idempotent Network}
Mathematically, an operation $f(\cdot)$ is called idempotent if and only if $f(f(\cdot)) = f(\cdot).$
This equation means a certain operation can be applied multiple times and still maintains the primary result. 

In this paper, we extend this concept to deep neural networks. Given an input $\mathbf{x}$, a network $\Phi$ with parameters $\Theta$ is called an idempotent network if and only if the outputs by the repeated implementation are the same, \ien,

\begin{equation}
\label{equa:idempotent_network}
\begin{aligned}
\Phi(\Phi(\mathbf{x}; \Theta); \Theta) = \Phi(\mathbf{x}; \Theta); \\
\Phi^k(\mathbf{x}; \Theta) = \Phi(\mathbf{x}; \Theta),
\end{aligned}
\end{equation}
where $k$ is a positive integer, indicating the number of repeating times. The above equation implies that for a deep deblurring network, given a pair of blurry and sharp images $(I_B, I_S)$, if we feed the blurry image $I_B$ into the network, the deblurred results $\hat{I}$ should be idempotent nevertheless how many times deblurring operations have been applied.

\subsection{Idempotent Network Architecture}

\begin{figure*}[t] 
\centering 
\includegraphics[width=1\textwidth]{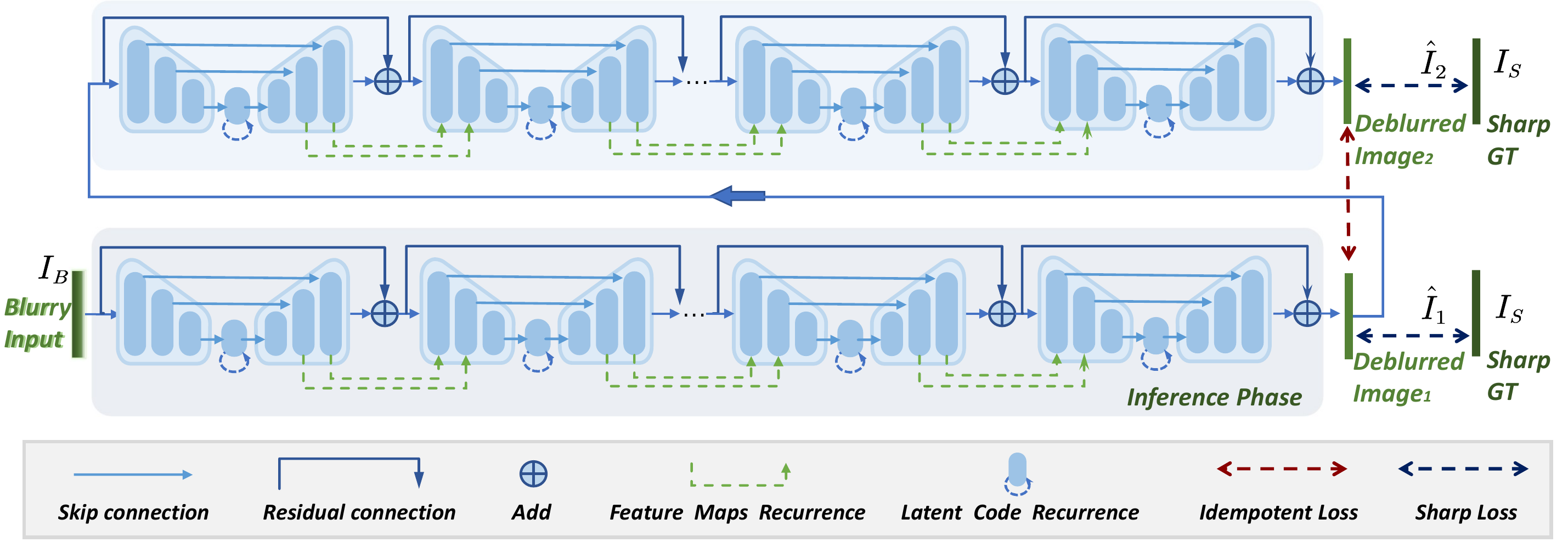}
\caption{\textbf{The overall framework of our idempotent deblurring network and idempotent constraint.} The deblurring network takes blurry images as input and outputs deblurred images by an iterative recurrent lightweight encoder-decoder structure. In all iterations, we only use the same basic model with shared weights and connect them by residual connection. The idempotent loss makes the outputs by repeating re-deblurring consistently in the training phase.}
\label{fig:OurFramework}
\end{figure*}

We propose a simple yet efficient single image blind deblurring network, the structure of this network is illustrated in Fig.~\ref{fig:OurFramework}.
This network uses a shared weight basic lightweight encoder-decoder module with a recurrent structure to achieve iterative residual deblurring.
Then we repetitively implement this network by taking the previous deblurred output as the next deblurring input, and then apply a novel idempotent constraint to these output results. Next, we will introduce each component of our idempotent network.


\begin{figure}[ht] 
\centering 
\includegraphics[width=0.48\textwidth]{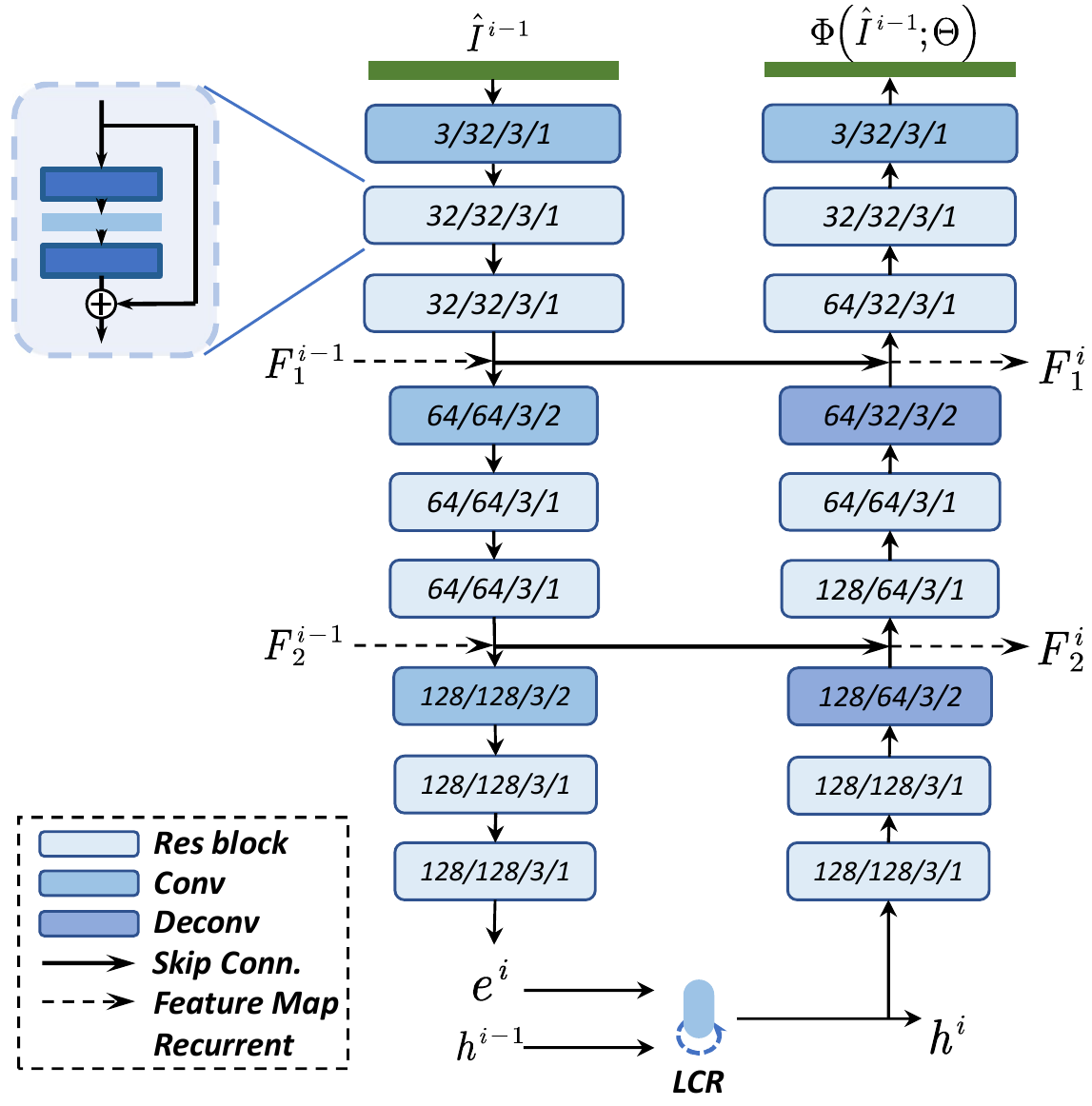}
\caption{\textbf{The structure of our encoder and decoder.} The number on each block denotes parameters of the residual block, convolution and deconvolution layers. From left to right are In Channel, Out Channel, Kernel size, and Stride. And the layers in the residual block are convolution-ReLU-convolution with skip connection.}
\label{fig:EncoderDecoder}
\end{figure}

\noindent\textbf{Basic Encoder-Decoder Deblurring Unit.}
To implement a simple network structure with minimal parameters, we first use a lightweight U-Net~\cite{ronneberger_UNet_2015} like encoder-decoder module as our basic unit for iterative residual deblurring,
\begin{equation}
\label{equa:Basic}
{\hat{I}^i} = \Phi_{\rm Basic}(\hat{I}^{i-1}; \Theta),
\end{equation}
where $i$ is the index of the iterations of the overall structure ($i=1,2,...,N$), and $N$ is the total number of iterations. $\hat{I}^i, \hat{I}^{i-1}$ are the deblurred images at the $i$-th and the ($i$-1)-th iteration. Specially, when $i$ equals to 1, $\hat{I}^{i-1}$ indicates the input blurry image.
$\Phi_{\rm Basic}$ is the basic deblurring unit, which has trainable parameters $\Theta$.

As shown in Fig.~\ref{fig:EncoderDecoder}, there are 15 convolutional layers with 6 residual connections~\cite{he_ResNet_CVPR_2016} and 6 ReLU activation functions in the encoder and decoder, respectively. Each of these residual blocks consists of two convolution layers with stride$=$1, a residual connection, and a ReLU layer, which do not change the size of feature maps. 
In addition, two convolution layers with stride$=$2 are interposed between the residual blocks in the encoder to downsample the feature maps.
Correspondingly, the decoder has two transpose convolutional layers in a symmetrical position for upsampling.
In all convolution layers, the kernel size is 3$\times$3. 
We also add a skip connection between corresponding encoder and decoder levels to fuse feature and enhance feature representation, this can facilitate network learning. Note that the residual connection uses an addition operation, while the skip connection uses concatenation.

\noindent\textbf{Progressive Residual Deblurring.} Iterative stacking or multi-scale structure has been widely used in existing deblurring networks~\cite{zhang_DMPHN_CVPR_2019, park_MTRNN_ECCV_2020, nah_DeepDeblur_CVPR_2017, tao_SRN_CVPR_2018}. To progressively achieve better deblurring performance, we use the basic deblurring unit with shared parameters to estimate the residual image for iterative deblurring. As shown in Fig.~\ref{fig:OurFramework}, each output of the deblurring unit is obtained as the sum of the input image and the residual images:
\begin{equation}
\label{equa:PRD}
\hat{I}^i = \hat{I}^{i-1} + \Phi_{\rm Basic}(\hat{I}^{i-1};\Theta).
\end{equation}

Given a blurry image ${I}_B = \hat{I}^{0}$ as input, our model outputs the final deblurred image $\hat{I}^{N}$ through iterative residual deblurring. Without using additional data supervision like MT-RNN~\cite{park_MTRNN_ECCV_2020}, our model learns to progressively deblur the input images. A visual comparison is shown in Fig.~\ref{fig:fig_progress}, which demonstrates the progressive learning ability of our residual deblurring structure.

\noindent\textbf{Feature Maps Recurrence.} We add a recurrent structure to the basic residual network to establish the relationship between two adjacent iterations. This mechanism recurs feature maps $F_1^{i-1}, F_2^{i-1}$ from the two last residual blocks in decoder after upsampling layer at the $(i-1)$-th iteration. And then concatenate $F_1^{i-1}, F_2^{i-1}$ corresponding with the feature maps in encoder at the $i$-th iteration before downsampling layer. 
(\cf~Fig.~\ref{fig:EncoderDecoder}). Specifically, this structure is modeled as:
\begin{equation}
\label{equa:FMR}
{\hat{I}^i, F_1^i, F_2^i} = \Phi_{\rm FMR}(\hat{I}^{i-1}, F_1^{i-1}, F_2^{i-1};\Theta),
\end{equation}
where $F_1^{i-1}, F_2^{i-1}$ are the feature maps from the decoder in the $(i-1)$-th iteration as the $i$-th iteration input, and output feature maps $F_1^{i}, F_2^{i}$ for next iteration.



\noindent\textbf{Latent Code Recurrence.}
The feature map recurrence we proposed above can only retain the information between two adjacent iterations.
As first used in SRN-Deblur~\cite{tao_SRN_CVPR_2018}, Long Short-Term Memory (LSTM)~\cite{hochreiter_LSTM_NC_1997, xingjian_ConvLSTM_NIPS_2015} can embed long term blur patterns across multiple deblurring iterations. The hidden state in LSTM retains the feature information from the previous iterations. 
To achieve similar performance with smaller parameters and cheaper computation, we use the Gated Recurrent Unit (GRU)~\cite{Cho_GRU_EMNLP_2014} as our Latent Code Recurrence (LCR) module in our network. 

Since the network is recurrent during multiple deblurring iterations, the feature maps from the last convolution layer in the encoder of each iteration are fed into GRU as the latent code. After passing through the memory unit, the feature maps can conditionally retain and forget the information of previous iterations, then can be used as the input of the decoder to restore the residual images of this iteration. 
The calculation process is modeled as follows:
\begin{equation}
\label{equa:GRU}
\begin{aligned}
z^{i} & = {\rm sigmoid}({\rm Conv}([h^{i-1}, e^{i}], \Theta_z)), \\
r^{i} & = {\rm sigmoid}({\rm Conv}([h^{i-1}, e^{i}], \Theta_r)), \\
\hat{h}^{i} & = {\rm tanh}({\rm Conv}([r^{i} \odot h^{i-1}, e^{i}], \Theta_h)), \\
h^{i} & = (1-z^{i}) \odot h^{i-1} + z^{i} \odot \hat{h}^{i},  \\
\end{aligned}
\end{equation}
where $[\cdot]$ is the concatenate operation and $\odot$ represents the Hadamard product. $z^{i}$, $r^{i}$ and $\hat{h}_{i}$ are the update gate, reset gate and candidate activation vector in GRU. $\Theta_z, \Theta_r, \Theta_h$ are the parameters of each convolution layer, respectively.

Thus, our basic unit for progressive residual deblurring with feature maps recurrence and latent code recurrence can be modeled as follows:
\begin{equation}
\label{equa:LTSR}
{\hat{I}^i, F_1^i, F_2^i, h^i} = \Phi_{\rm LCR}(\hat{I}^{i-1}, F_1^{i-1}, F_2^{i-1}, h^{i-1};\Theta),
\end{equation}
where $h^{i-1}, h^i$ represent the hidden state output by previous iteration $(i-1)$ and current iteration $i$.

\noindent\textbf{Idempotent Re-Deblurring by Multiple Times.}
\label{section:idemrepetredeblur}
It is difficult for a deep network to output the same results as the input, so we propose a idempotent constraint to train our deblurring network.
As shown in Fig.~\ref{fig:OurFramework}, our model first outputs the deblurring result $\hat{I}_1 = \hat{I}^{N}_1$, then takes it as next input and get the re-deblurring result $\hat{I}_2 = \hat{I}^{N}_2$ ($N$ denotes the number of iterations) for multiple times. 

Note above re-deblurring only exists in training, and in the inference phase, we only need to apply our model once to get the deblurred result.
Because of the residual deblurring structure, the idempotent constraint can be satisfied when the sum of all the residual outputs is close to zero. 


\subsection{Idempotent Constraint}
To keep the deblurring network output idempotent and enhance the deblurring performance with idempotent property, we enforce the idempotent constraint as a loss function to supervise the training process. We term such loss function as an idempotent loss. 
As shown in Fig.~\ref{fig:OurFramework}, the network is repeated twice and we directly constrain these two deblurring outputs to be consistent. Therefore, the idempotent loss is defined \Fix{by the ${L}_1$ distance} as:
\begin{equation}
\label{equa:Idem_loss}
\mathcal{L}_{Idem} = \|\hat{I}_1 - \hat{I}_2\|_1,
\end{equation}
where $\hat{I}_1$ and $\hat{I}_2$ denote two latent sharp image outputs by deblurring and re-deblurring.


We use pairs of blurry and sharp images ($I_B$, $I_S$) to supervise the training of our deblurring network, and then calculate the \Fix{${L}_1$ distance} as our deblurring loss at each output by re-deblurring: 
\begin{equation}
\label{equa:sharp_loss}
\mathcal{L}_{Sharp} = \sum_{j=1}^{2} {\alpha_{j} \|\hat{I}_{j} - I_S\|_1}.
\end{equation}

The final objective function is reached as:
\begin{equation}
\label{equa:Total_loss}
\mathcal{L} = \lambda \mathcal{L}_{Idem} + \mathcal{L}_{Sharp},
\end{equation}
where $\alpha_{j}$ and $\lambda$ are the trade-off parameters.

\section{Experiments}
\subsection{Experimental Details}
\noindent\textbf{Datasets.} \Fix{Following the general setting of single image deblurring task~\cite{nah_DeepDeblur_CVPR_2017, kupyn_DeblurGAN_CVPR_2018, zhang_DMPHN_CVPR_2019, park_MTRNN_ECCV_2020, cai_DBCPeNet_TIP_2020, waqas_MPRNet_CVPR_2021, purohit_RADN_AAAI_2020, tao_SRN_CVPR_2018}},
we use the GoPro dataset~\cite{nah_DeepDeblur_CVPR_2017} to train our proposed model.
The blurry images of the GoPro dataset are synthesized by averaging different numbers (7–13) of successive latent frames from 240 FPS video sequences captured by a GoPro Hero 4 camera.
As a common benchmark for image motion deblurring, it contains 3,214 blurry-sharp image pairs. We follow the widely used split method as~\cite{nah_DeepDeblur_CVPR_2017} and use 2,103 pairs from the linear subset for training and the remaining 1,111 pairs as the test set for evaluation.

\Fix{We also evaluate the generalization ability of our method on a real-world blurry scenes dataset, \ien, RealBlur~\cite{rim_RealBlurDataset_ECCV_2020}, a commonly used dataset in recent years.}
It contains two versions named as \textit{RealBlur-J} (JPEG compressed) and \textit{RealBlur-R} (RAW), where each version includes 980 pairs of geometrically aligned real-world blurry and ground-truth sharp images captured by a well-designed image acquisition system composed of a beam splitter. Following the original settings of the RealBlur dataset, we also conduct photometric alignment between the outputted deblurred images and ground-truth sharp images before computing PSNR and SSIM.

\noindent\textbf{Implementation Details.}
We implement our model\footnote{Our code and pre-trained model will be released.} by PyTorch~\cite{paszke_pytorch_NIPS_2019} with two NVIDIA RTX 2080Ti GPU for training and one for evaluation. The trade-off parameter $\alpha_{1}$, $\alpha_{2}$ is set to $1.0$ and $\lambda$ is $0.1$, respectively. We define the number of iterations $N$ as 6. All weights are initialized from scratch by the Xavier method~\cite{glorot_Xavier_ICAIS_2010}. The Adam~\cite{Kingma_Adam_ICLR_2015} solver is used to optimize our network for 3,000 epochs with default parameters $\beta_1=0.9$, $\beta_2=0.999$ and $\epsilon=10^{-8}$. The training mini-batch size is set to 6. The initial learning rate is $10^{-4}$ and decays by $0.5$ after every 500 epochs. Following DMPHN~\cite{zhang_DMPHN_CVPR_2019}, the input blurry images are normalized to range $[0, 1]$ and subtracted by $0.5$. In training, the input blurry images and corresponding ground-truth sharp images are randomly cropped to 256$\times$256 pixels patches. Additionally, we randomly rotate and/or flip the image patches for data augmentation. The color saturation is also randomly changed on the input images for robust learning.

\subsection{Comparison with the State-of-the-art Methods}
In this subsection, we perform quantitative and qualitative studies with existing state-of-the-art methods on the benchmark datasets (\ien~GoPro and RealBlur).

\begin{table}[t]
    \caption{\textbf{Quantitative results on the GoPro test dataset~\cite{nah_DeepDeblur_CVPR_2017}} in terms of PSNR, SSIM, number of parameters and inference time. The 1st, 2nd and 3rd best performances are highlighted with \textcolor{red}{red}, \textcolor{blue}{blue} and \textcolor{green}{green} (Best viewed in color).}
    \label{tab:GoPro_benchmark}
    \centering
        \begin{threeparttable}
        \footnotesize
        \begin{tabular}{lccrr}
        \toprule
        Methods                                   & PSNR~(dB)        & SSIM    & Param.  & Time(s) \cr
        \midrule
        Sun~\etal~\cite{sun_CVPR_2015}             & 24.64    & 0.843    & -     & -   \cr
        MS-CNN~\cite{nah_DeepDeblur_CVPR_2017}     & 29.23    & 0.914    & 21.0M & 0.94 \cr
        DeblurGAN~\cite{kupyn_DeblurGAN_CVPR_2018} & 28.70    & \textcolor{blue}{0.958}   & 3.50M & 0.12 \cr
        DeblurGANv2~\cite{kupyn_DeblurGANv2_ICCV_2019}  & 29.55 & 0.934  & 60.9M & 0.058 \cr
        SRN-Deblur~\cite{tao_SRN_CVPR_2018}          & 30.26  & 0.931  & 3.76M & 0.25 \cr
        Gao~\etal~\cite{gao_NestedSkip_CVPR_2019}    & 30.92  & 0.942  & \textcolor{blue}{2.84M} & 0.76  \cr
        MT-RNN~\cite{park_MTRNN_ECCV_2020}           & 31.15  & 0.945  & \textcolor{red}{2.63M} & 0.048 \cr 
        DMPHN~\cite{zhang_DMPHN_CVPR_2019}           & 30.25  & 0.935  & 7.23M & \textcolor{blue}{0.032} \cr
        Stack(4)-DMPHN~\cite{zhang_DMPHN_CVPR_2019}  & 31.20  & 0.945  & 28.9M & 0.35  \cr
        Stack(2)-VMPHN~\cite{zhang_DMPHN_CVPR_2019}  & 31.50  & 0.948  & 28.9M & (0.55)  \cr
        MSCAN~\cite{wan_channel_attention_TCSVT_2021} & 31.24  & 0.945  & 7.5M  & 0.42 \cr
        RADN~\cite{purohit_RADN_AAAI_2020}           & 31.76  & \textcolor{green}{0.953}  & -   & \textcolor{green}{(0.038)} \cr
        SAPHN~\cite{suin_AttentionDMPHN_CVPR_2020}   & \textcolor{green}{31.85}  & 0.948  & 24.0M & (0.28)  \cr
        MPRNet(1)~\cite{waqas_MPRNet_CVPR_2021}      & 30.43  & - & 5.6M & (0.04)  \cr
        MPRNet(2)~\cite{waqas_MPRNet_CVPR_2021}      & 31.81  & - & 11.3M & (0.08)  \cr
        MPRNet(3)~\cite{waqas_MPRNet_CVPR_2021}      & \textcolor{red}{32.66}  & \textcolor{red}{0.959} & 20.1M & 0.15  \cr
        \midrule
        \textbf{Ours}(w/o Idem.)  & 31.80   & 0.949   & \textcolor{green}{3.11M}   & \textcolor{red}{0.028} \cr 
        \textbf{Ours}(w/ Idem.)   & \textcolor{blue}{31.92}   & \textcolor{green}{0.953}   & \textcolor{green}{3.11M}   & \textcolor{red}{0.028} \cr 
        \toprule
        \end{tabular}
        \end{threeparttable}
\end{table}

\noindent\textbf{Quantitative Evaluations.}
Following the common experimental setting as previous works~\cite{suin_AttentionDMPHN_CVPR_2020, purohit_RADN_AAAI_2020, zhang_DMPHN_CVPR_2019, park_MTRNN_ECCV_2020}, we compare the performance and generalization ability across different datasets of our deep idempotent deblurring network with previous state-of-the-art deblurring methods in a quantitative way. The experimental results in terms of \textit{PSNR}, \textit{SSIM}, \textit{Parameters} and \textit{Time} for different deblurring methods on GoPro test dataset are shown in Table~\ref{tab:GoPro_benchmark}. For a fair comparison, we perform the experiments of running time on a single RTX 2080Ti GPU. Except for Stack(2)-VMPHN~\cite{zhang_DMPHN_CVPR_2019}, RADN~\cite{purohit_RADN_AAAI_2020} and MPRNet-(1)\&(2)~\cite{waqas_MPRNet_CVPR_2021} are from their paper which do not provide the opensource code or the corresponding model, SAPHN~\cite{suin_AttentionDMPHN_CVPR_2020} are provided by the authors (bracketed 
in Table~\ref{tab:GoPro_benchmark}).

On the GoPro test dataset, our method achieves comparable high performance to state-of-the-art methods with smaller parameters and faster inference time. 
In particular, although MPRNet(3)~\cite{waqas_MPRNet_CVPR_2021} achieves higher performance, our model is nearly 6.5$\times$ smaller and 6.4$\times$ faster than it.
The comparison with MPRNet(2) shows that we achieve equivalent performance with fewer parameters and shorter inference time. 
Moreover, the adversarial loss improves the visual quality but may sacrifice the pixel-wise metric results, while the SSIM metric prefers the structural similarity rather than the pixel-wise intensity similarity. Therefore, the GAN-based methods~\cite{kupyn_DeblurGAN_CVPR_2018} tend to achieve better SSIM than other methods trained with the $L_1$ loss. And the $L_1$ loss often leads to smooth results.

We also perform a quantitative comparison of generalization results on the RealBlur~\cite{rim_RealBlurDataset_ECCV_2020} dataset for models only pre-trained on the GoPro dataset. The quantitative results are reported in Table~\ref{tab:RealBlur_benchmark}. We can observe that our model performs better compared to previous state-of-the-art methods, which confirms that our model is more robust in real-world scenes with better generalization ability across different datasets for image deblurring. 
Note that these results are from~\cite{rim_RealBlurDataset_ECCV_2020}, except for MT-RNN~\cite{park_MTRNN_ECCV_2020} which is reproduced by us. 
Due to the different image sizes of the RealBlur dataset, we test the average inference time of each model, and our model maintains a very fast inference speed.

\begin{table}[t] 
    \caption{\textbf{Quantitative analysis on the \textit{RealBlur} test dataset~\cite{rim_RealBlurDataset_ECCV_2020}} for models only pre-trained on GoPro dataset~\cite{nah_DeepDeblur_CVPR_2017}}
    \label{tab:RealBlur_benchmark}
    \centering
    \resizebox{0.5\textwidth}{!}{
        \begin{threeparttable}
        \begin{tabular}{lccccc}
        \toprule
        & \multicolumn{2}{c}{\textit{RealBlur-J}} & \multicolumn{2}{c}{\textit{RealBlur-R}} & \cr
        \cmidrule(lr){2-3}\cmidrule(lr){4-5}
        Methods & PSNR~(dB) & SSIM & PSNR~(dB) & SSIM & Time(s)\cr
        \midrule
        MS-CNN~\cite{nah_DeepDeblur_CVPR_2017}      & 27.87    & 0.827  & 32.51 & 0.841 & 0.77 \cr
        Stack(4)-DMPHN~\cite{zhang_DMPHN_CVPR_2019} & 27.80    & 0.847  & 35.48 & 0.947 & 0.29 \cr
        DeblurGAN~\cite{kupyn_DeblurGAN_CVPR_2018}  & 27.97    & 0.834  & 33.79 & 0.903 & 0.098 \cr
        MT-RNN~\cite{park_MTRNN_ECCV_2020}          & 28.44    & 0.862  & 35.77 & \underline{0.951} & \underline{0.039} \cr
        SRN-Deblur~\cite{tao_SRN_CVPR_2018}         & 28.56    & 0.867  & 35.66 & 0.947 & 0.20 \cr
        DeblurGANv2~\cite{kupyn_DeblurGANv2_ICCV_2019} & \underline{28.70} & 0.867  & 35.26 & 0.944 & 0.048 \cr
        MPRNet(3)~\cite{waqas_MPRNet_CVPR_2021}      & \underline{28.70} & \underline{0.873} & \textbf{35.99} & \textbf{0.952} & 0.16 \cr
        \midrule
        \textbf{Ours}(w/o Idem.)   & 28.69  & 0.868 & 35.79 & 0.949 & 0.023\cr
        \textbf{Ours}(w/ Idem.)    & \textbf{28.72}  & \textbf{0.876} & \underline{35.86}  & \underline{0.951} & \textbf{0.023} \cr
        \toprule
        \end{tabular}
        \end{threeparttable}
    }
\end{table}

\begin{table}[t]
    \caption{\textbf{Quantitative analysis on multi-blurring level synthetic dataset}. Different number in the first row means the number of successive latent frames used to synthesize the blurry images.}
    \label{tab:MultiLevelData}
    \centering
    \resizebox{0.48\textwidth}{!}{
        \begin{threeparttable}
        \begin{tabular}{ccccccc}
    \toprule
     Methods $\backslash$ PSNR(dB)        & 5       & 7       & 9       & 11      & 13      & 15  \cr
    \midrule
    MT-RNN~\cite{park_MTRNN_ECCV_2020}     & \textbf{35.17}   & 33.67   & 32.37   & 31.12   & 29.91   & 28.68 \cr
    Stack(4)-DMPHN~\cite{zhang_DMPHN_CVPR_2019} & 33.15   & 32.90   & 32.26   & 31.19   & 29.85   & 27.73 \cr
    \midrule
    \textbf{Ours}(w/o Idem.)               & 34.77   & 33.82   & 32.81   & 31.75   & 30.55   & 29.14 \cr
    \textbf{Ours}(w/ Idem.)                & 35.05 & \textbf{33.98} & \textbf{33.00} 
                                          & \textbf{31.92} & \textbf{30.72} & \textbf{29.31} \cr
    \toprule
        \end{tabular}
        \end{threeparttable}
    }
\end{table}

To further demonstrate the deblurring performance and generalization ability for different blur levels, we resynthesized a multi-blurring level dataset following the synthesis pipeline of the GoPro dataset~\cite{nah_DeepDeblur_CVPR_2017}. 
\Fix{Quantitative results in Table~\ref{tab:MultiLevelData} show that our model achieves better deblurring performance on the multi-blurring level dataset, indicating the superiority of our idempotent framework for dynamic scene non-uniform deblurring. 
Note that MT-RNN~\cite{park_MTRNN_ECCV_2020} uses blurry images of different blur levels for supervision during the training process. This strategy makes MT-RNN more advantageous when using datasets synthesized at blur level 5 or 7, and causes the performance of our network to be inferior to MT-RNN when the blur level is 5. However, when the blur level is 7-15, the performance of our proposed idempotent deblurring network is significantly better than MT-RNN and DMPHN.}

\begin{figure*}[ht!]
    \centering
    {\begin{minipage}[t]{0.24\textwidth}
    \centering\includegraphics[width=1\textwidth]{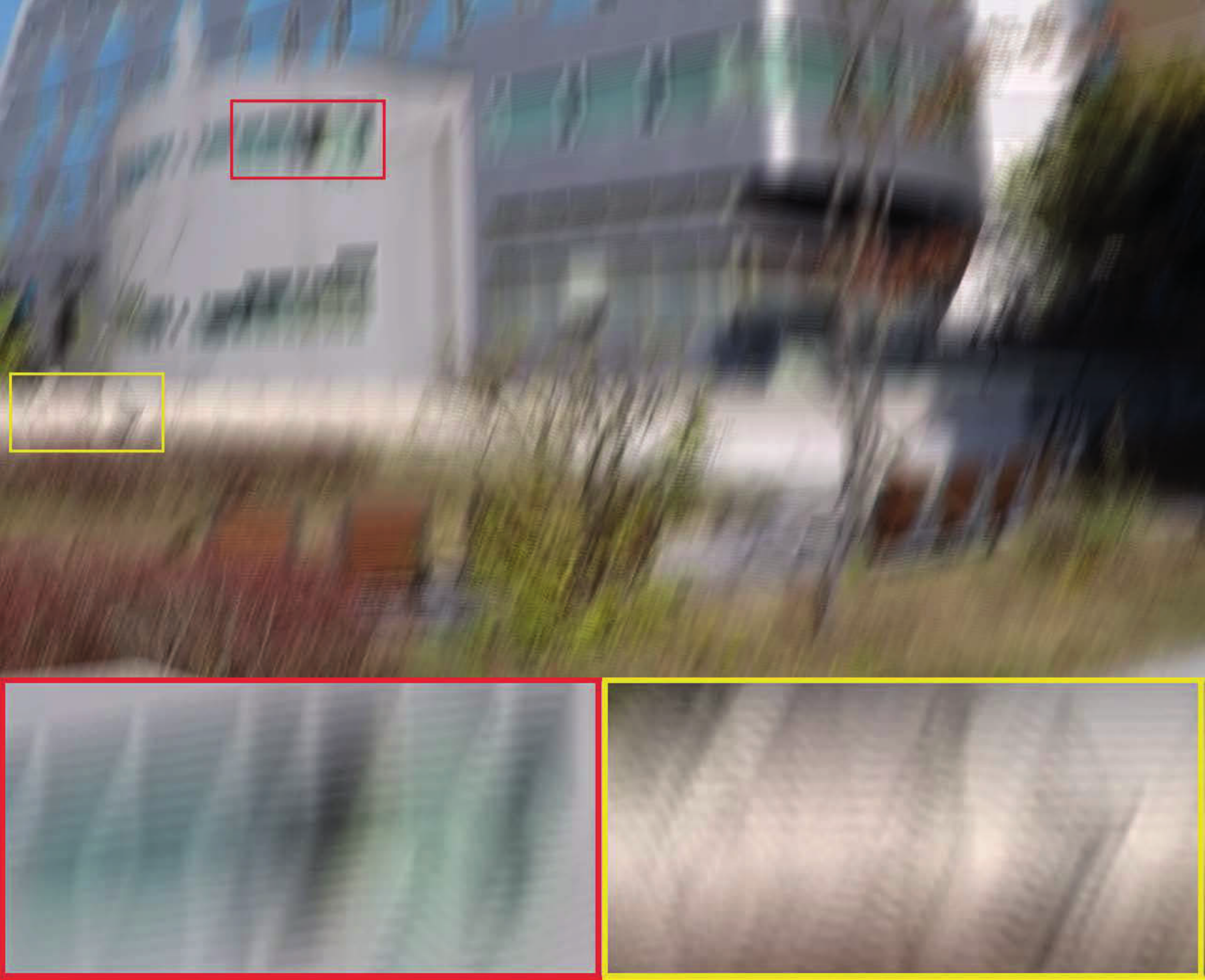}
    \end{minipage}}
    {\begin{minipage}[t]{0.24\textwidth}
    \centering\includegraphics[width=1\textwidth]{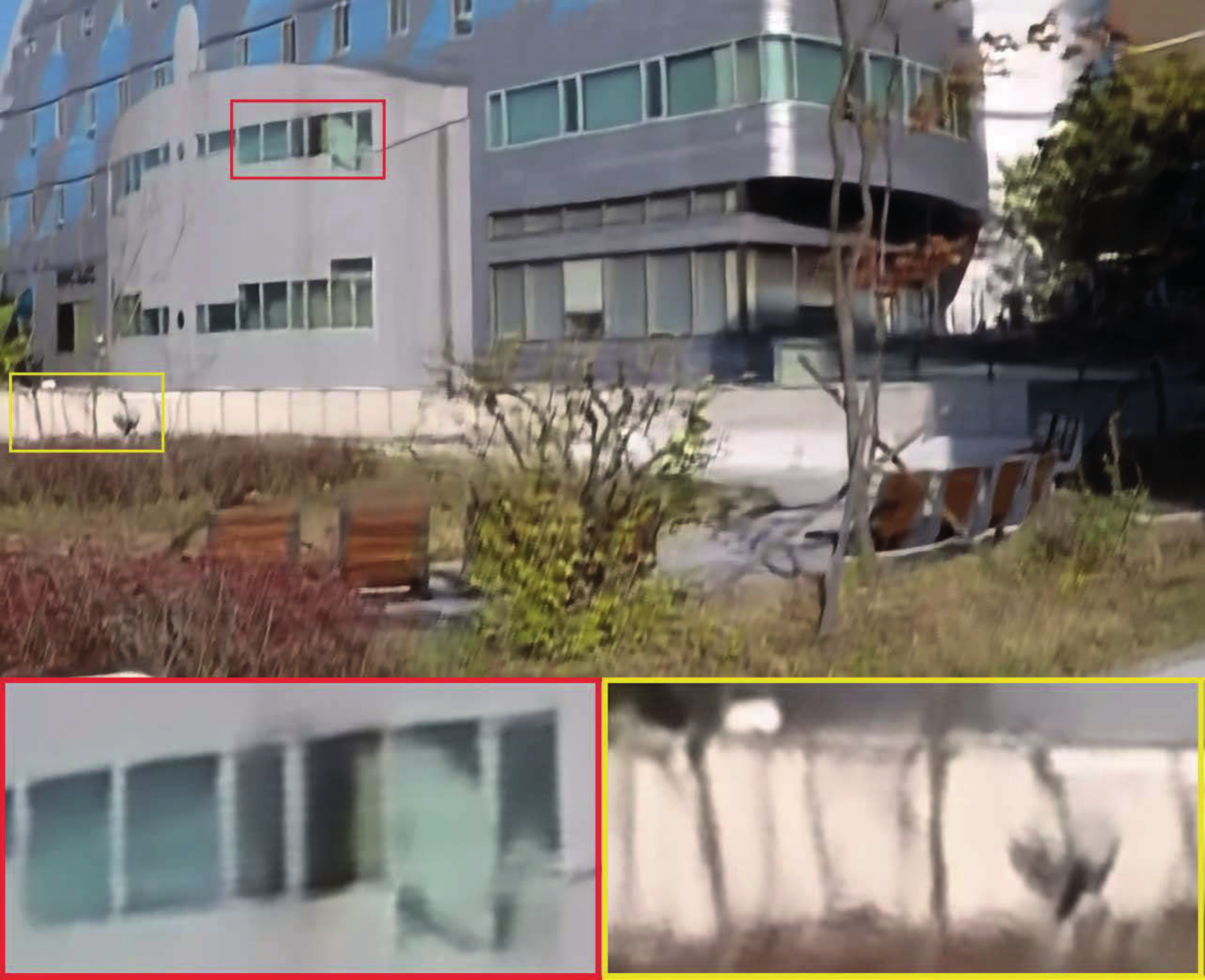}
    \end{minipage}}
    {\begin{minipage}[t]{0.24\textwidth}
    \centering\includegraphics[width=1\textwidth]{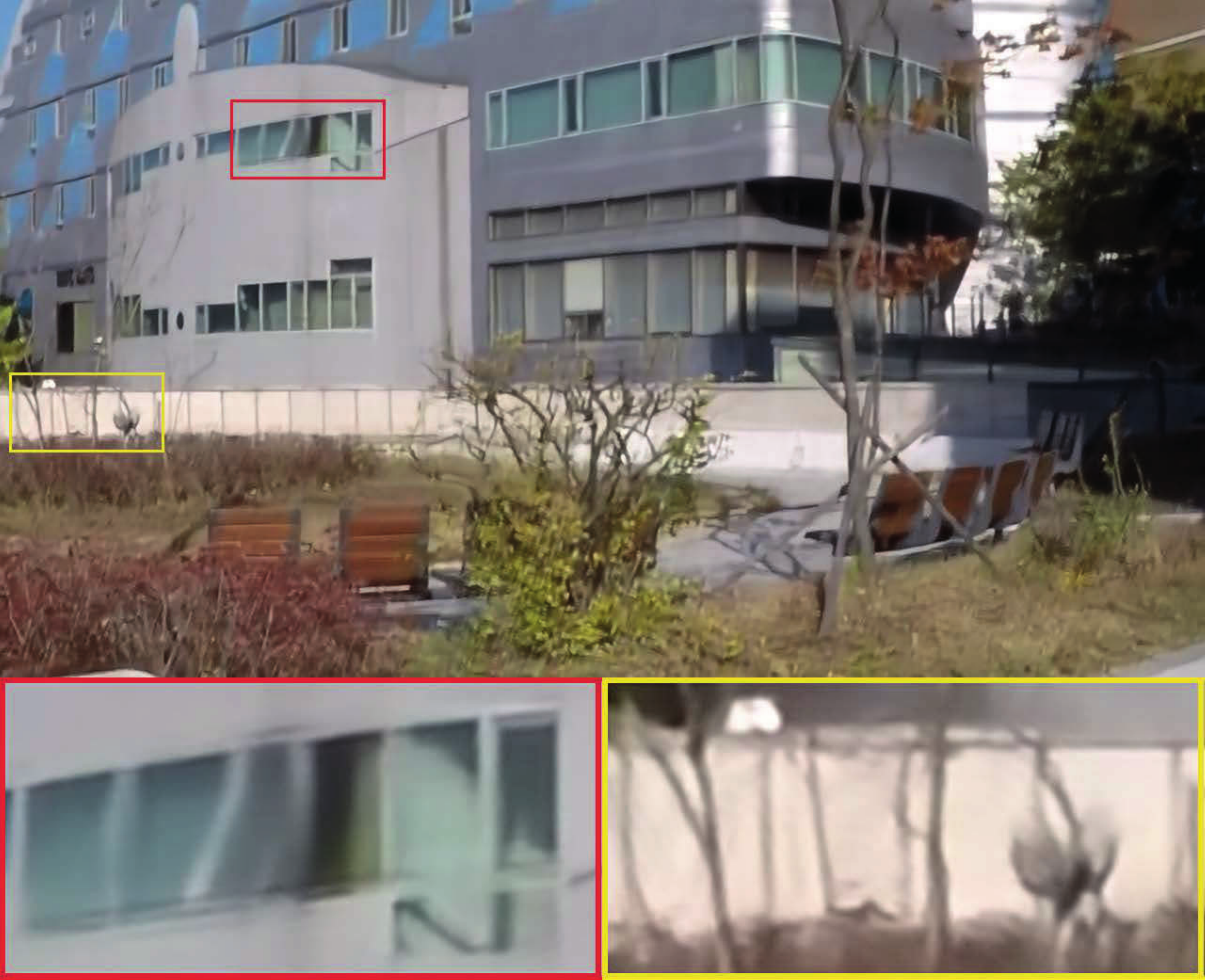}
    \end{minipage}}
    {\begin{minipage}[t]{0.24\textwidth}
    \centering\includegraphics[width=1\textwidth]{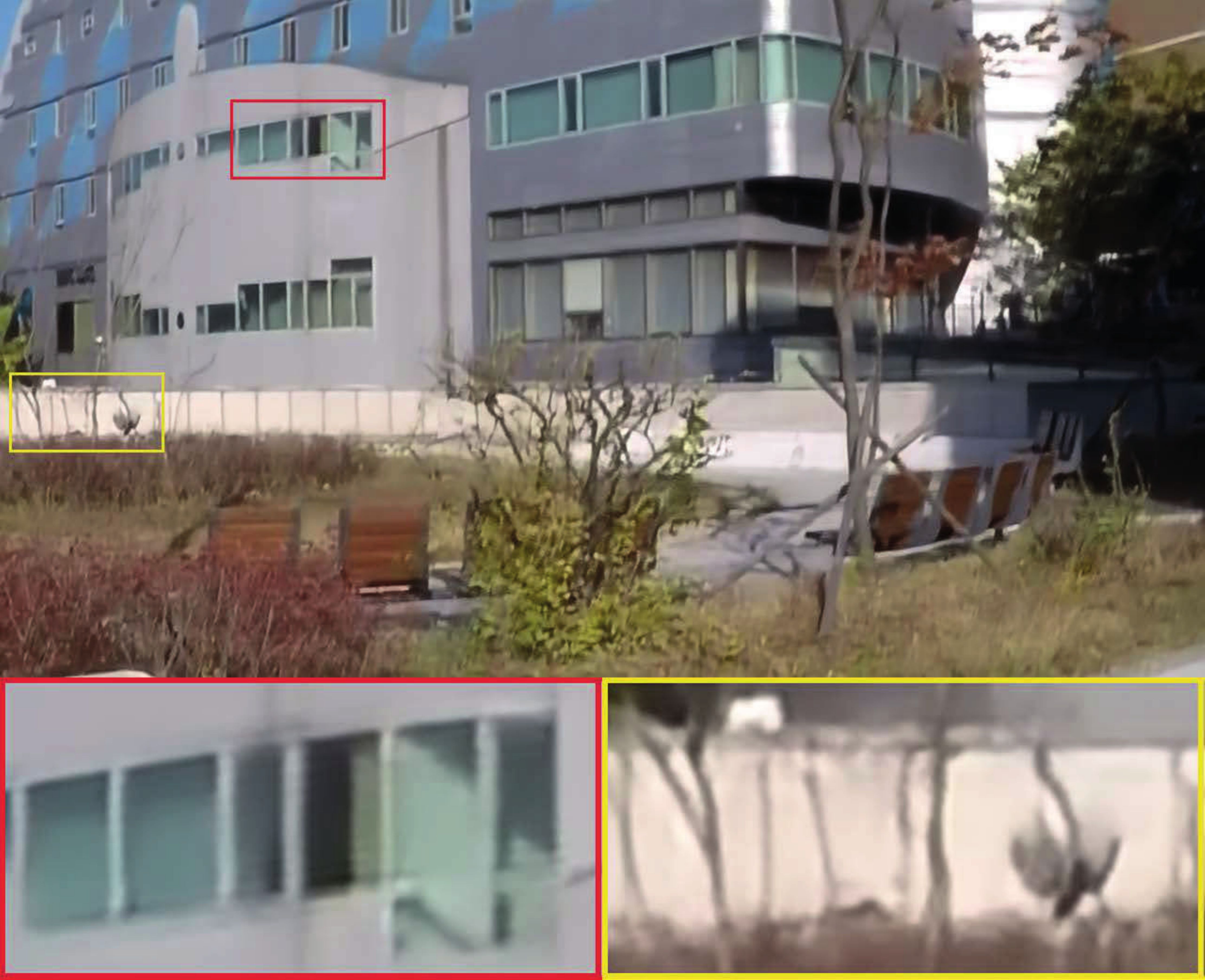}
    \end{minipage}} \\
    \vspace{-0.15cm}
    \subfigure[Blur input]{\begin{minipage}[t]{0.24\textwidth}
    \centering\includegraphics[width=1\textwidth]{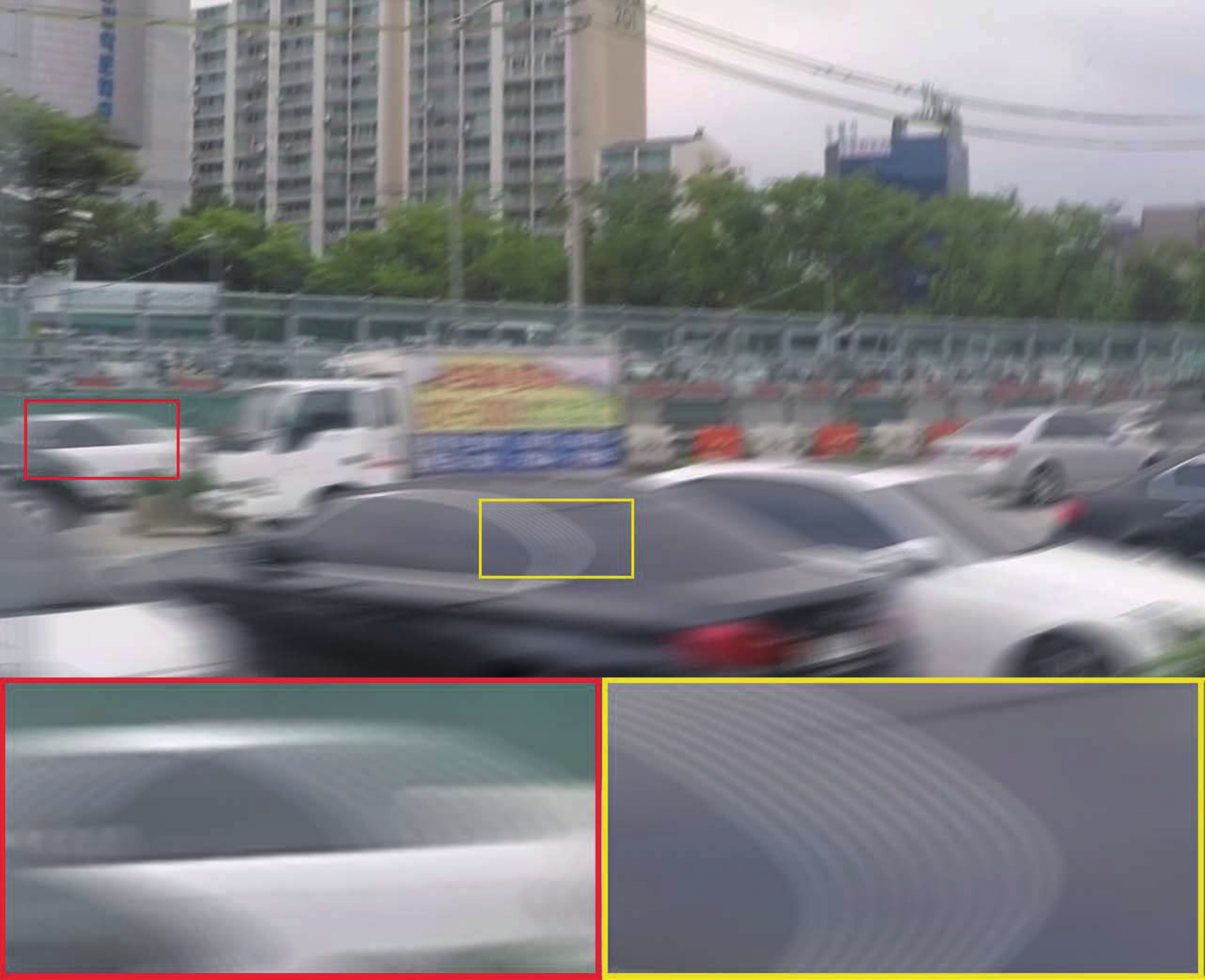}\end{minipage}}
    \subfigure[MT-RNN~\cite{park_MTRNN_ECCV_2020}]{\begin{minipage}[t]{0.24\textwidth}
    \centering\includegraphics[width=1\textwidth]{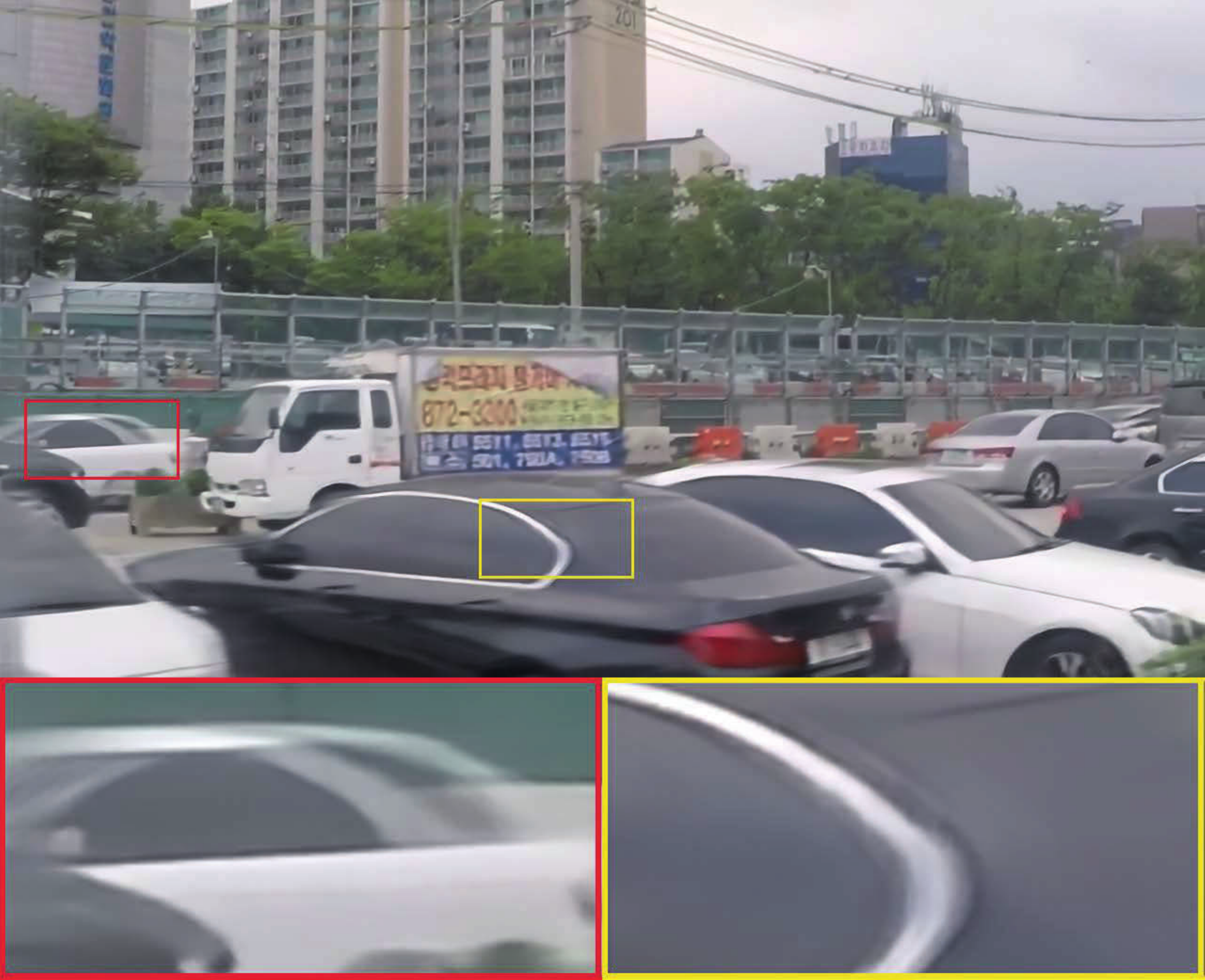}\end{minipage}}
    \subfigure[Stack(4)-DMPHN~\cite{zhang_DMPHN_CVPR_2019}]{\begin{minipage}[t]{0.24\textwidth}
    \centering\includegraphics[width=1\textwidth]{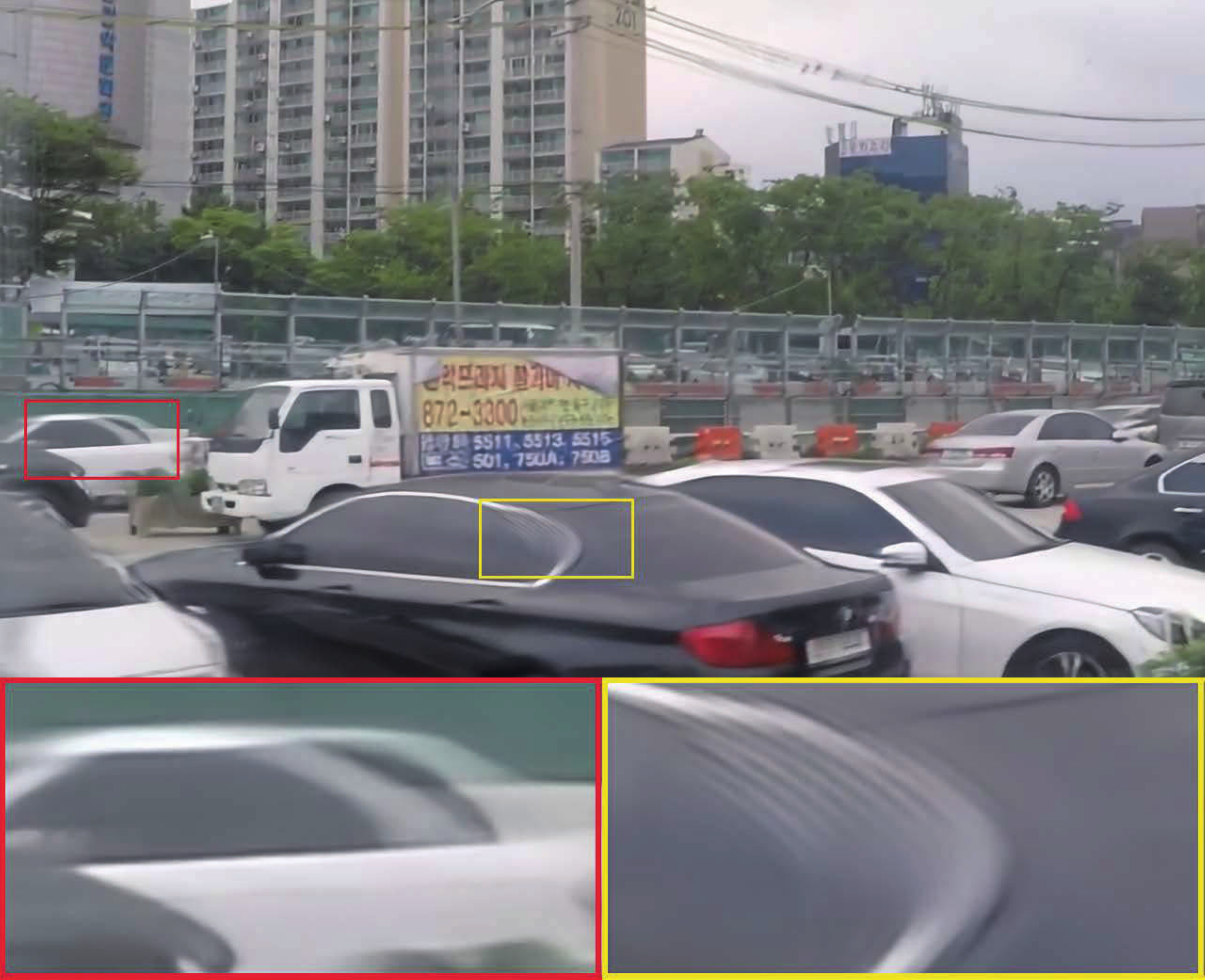}\end{minipage}}
    \subfigure[Ours]{\begin{minipage}[t]{0.24\textwidth}
    \centering\includegraphics[width=1\textwidth]{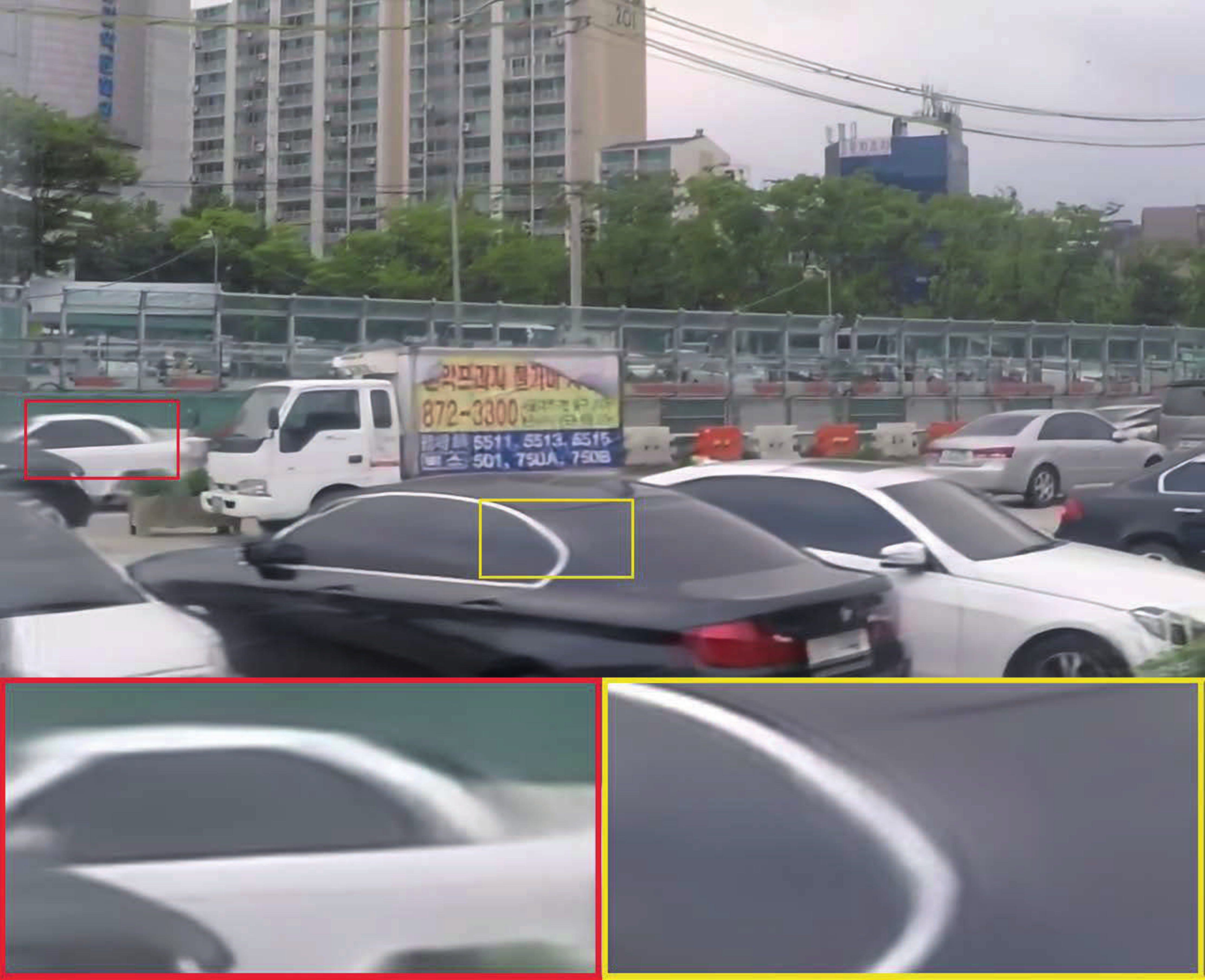}\end{minipage}} \\
    \centering
    \caption{\textbf{Visual comparisons on the GoPro testing dataset.} Column (a) is the original blurry images, (b), (c), (d) are the deblurring results from MT-RNN~\cite{park_MTRNN_ECCV_2020}, Stack(4)-DMPHN~\cite{zhang_DMPHN_CVPR_2019} and our method, respectively. Best Viewed on Screen.}
    \label{fig:compare_results_GoPro}
\end{figure*}

\begin{figure*}[ht!]
    \centering
    {\begin{minipage}[t]{0.24\textwidth}
    \centering\includegraphics[width=1\textwidth]{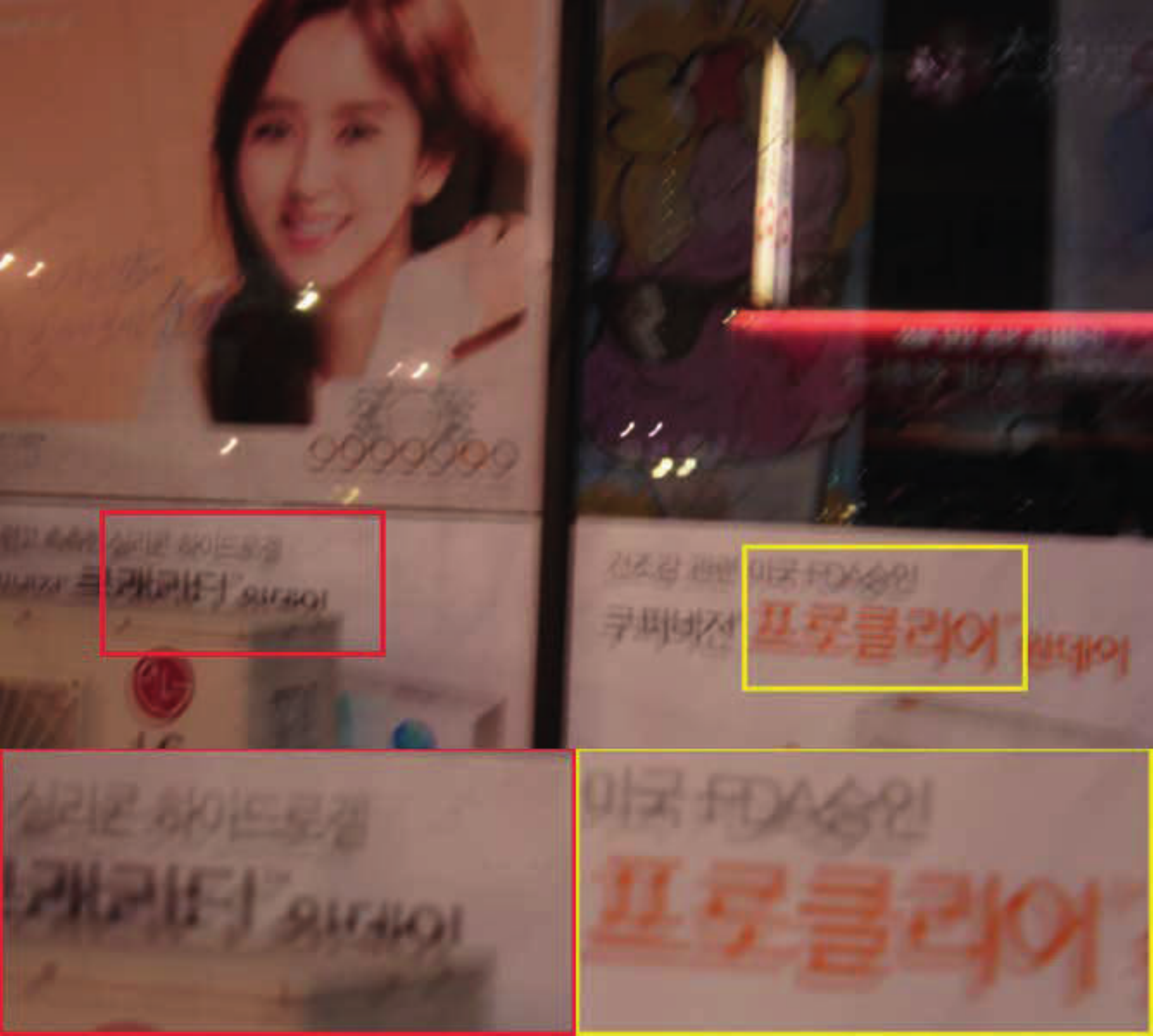}
    \end{minipage}}
    {\begin{minipage}[t]{0.24\textwidth}
    \centering\includegraphics[width=1\textwidth]{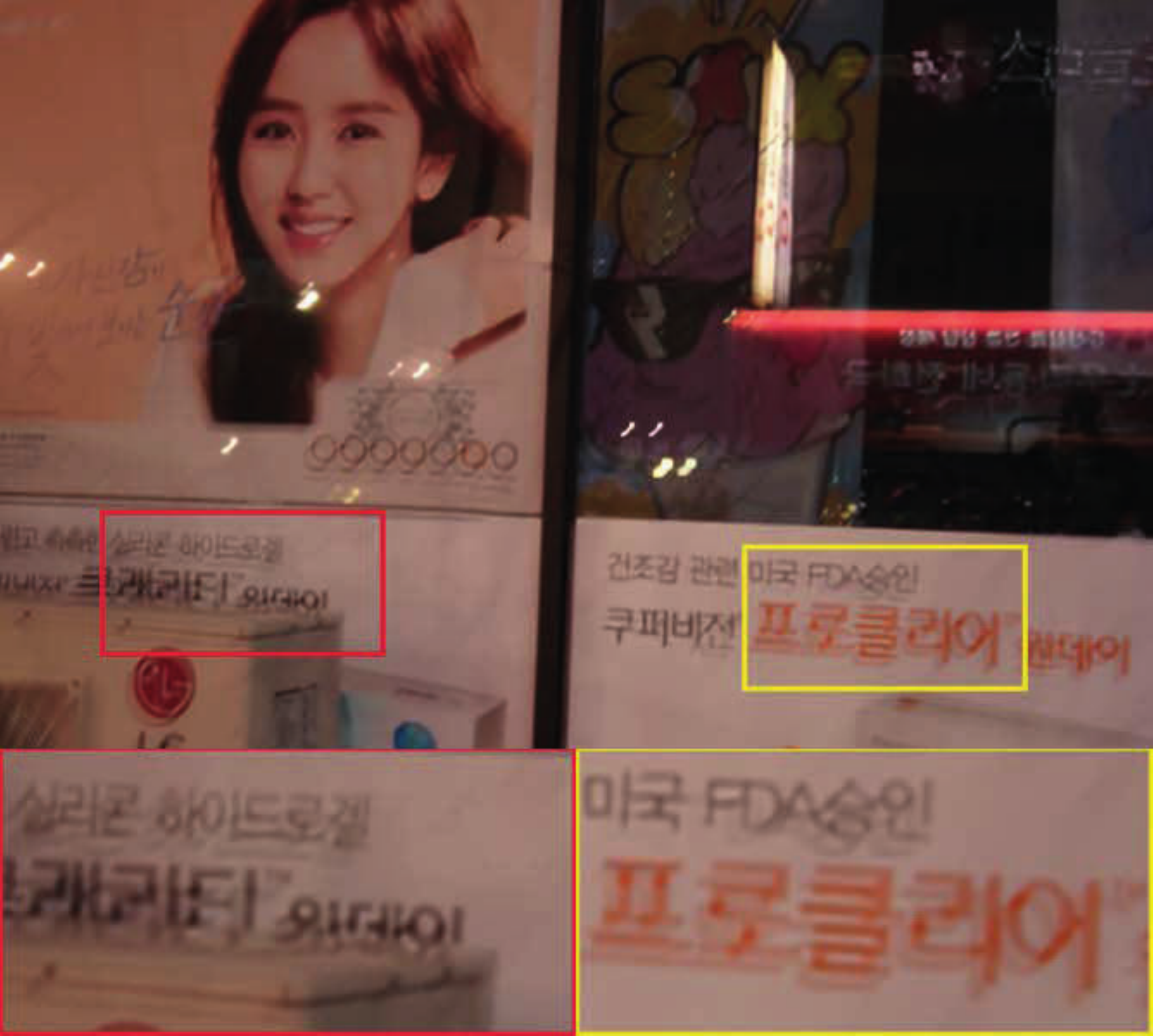}
    \end{minipage}}
    {\begin{minipage}[t]{0.24\textwidth}
    \centering\includegraphics[width=1\textwidth]{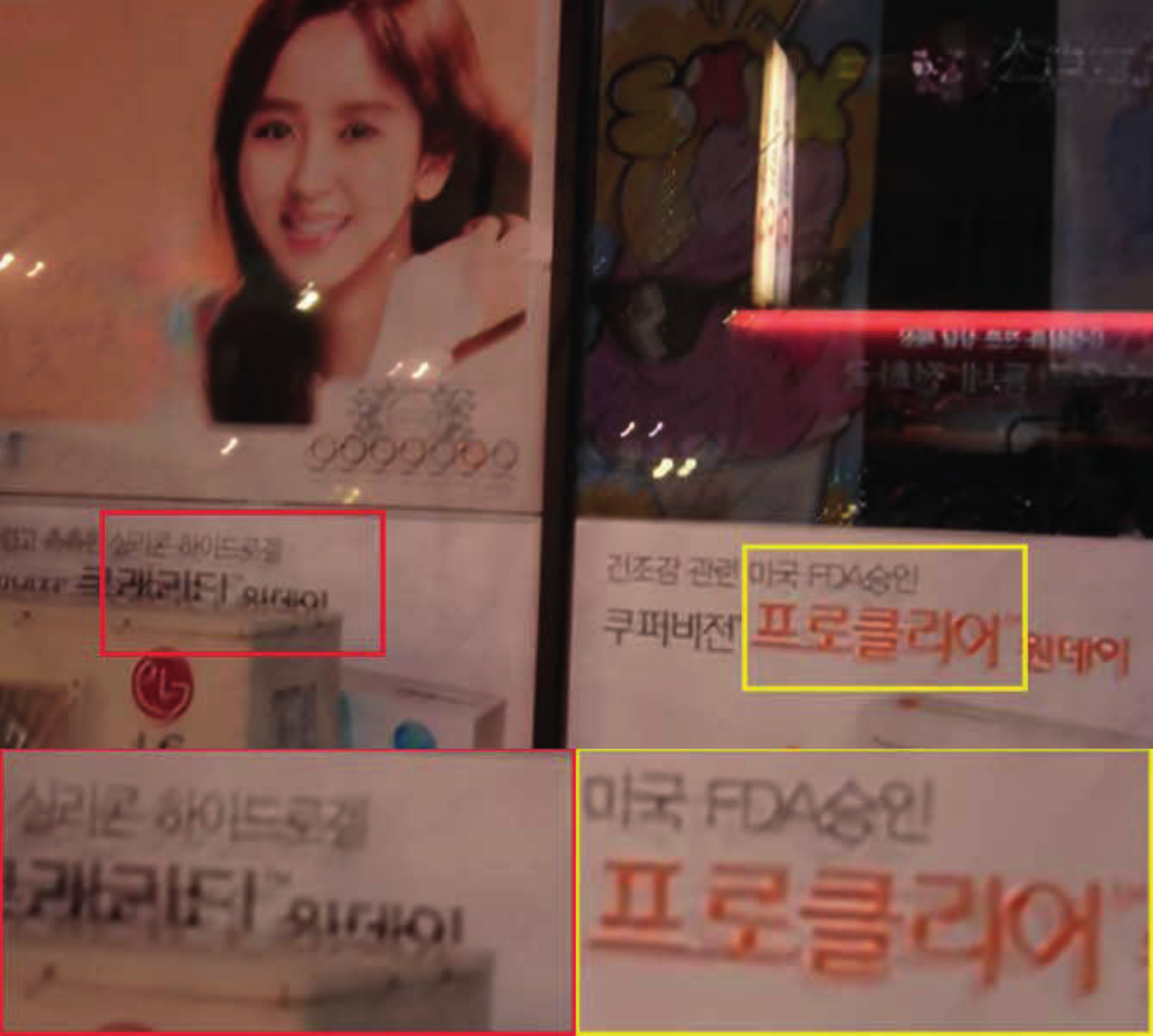}
    \end{minipage}}
    {\begin{minipage}[t]{0.24\textwidth}
    \centering\includegraphics[width=1\textwidth]{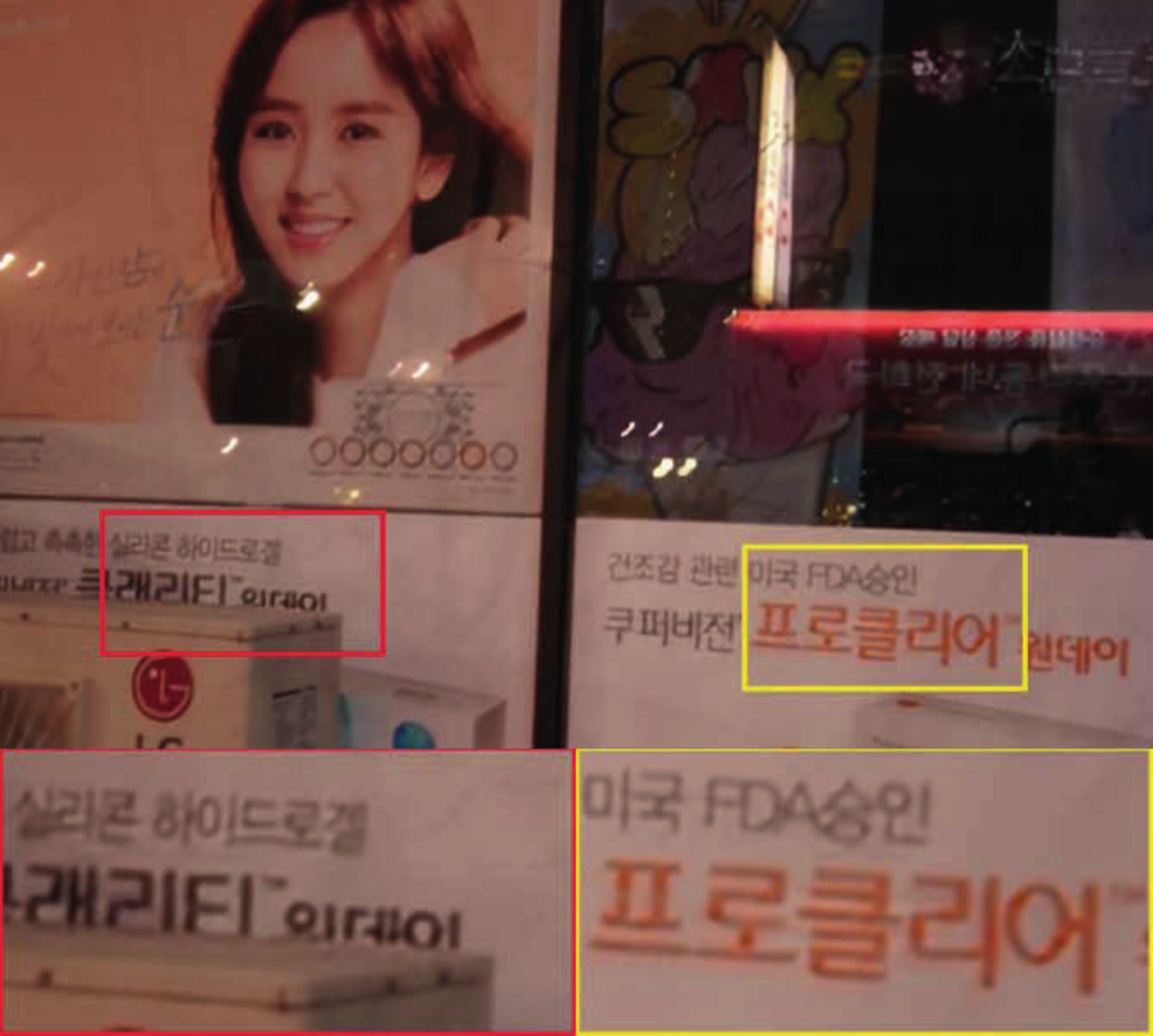}
    \end{minipage}} \\
    {\begin{minipage}[t]{0.24\textwidth}
    \centering\includegraphics[width=1\textwidth]{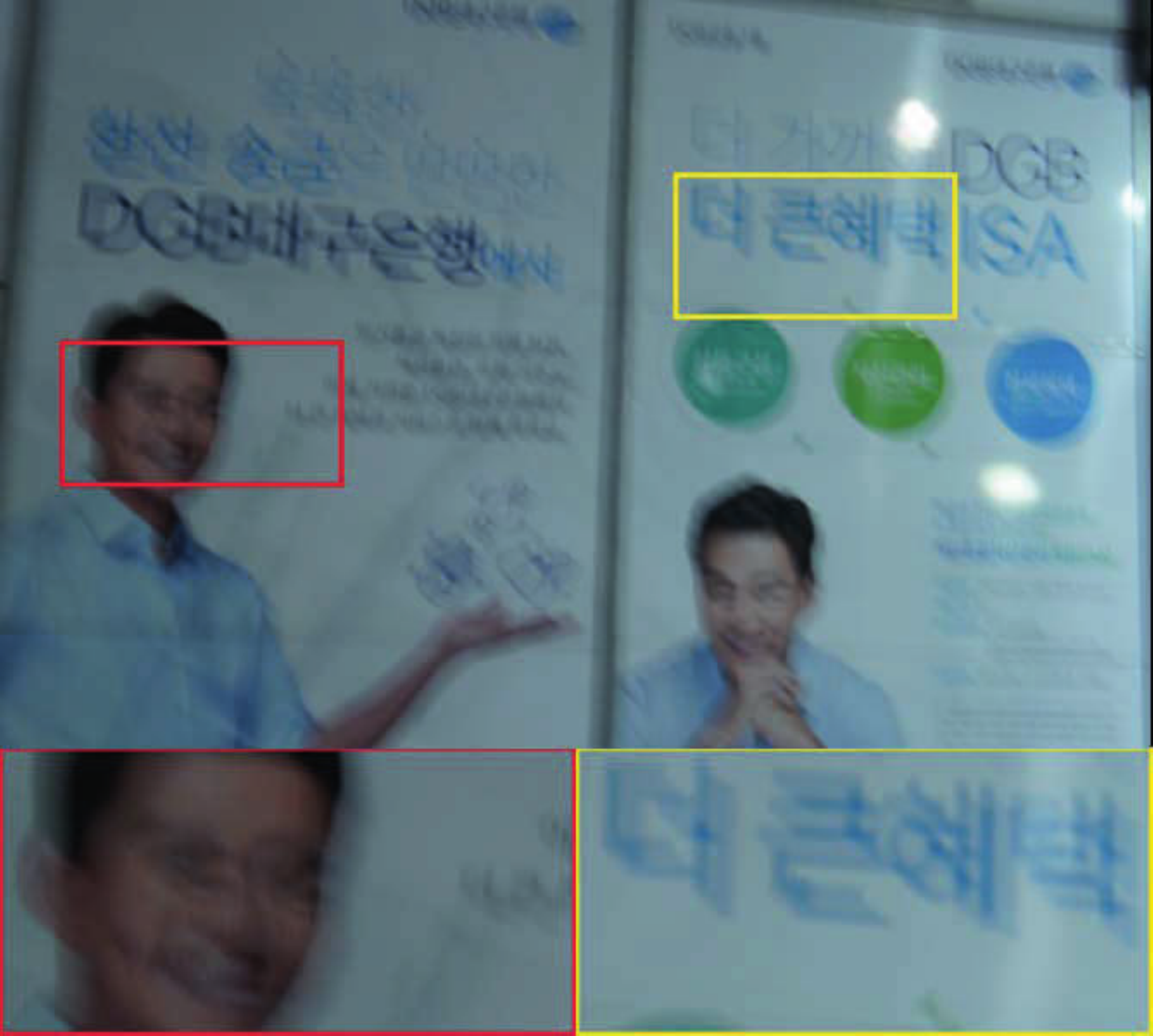}
    \end{minipage}}
    {\begin{minipage}[t]{0.24\textwidth}
    \centering\includegraphics[width=1\textwidth]{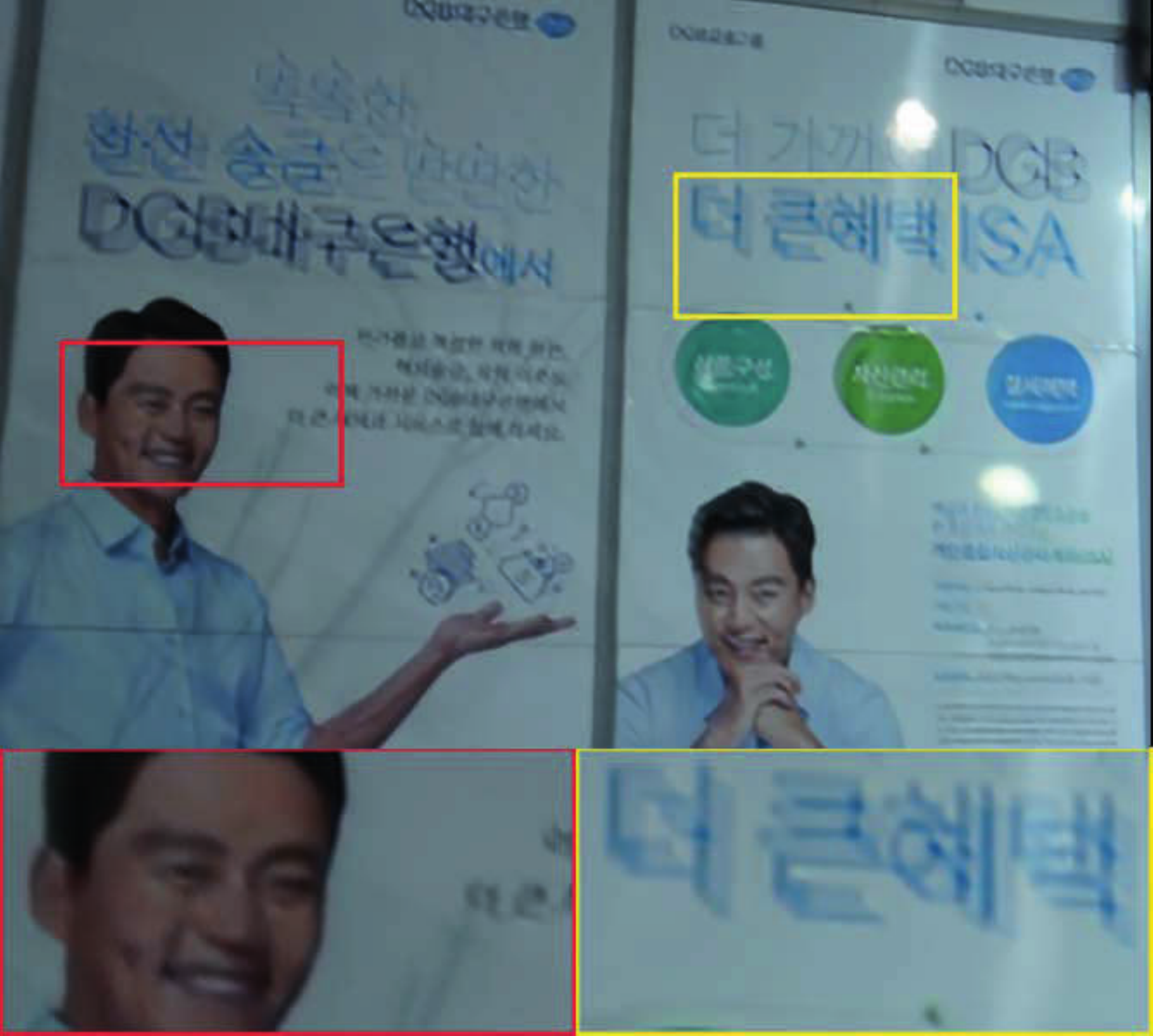}
    \end{minipage}}
    {\begin{minipage}[t]{0.24\textwidth}
    \centering\includegraphics[width=1\textwidth]{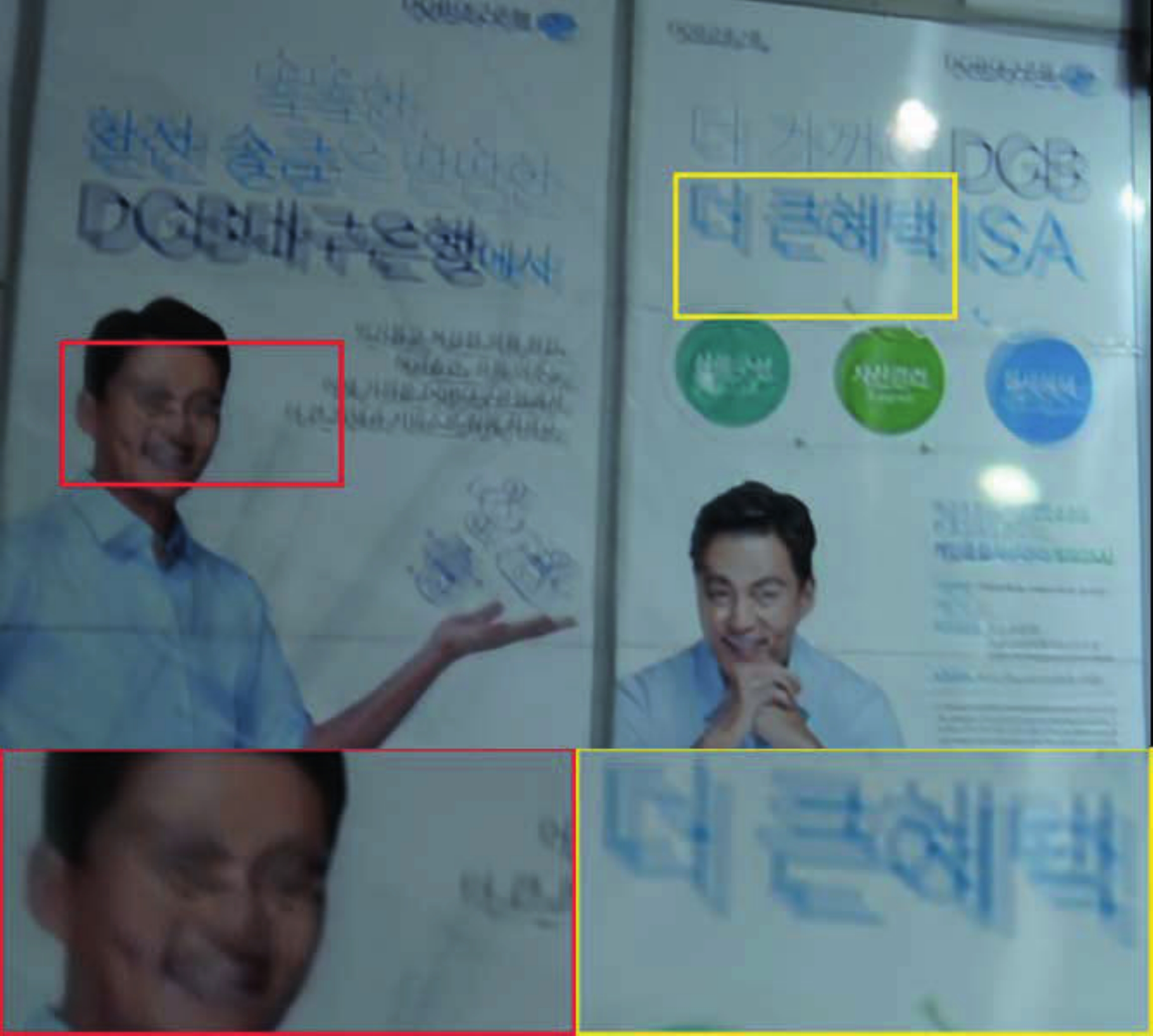}
    \end{minipage}}
    {\begin{minipage}[t]{0.24\textwidth}
    \centering\includegraphics[width=1\textwidth]{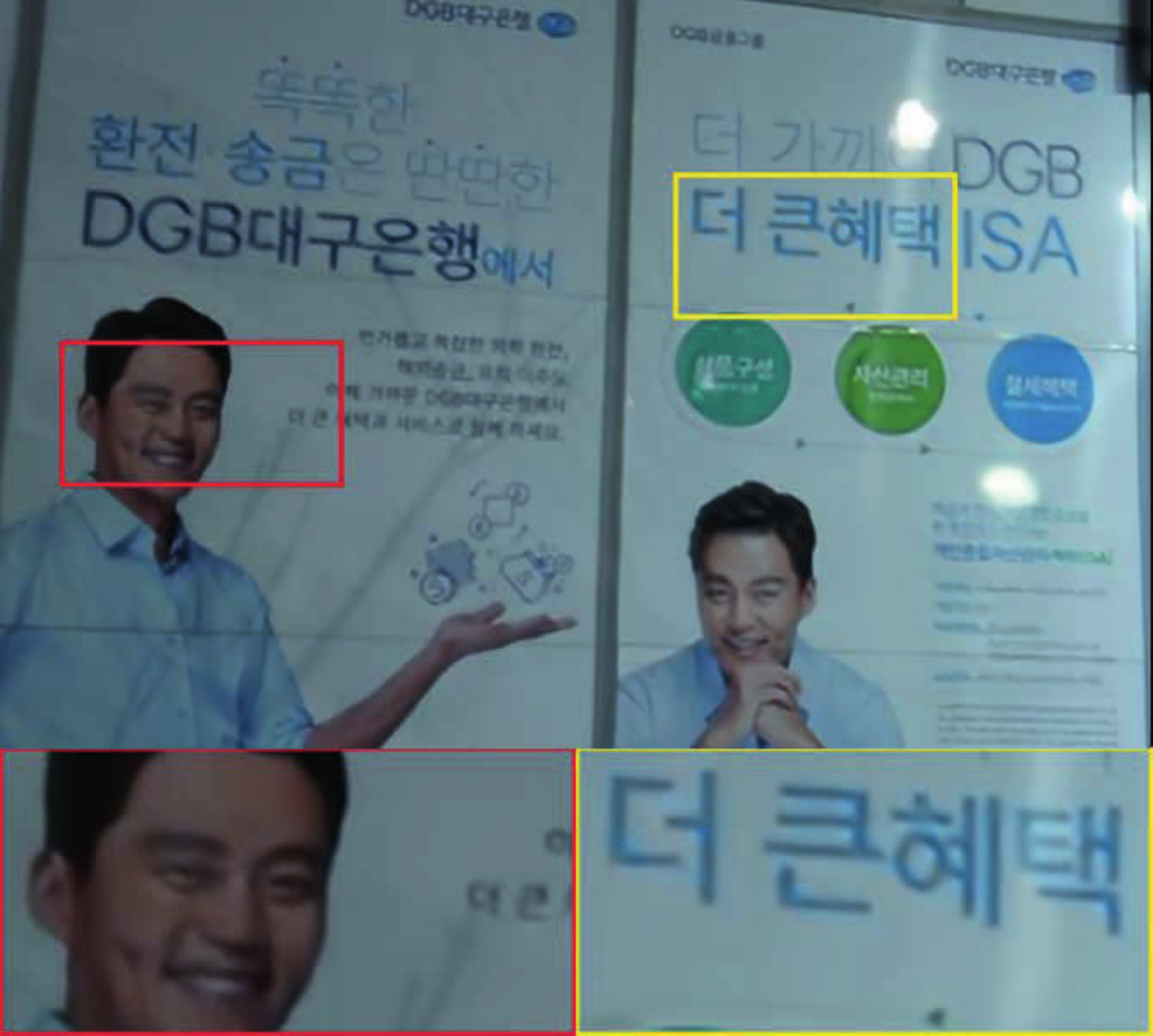}
    \end{minipage}} \\
    \vspace{-0.15cm}
    \subfigure[Blur input]{\begin{minipage}[t]{0.24\textwidth}
    \centering\includegraphics[width=1\textwidth]{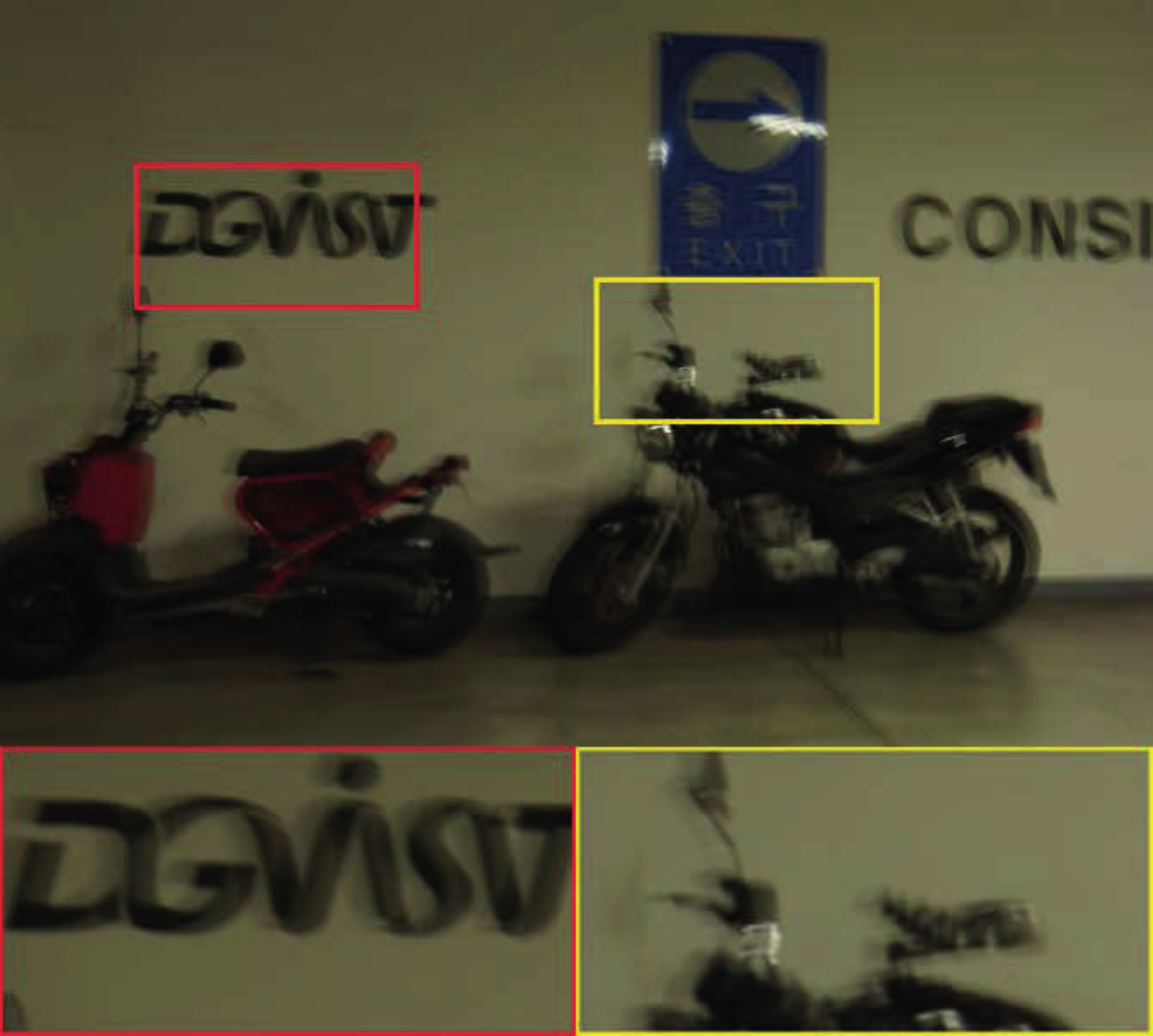}\end{minipage}}
    \subfigure[MT-RNN~\cite{park_MTRNN_ECCV_2020}]{\begin{minipage}[t]{0.24\textwidth}
    \centering\includegraphics[width=1\textwidth]{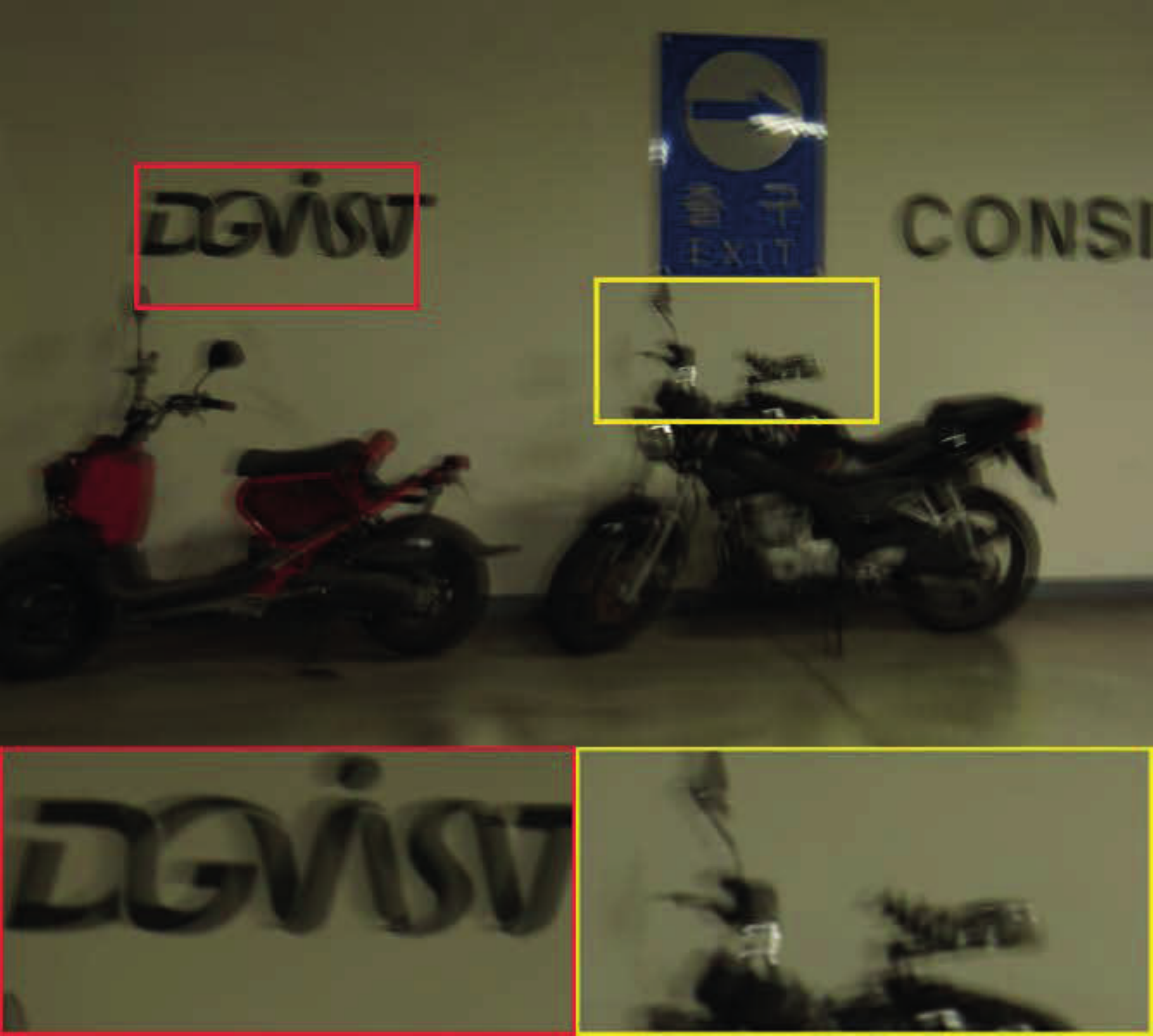}\end{minipage}}
    \subfigure[Stack(4)-DMPHN~\cite{zhang_DMPHN_CVPR_2019}]{\begin{minipage}[t]{0.24\textwidth}
    \centering\includegraphics[width=1\textwidth]{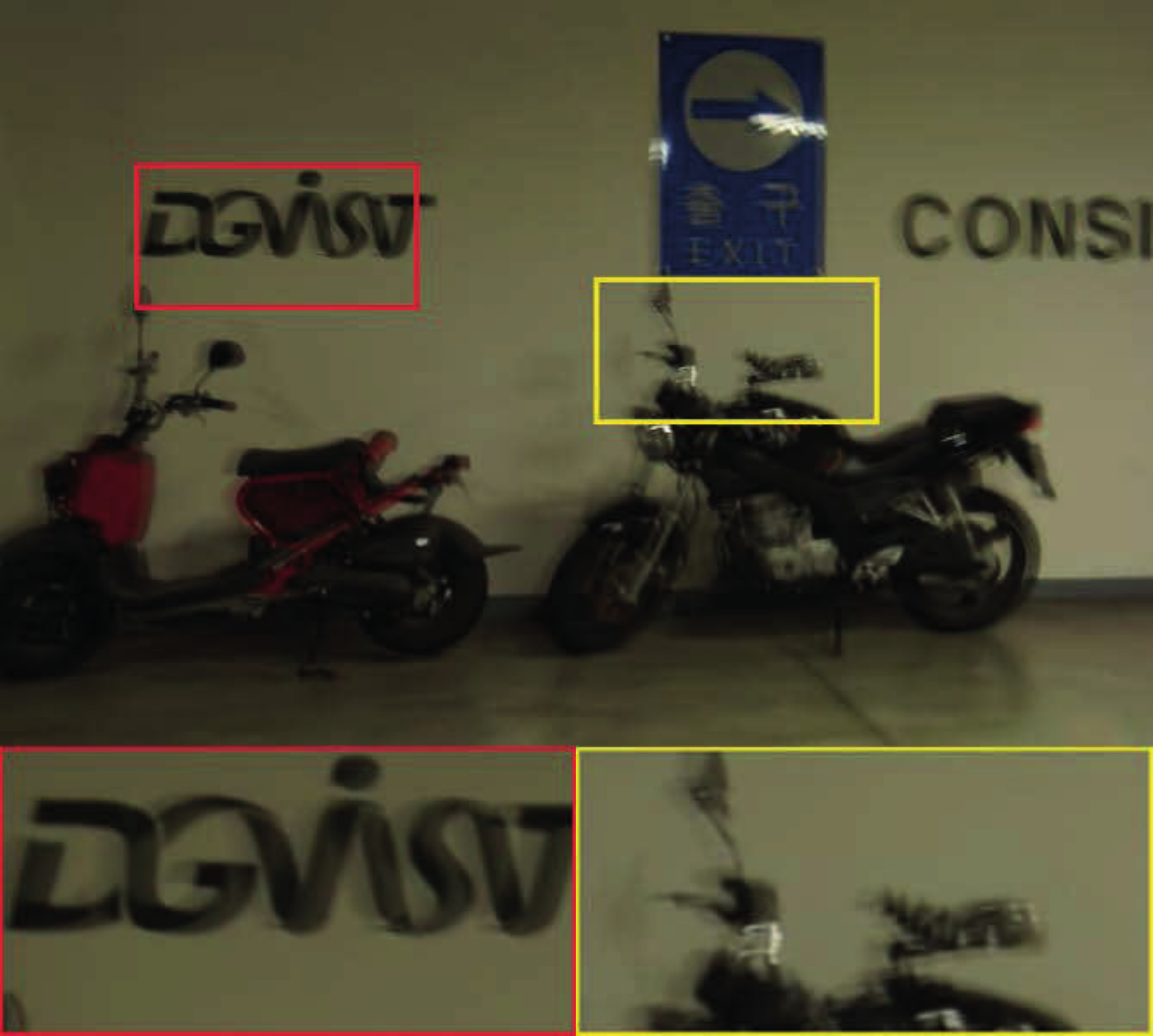}\end{minipage}}
    \subfigure[Ours]{\begin{minipage}[t]{0.24\textwidth}
    \centering\includegraphics[width=1\textwidth]{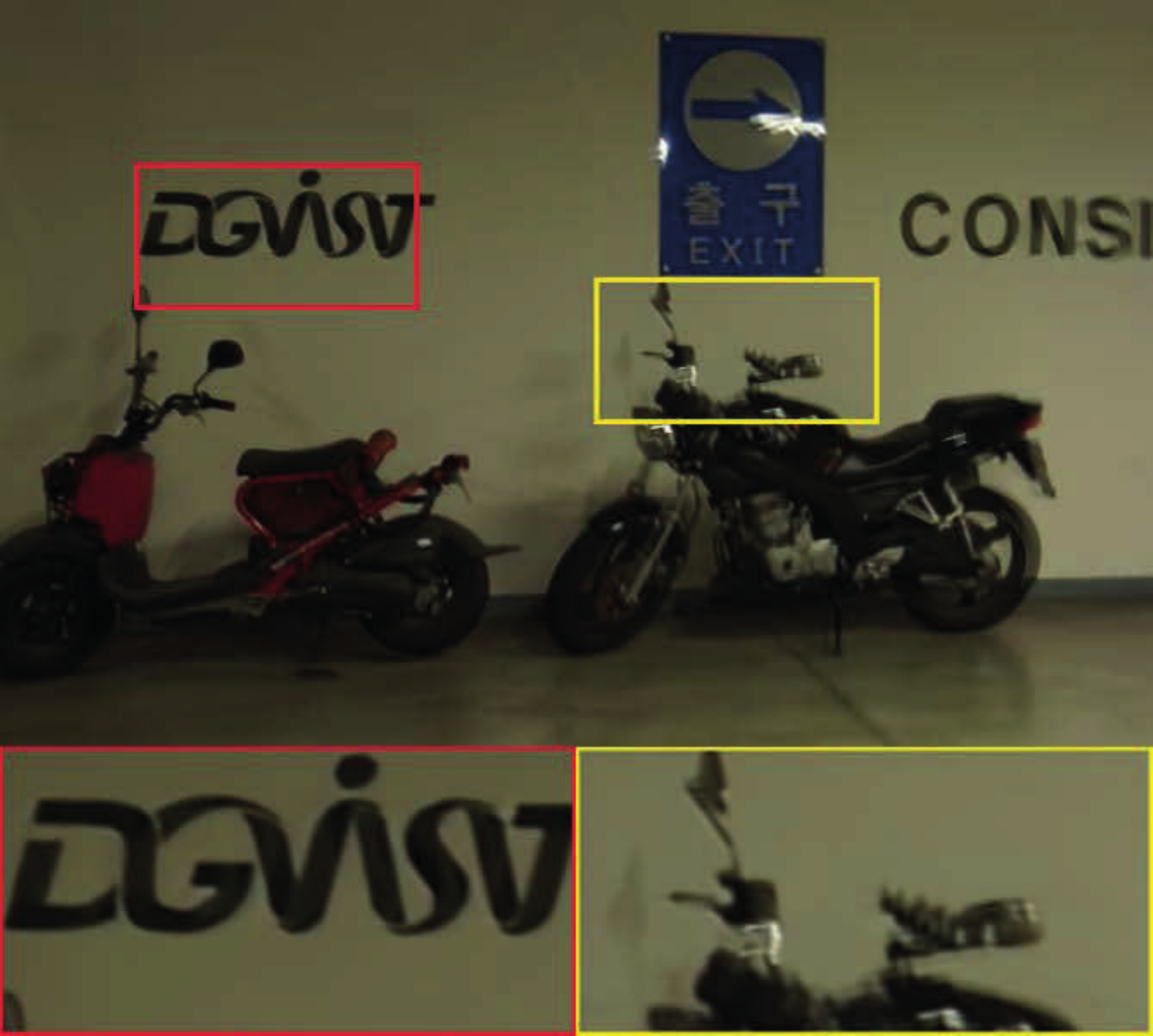}\end{minipage}} \\
    \centering
    \caption{\textbf{Visual comparisons on the RealBlur dataset.} Column (a) is the original blurry images, (b), (c), (d) are the deblurring results from MT-RNN~\cite{park_MTRNN_ECCV_2020}, Stack(4)-DMPHN~\cite{zhang_DMPHN_CVPR_2019} and our method, respectively. Best Viewed on Screen.}
    \label{fig:compare_results_RealBlur}
\end{figure*}

\noindent\textbf{Qualitative Evaluations.}
We perform the visual quality comparison of deblurred images by our proposed model and recent CNN-based dynamic scene deblurring networks, including MT-RNN~\cite{park_MTRNN_ECCV_2020} and Stack(4)-DMPHN~\cite{zhang_DMPHN_CVPR_2019} (Considering the space, we only tested the best of the two open-source methods). Fig.~\ref{fig:compare_results_GoPro} shows several blurry images from the GoPro test dataset and their corresponding deblurring results produced by the above methods. We can observe that although the above methods can play a good deblurring effect, the handling of some details such as blurred structure recovery and blurred edges is not satisfactory. For example, on the first row in Fig.~\ref{fig:compare_results_GoPro}, our proposed model can better handle highly blurred scenes, especially in the zoom-in region (Such as recovering the structure of the ``window'' is better than Stack(4)-DMPHN). And on the second row in Fig.~\ref{fig:compare_results_GoPro}, our proposed model can also perform better on recovering the blurred edges caused by the large depth of field and highly dynamic moving objects.

We compare the qualitative results on the RealBlur test datasets, as shown in Fig.~\ref{fig:compare_results_RealBlur}. We can observe that our proposed model has a very outstanding advantage for deblurring the text in the scene accompanied by uneven lighting, while there are still noticeable artifacts for the results of MT-RNN and Stack(4)-DMPHN. Moreover, our model performs better for deblurring faces (Fig.~\ref{fig:compare_results_RealBlur} second row) and tiny objects with intricate details (Fig.~\ref{fig:compare_results_RealBlur} third row).

\begin{figure*}[htbp]
    \centering
\setlength{\abovecaptionskip}{0.cm}
    \subfigure[Blur input]{
    \begin{minipage}[t]{0.44\textwidth}
        \centering
        \includegraphics[width=\textwidth]{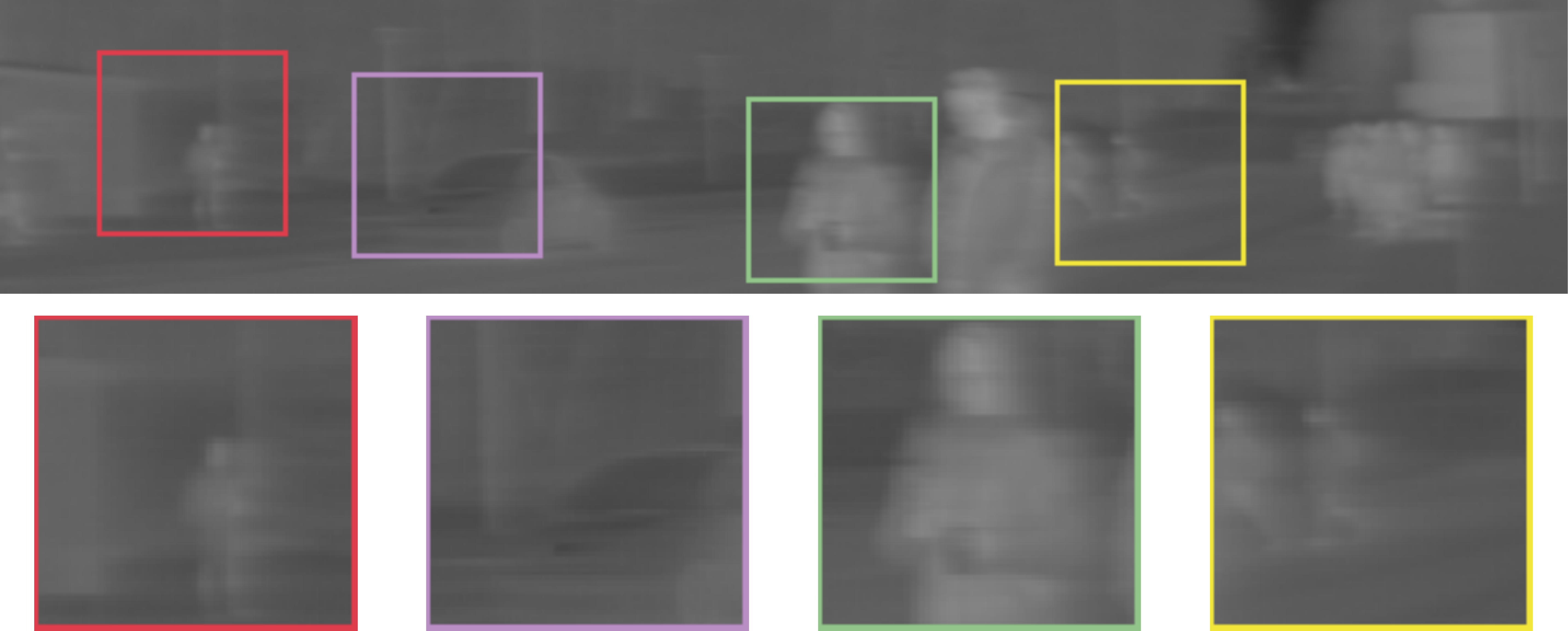}
    \end{minipage}
    }
    \subfigure[MT-RNN~\cite{park_MTRNN_ECCV_2020}]{
    \begin{minipage}[t]{0.44\textwidth}
        \centering
        \includegraphics[width=\textwidth]{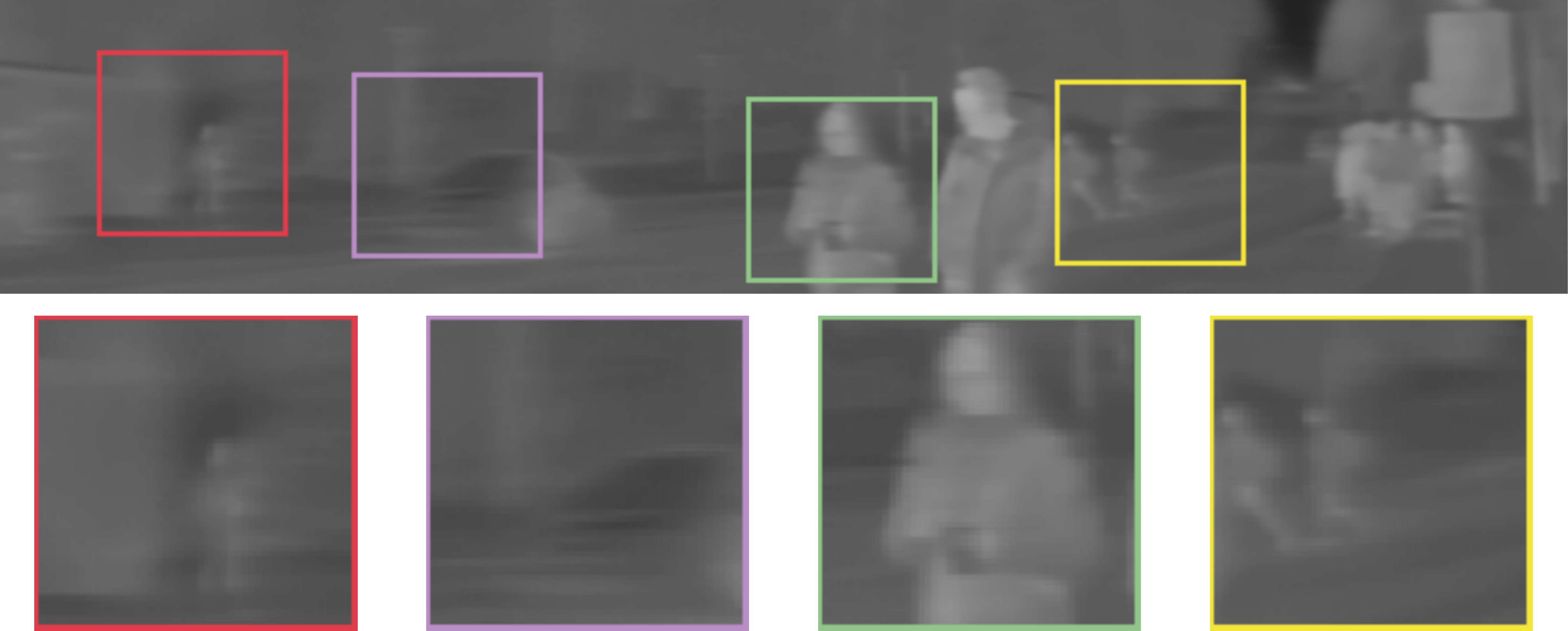}
    \end{minipage}
    } \\

    \subfigure[Stack(4)-DMPHN~\cite{zhang_DMPHN_CVPR_2019}]{
    \begin{minipage}[t]{0.44\textwidth}
        \centering
        \includegraphics[width=\textwidth]{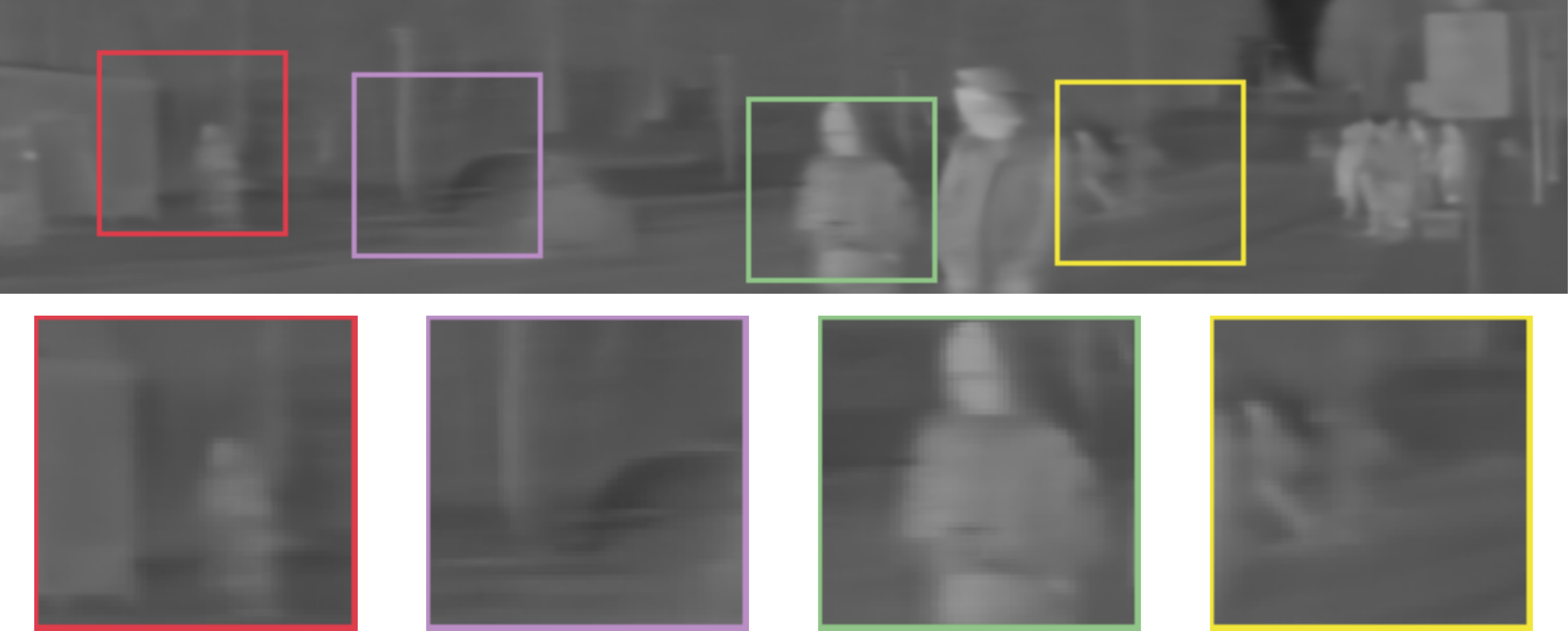}
    \end{minipage}
    }
    \subfigure[Ours]{
    \begin{minipage}[t]{0.44\textwidth}
        \centering
        \includegraphics[width=\textwidth]{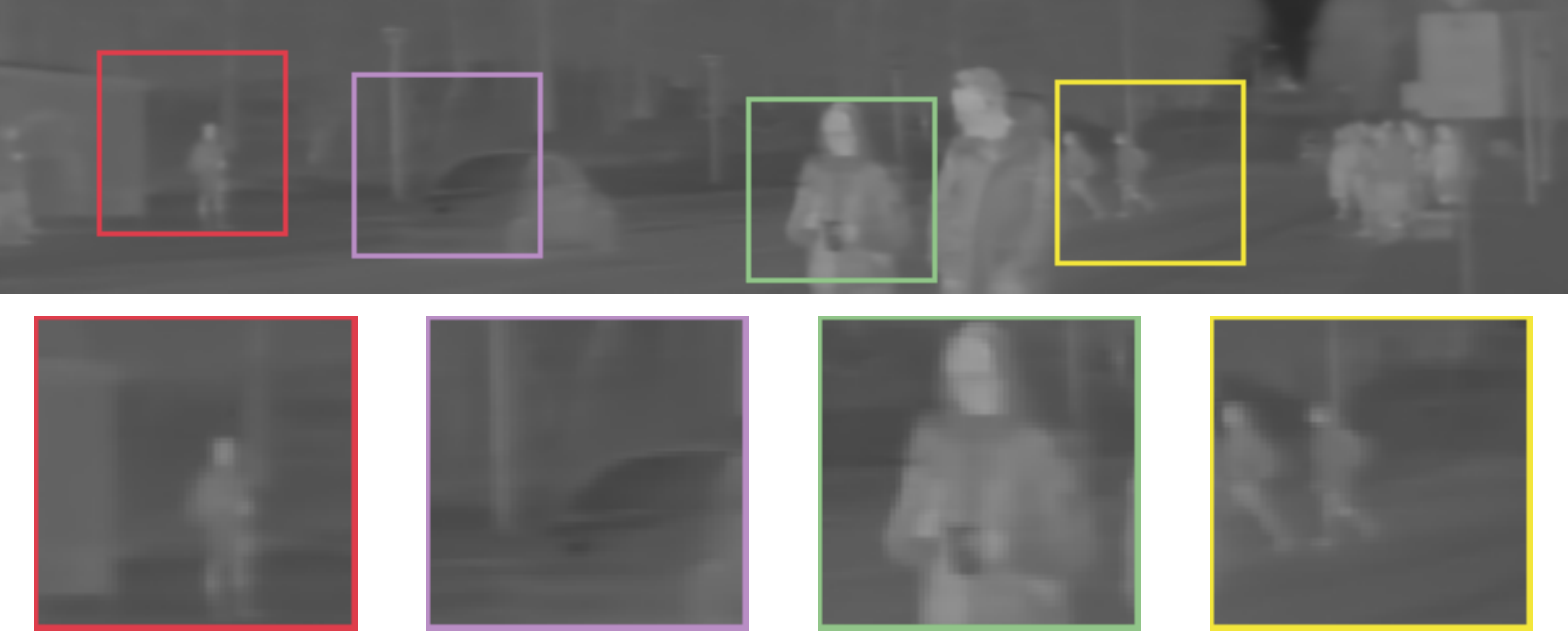}
    \end{minipage}
    }

    \centering

    \caption{\Fix{\textbf{Visual comparisons on the motion blurred thermal samples from ~\cite{Ramanagopal_Thermal_deblur_RSS_20}.} Column (a) is the original blurred thermal images, (b), (c), (d) are the deblurring results from MT-RNN~\cite{park_MTRNN_ECCV_2020}, Stack(4)-DMPHN~\cite{zhang_DMPHN_CVPR_2019} and our method, respectively. Best Viewed on Screen.}}
    \label{fig:compare_results_thermal}
\end{figure*}

\Fix{Our proposed idempotent deblurring network can be applied to other imaging modalities. Following~\cite{Ramanagopal_Thermal_deblur_RSS_20}, we perform a qualitative analysis of our pre-trained model and MT-RNN, Stack(4)-DMPHN on motion blurred thermal images on the dataset of~\cite{Ramanagopal_Thermal_deblur_RSS_20}. The visualization comparisons are shown in Fig.~\ref{fig:compare_results_thermal}. We can observe that our model recovers sharper details from the blurred thermal inputs, especially for highly dynamic moving objects and structural details of the image. The experiments on the thermal images show that our proposed deblurring framework can generalize well to other image modalities. }


\subsection{Ablation studies}

\begin{table}[tb!]
    \caption{\textbf{Quantitative analysis of ablation studies}.}
    \label{tab:AblationStudy}
    \centering
    \resizebox{0.48\textwidth}{!}{
        \begin{threeparttable}
        \begin{tabular}{cccccc|cc}
            \toprule
                                       & FMR   & LCR    & Times & Idem.    & Iters  & PSNR~(dB)      & SSIM \cr
            \midrule
            (\textit{a})                & -     & -     & 1       & -   & 1      & 29.463 & 0.9259  \cr
            (\textit{b})                & -     & -     & 1       & -   & 2      & 29.958 & 0.9323  \cr
            (\textit{c})                & -     & -     & 1       & -   & 4      & 30.933 & 0.9434  \cr
            (\textit{d})                & -     & -     & 1       & -   & 6      & 31.337 & 0.9461  \cr
            (\textit{e})                & -     & -     & 1       & -   & 8      & 31.464 & 0.9468  \cr
            \midrule
            (\textit{f})                & \checkmark    & -     & 1       & -   & 6      & 31.479 & 0.9468 \cr
            (\textit{g})                & \checkmark    & -     & 2       & -   & 6      & 31.403 & 0.9465 \cr
            (\textit{h})                & \checkmark    & -     & 2       & \checkmark & 6      & 31.521 & 0.9469 \cr
            \midrule
            (\textit{i})& \checkmark    & \checkmark   & 1  & -   & 6  & 31.796 & 0.9487    \cr
            (\textit{j})& \checkmark    & \checkmark   & 2  & -          & 6  & 31.684 & 0.9471   \cr
            \midrule
            \Fix{(\textit{k})}& \Fix{\checkmark}& \Fix{\checkmark}   & \Fix{3}  & \Fix{\checkmark} & \Fix{6}  & \Fix{31.892} & \Fix{0.9523} \cr
            \midrule
            (\textit{l})& \checkmark    & \checkmark   & 2  & \checkmark & 6  & 31.917   & 0.9527  \cr
            (\textit{m})& \checkmark    & \checkmark   & 2  & \checkmark & 8  & 31.972  & 0.9529  \cr
            \bottomrule
        \end{tabular}
        \end{threeparttable}
    }
\end{table}

In this section, we perform ablation studies on the GoPro test dataset to analyze the effectiveness of each component of our proposed method. The relevant experimental results are reported in Table~\ref{tab:AblationStudy}. \textit{FMR} represents whether using feature map recurrence. \textit{LCR} denotes whether to use latent code recurrence to embed states during iterations. \textit{Times} means \Fix{the deblurring times in training, including deblurring and re-deblurring}. \textit{Idem.} means whether using an idempotent constraint on the repeating re-deblurring outputs. \textit{Iters} is the total number of iterations $N$ in a single deblurring process.

\noindent\textbf{Effectiveness of the Idempotent Constraint.}
We first conduct two experiments to evaluate the effectiveness of the idempotent constraint. As shown in Table~\ref{tab:AblationStudy} (\textit{i}) and (\textit{l}), after using the idempotent constraint during the training process, the deblurring performance can be improved from 31.796dB to 31.917dB in terms of PSNR. Real-world deblurring performance on the RealBlur dataset reported in Table~\ref{tab:RealBlur_benchmark} also demonstrates the effectiveness of our idempotent constraint. 
If we only perform re-deblur and use sharp loss $\mathcal{L}_{Sharp}$ to supervise without idempotent loss $\mathcal{L}_{Idem}$ (\cf Table~\ref{tab:AblationStudy} (\textit{j})), the result will be severely reduced. We analyze that directly repeating the whole deblurring model will cause a bottleneck in the information flow, which explains the inferior result of (\textit{j}) to (\textit{i}), and demonstrates the effectiveness of our proposed idempotent constraint. 
\Fix{We experiment with deblurring 3 times with the idempotent constraint. The results of (\textit{k}) and (\textit{j}) show a significant performance improvement. This comparison demonstrates the effectiveness of the idempotent constraint in deblurring multiple times. However, as the performance of (\textit{k}) is comparable to (\textit{l}), considering the faster training speed, we set the deblurring times of our idempotent framework to 2.}

\begin{figure*}[ht]
    \centering
    \subfigure[MT-RNN~\cite{park_MTRNN_ECCV_2020}]{\begin{minipage}[t]{0.24\textwidth}
    \centering\includegraphics[width=1\textwidth]{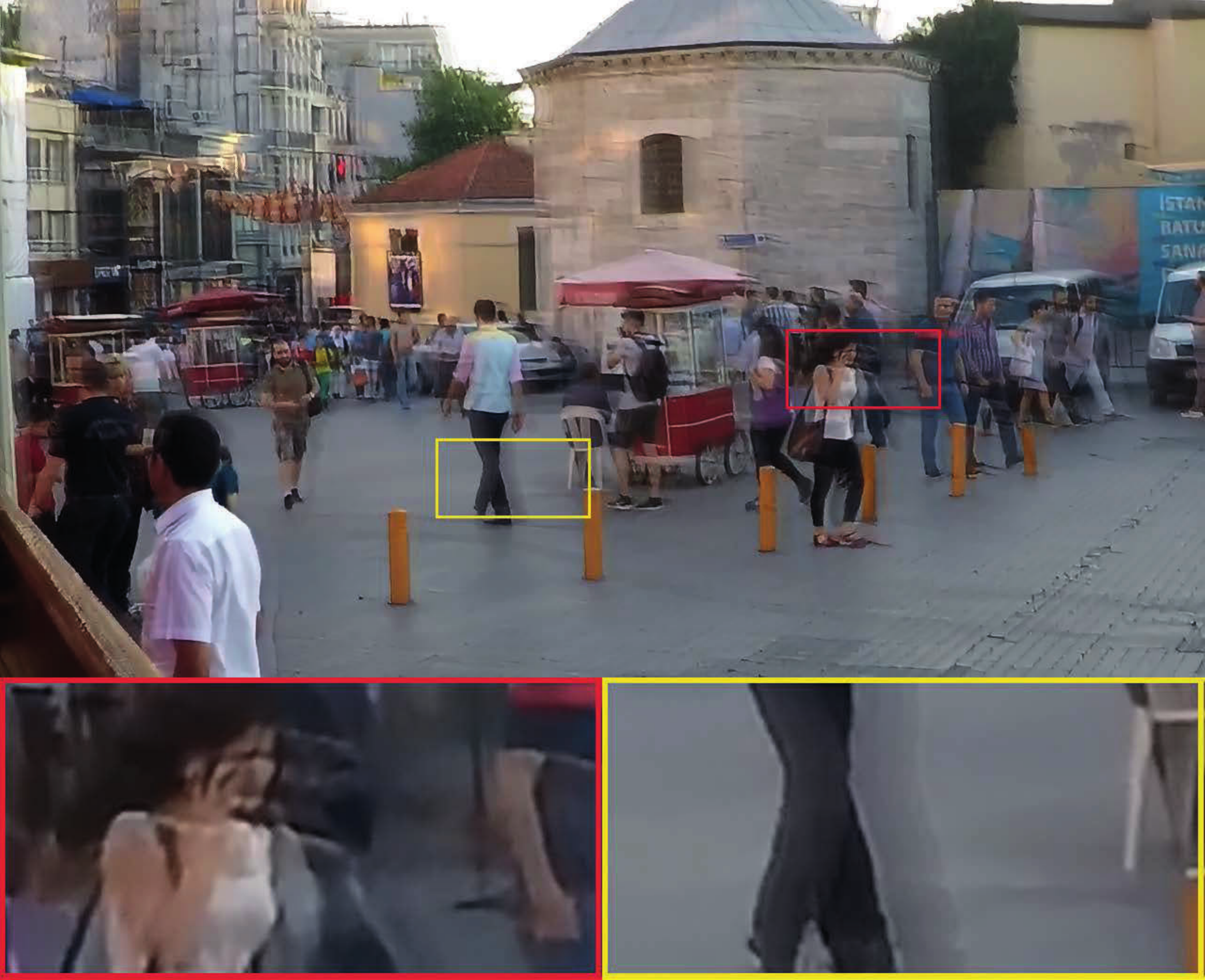}\end{minipage}}
    \subfigure[Stack(4)-DMPHN~\cite{zhang_DMPHN_CVPR_2019}]{\begin{minipage}[t]{0.24\textwidth}
    \centering\includegraphics[width=1\textwidth]{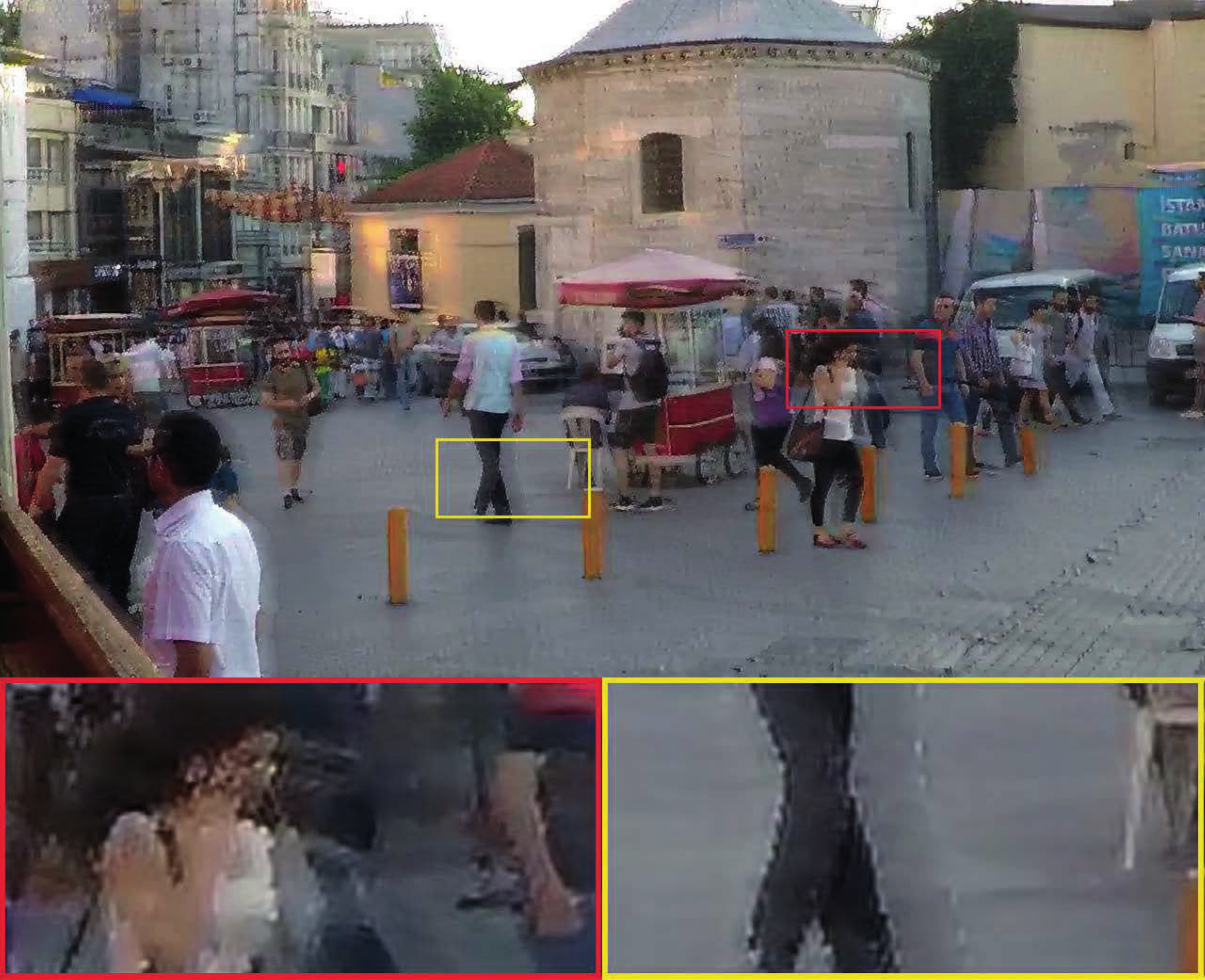}\end{minipage}}
    \subfigure[Ours]{\begin{minipage}[t]{0.24\textwidth}
    \centering\includegraphics[width=1\textwidth]{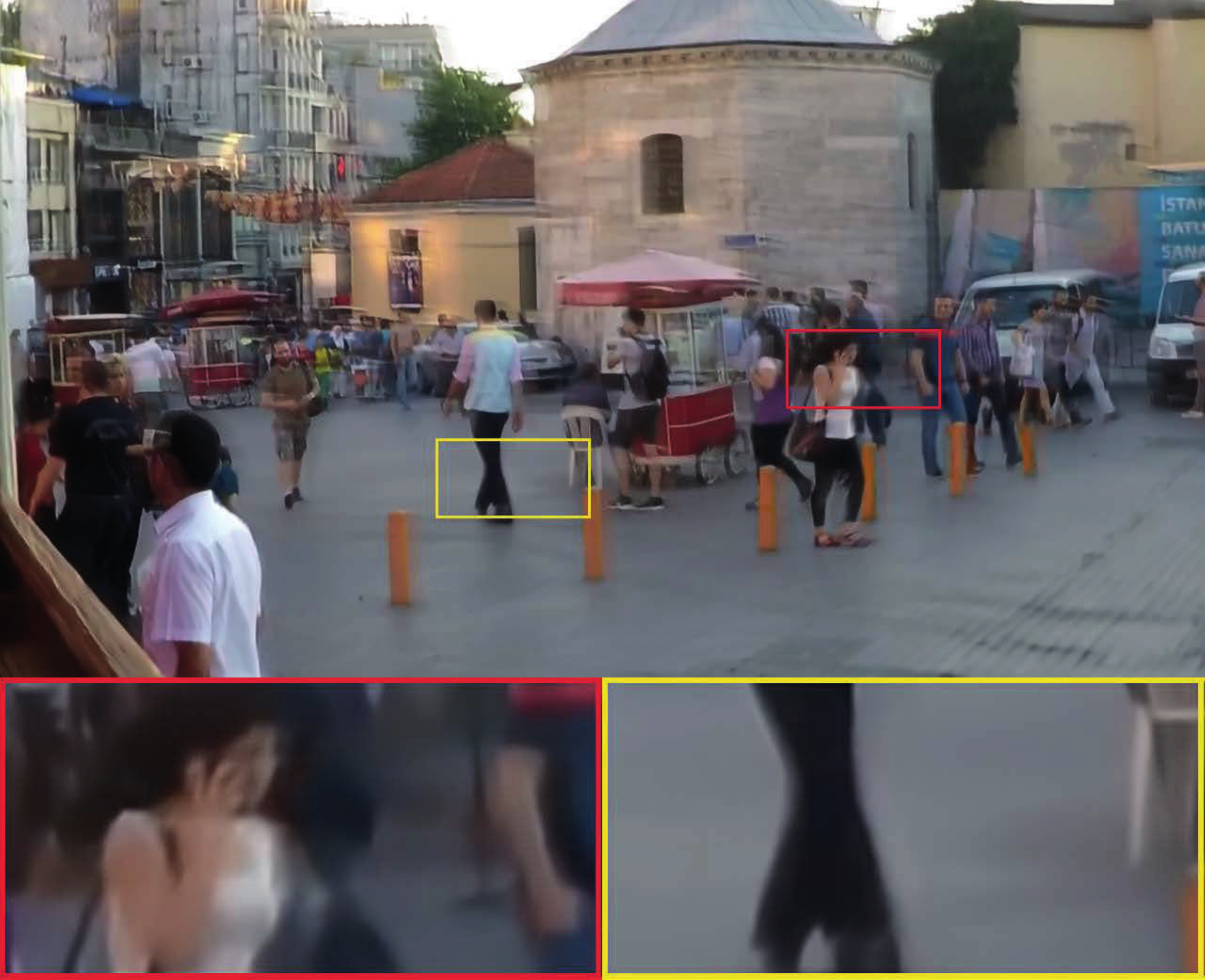}\end{minipage}}
    \subfigure[Ground-Truth]{\begin{minipage}[t]{0.24\textwidth}
    \centering\includegraphics[width=1\textwidth]{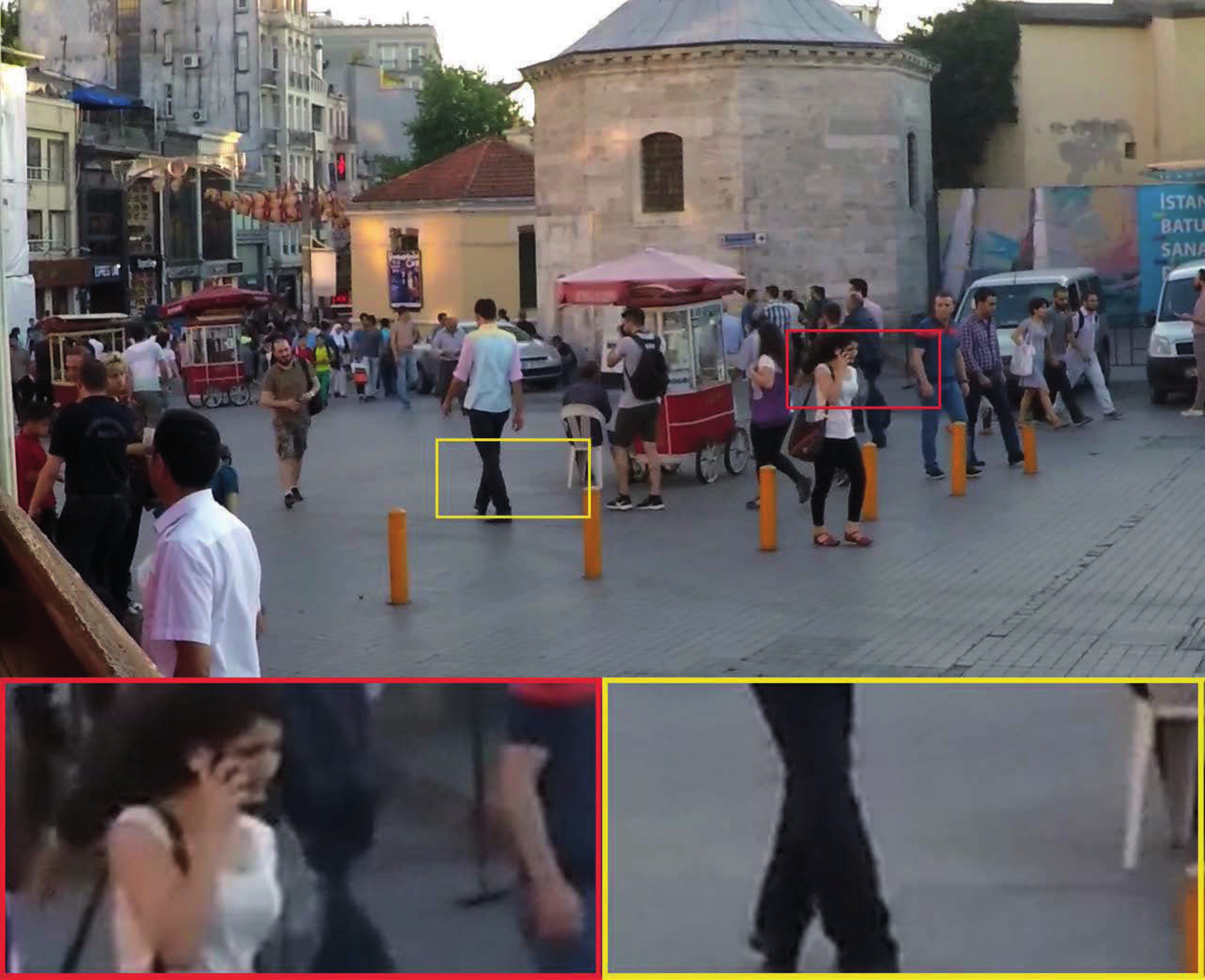}\end{minipage}} 
    \quad
    \centering
    \caption{\textbf{Visual comparisons of model idempotence}. Columns 1 to 3 show the deblurred results on the GoPro dataset when re-deblurring 10 times, and column 4 is the ground-truth sharp image for visual comparison.}
    \label{fig:idempotence_results}
\end{figure*}

In Fig.~\ref{fig:idem_blur}, compared with others, our model trained with the idempotent constraint could retain stable performance when re-deblurring multiple times. 
We compare our model with MT-RNN~\cite{park_MTRNN_ECCV_2020} and Stack(4)-DMPHN~\cite{zhang_DMPHN_CVPR_2019} by repeating the re-deblurring process for 10 times. As shown in Fig.~\ref{fig:idempotence_results}, after repeating 10 times, more noise appears in the outputs of Stack(4)-DMPHN while the result of MT-RNN shows overly smoothed and unrealistic effects. In contrast, our model with the idempotent constraint produces a visually realistic result that is close to the sharp ground-truth image.
The re-deblurring results are repeated 10 times just to highlight the degradation trend of these methods. We only need to repeat twice in training, and for real application, the deblurring algorithm only needs to be done once \ie no re-deblurring. Benefiting from the idempotent constraint, our model maintains stable idempotence for multiple re-deblurring results. 

\noindent\textbf{Effectiveness of Progressive Residual Deblurring.}
By training the model with progressive residual deblurring, we allow the network to consider wider image contexts and gradually restore the sharp image. 
In Table~\ref{tab:AblationStudy},  we experiment on the influence of iterations when training the progressively deblurring model. As shown in Line (\textit{a})-(\textit{e}), our model achieves improved performance as the iteration number increase, which validates the effectiveness of the progressive residual learning mechanism. The results in Line (\textit{k}) and (\textit{l}) for iterations 6 and 8 also demonstrate this.
\Fix{The iteration number is a hyperparameter that we can manually choose to achieve either better deblurring performance or faster inference. Considering the trade-off between inference time and deblurring performance, we set the iteration number to 6.}
    
\noindent\textbf{Effectiveness of the Feature Maps Recurrence.}
We observe that results become better with the help of the feature maps recurrence structure in Table~\ref{tab:AblationStudy} (\textit{d}) and (\textit{f}).
Because it inherently passes the high-frequency features from the decoder to the next encoder, the high-frequency features will be further emphasized. The highlighted high-frequency features thus lead to better deblurring performance.


\begin{figure*}[ht]
    \centering
    \includegraphics[width=1\textwidth]{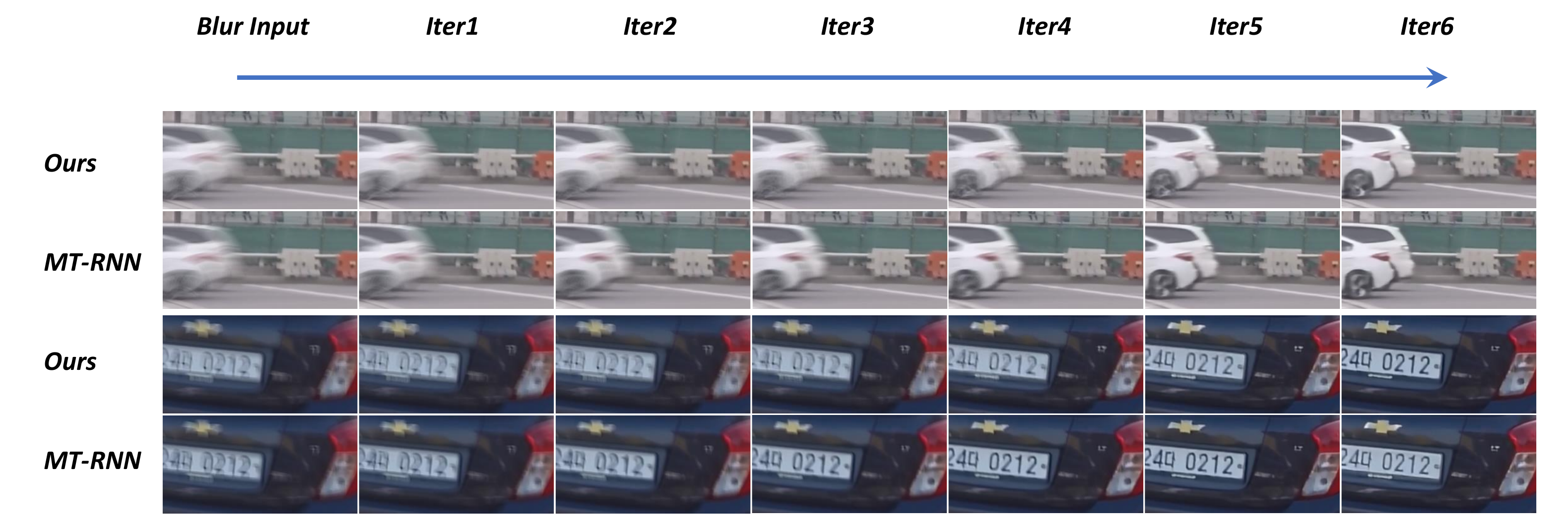}
    \caption{\textbf{Progressively deblurred results of our method and MT-RNN~\cite{park_MTRNN_ECCV_2020}.} Same as MTRNN, our model can achieve progressive iterative deblurring over multiple iterations, but without their temporal data augmentation in the training process.}
    \label{fig:fig_progress}
\end{figure*}

\begin{table}[t]
    \caption{\textbf{Quantitative results of retrained DMPHN and MT-RNN with our idempotent constraint}.}
    \label{tab:IdemInOthers}
    \centering
    \resizebox{0.48\textwidth}{!}{
        \begin{threeparttable}
        \begin{tabular}{ccccc}
            \toprule
             Methods  & DMPHN & DMPHN(w/ Idem) & MT-RNN & MT-RNN(w/ Idem)  \cr
            \midrule
            PSNR~(dB)  & 30.25    & 30.36         & 31.15   & 31.28  \cr
            SSIM       & 0.935    & 0.936         & 0.945   & 0.945  \cr
            \toprule
        \end{tabular}
        \end{threeparttable}
    }
\end{table}

\begin{table}[ht]
    \caption{
    \Fixvv{\textbf{Quantitative analysis of re-deblurring performance.} The first two columns represent using different methods to deblur the original blurry images from the GoPro dataset. The remaining three columns represent each method's deblurring results when re-deblurring the deblurred images by the first column's methods.}
    }
    \label{tab:redeblurring}
    \centering
    \resizebox{0.48\textwidth}{!}{
    \begin{tabular}{cc|ccc}
        \toprule
        Deblur Method & Deblur & Re-deblur Method & 1st Re-deblur & 2nd Re-deblur  \\
        \midrule
        Ours & 31.917 & Ours & 31.916 & 31.914   \\
        \midrule
        \multirow{2}*{Stack(4)-DMPHN~\cite{zhang_DMPHN_CVPR_2019}} &\multirow{2}*{31.402} & Stack(4)-DMPHN~\cite{zhang_DMPHN_CVPR_2019} & 30.192 & 29.896 \\
        & & Ours & 31.370 & 31.369   \\
        \midrule
        \multirow{2}*{MTRNN~\cite{park_MTRNN_ECCV_2020}} & \multirow{2}*{31.149} & MTRNN~\cite{park_MTRNN_ECCV_2020} & 30.538 & 30.322 \\
         & & Ours & 31.041 & 31.039\\
        \bottomrule
    \end{tabular}
    }
\end{table}

\noindent\textbf{Effectiveness of the Latent Code Recurrence.}
In Table~\ref{tab:AblationStudy} (\textit{h}) and (\textit{k}), the latent code recurrence improves PSNR by about 0.396dB. 
This demonstrates that our LCR indeed mitigates the problem of fewer channel numbers at the bottleneck caused by the lightweight network. The GRU module encodes blurry patterns into hidden states among multiple iterations, thus the progressive residual deblurring process can fully utilize features that are more helpful to restore the sharp image.


\subsection{Idempotent Property on Previous Methods}
To demonstrate the effectiveness of our proposed idempotent constraint, we retrain DMPHN and MT-RNN with our idempotent constraint. The evaluation results on the GoPro test dataset are reported in Table~\ref{tab:IdemInOthers}. Our proposed idempotent constraint can improve the deblurring performance of these two methods without any modification of the network.

\Fixvv{To verify the robustness of our method when dealing with deblurred images, we further perform experiments by feeding images obtained from other deblurring methods into our model. In Table~\ref{tab:redeblurring}, we report the re-deblurring performance of our model and the re-deblurring performance of Stack(4)-DMPHN and MT-RNN. 
We can observe that the re-deblurring performance of our method is significantly better than Stack(4)-DMPHN and MT-RNN. This demonstrates that our proposed idempotent framework can effectively deal with images acquired from arbitrary sources (\ien, original blurry images and deblurred images from other deblurring methods).}

\begin{table*}[ht!]
    \caption{\textbf{Quantitative analysis of iterative residual deblurring compared with MT-RNN~\cite{park_MTRNN_ECCV_2020}}.}
    \label{tab:iterative_results}
    \centering
    \begin{threeparttable}
    \begin{tabular}{ccccccccc}
    \toprule
    Method     &   Deblurring Times    &    Calculation Data      &     1    & 2    & 3    & 4    & 5     & 6    \\
    \midrule
    \multirow{6}{*}{Ours} 
    & \multirow{3}{*}{1} 
    & PSNR~(dB) & 25.475 & 25.556 & 26.347 & 28.191 & 30.975 & 31.917 \\
    & & Every  & 3.614  & 4.992  & 3.834  & 3.123  & 4.302  & 3.903  \\
    & & Sum    & 3.614  & 6.235  & 5.492  & 8.329  & 11.674 & 13.026 \\
    \cmidrule(lr){2-9}
    & \multirow{3}{*}{2}
    & PSNR~(dB) & 31.903 & 31.903 & 31.897 & 31.895 & 31.895 & 31.973 \\
    & & Every  & 0.359  & 0.422  & 0.142  & 0.381  & 0.675  & 0.673 \\
    & & Sum    & 0.359  & 0.298  & 0.540  & 0.531  & 0.359  & 0.071 \\
    \midrule
    \multirow{6}{*}{MT-RNN~\cite{park_MTRNN_ECCV_2020}} 
    & \multirow{3}{*}{1}
    & PSNR~(dB) & 26.432 & 27.606 & 29.001 & 30.363 & 31.083 & 31.127 \\
    & & Every  & 2.324  & 3.118  & 3.283  & 3.203  & 2.572  & 1.217  \\
    & & Sum    & 2.324  & 5.102  & 7.811  & 10.226 & 11.943 & 12.469 \\
    \cmidrule(lr){2-9}
    & \multirow{3}{*}{2}
    & PSNR~(dB) & 31.093 & 31.002 & 30.884 & 30.762 & 30.637 & 30.509 \\
    & & Every  & 0.386  & 0.535  & 0.584  & 0.586  & 0.580  & 0.579  \\
    & & Sum    & 0.386  & 0.803  & 1.175  & 1.491  & 1.765  & 2.020  \\
    \toprule
    \end{tabular}
    \end{threeparttable}
\end{table*}

\subsection{Comparison of Our Method and MT-RNN}
Fig.~\ref{fig:fig_progress} shows the output of progressively deblurred results of our method and MT-RNN~\cite{park_MTRNN_ECCV_2020} in different deblurring iterations. Our model archives progressively deblurring similar to MT-RNN, but do not need their temporal data augmentation. The degradation curve of MT-RNN in Fig.~\ref{fig:idem_blur} is relatively more stable than other methods without our idempotent constraint (except for Ours (w/Idem.)). We believe this is mainly because they train the model using additional temporally augmented data. This data is composed of 3 to 13 (odd numbers) frames respectively, which is more than the commonly used split by GoPro~\cite{nah_DeepDeblur_CVPR_2017}. The results of our re-trained MT-RNN without their data augmentation also prove this belief.

We report the quantitative results of iterative residual deblurring from every deblurring unit on the GoPro test dataset~\cite{nah_DeepDeblur_CVPR_2017} in Table~\ref{tab:iterative_results}. 
The first deblurring (Deblurring Times 1) uses the blurry image as the model input, and the second deblurring (Deblurring Times 2) uses the deblurred image at the first time as the input.
\textit{PSNR} represents the distortion between the deblurred image at each iteration and the ground-truth sharp image. 
\textit{Every} is the absolute value of the residual image output by the network in each iteration. 
\textit{Sum} is the absolute value of the sum of the residual images from the current and previous iterations, which is equivalent to the difference between the current deblurred image and the input image of the network at the first iteration. 
Note that these absolute values are defined in the image value range from [0, 255].

For the first deblurring time, the deblurring performance improves with the increase of the number of iterations. Although the trend is similar, the variation (Every and Sum) of each iteration is quite different. In the process of the second deblurring (\ien, re-deblurring), our performance is more stable and the PSNR of the final output is better.

In particular, the value of each residual is very small, and the sum of the residual is very close to zero. As mentioned in Section~\ref{section:idemrepetredeblur}, the idempotent constraint can be satisfied when the sum of all the residual outputs is close to zero in the re-deblurring process. On the contrary, the Sum value of MT-RNN~\cite{park_MTRNN_ECCV_2020} increases with multiple iterations step by step. This also leads to its worse and unstable performance in the second deblurring process and shows the effectiveness of our \emph{idempotent constraint} by comparison.

\begin{figure*}[htbp]
    \centering
    {\begin{minipage}[t]{0.24\textwidth}
    \centering\includegraphics[width=1\textwidth]{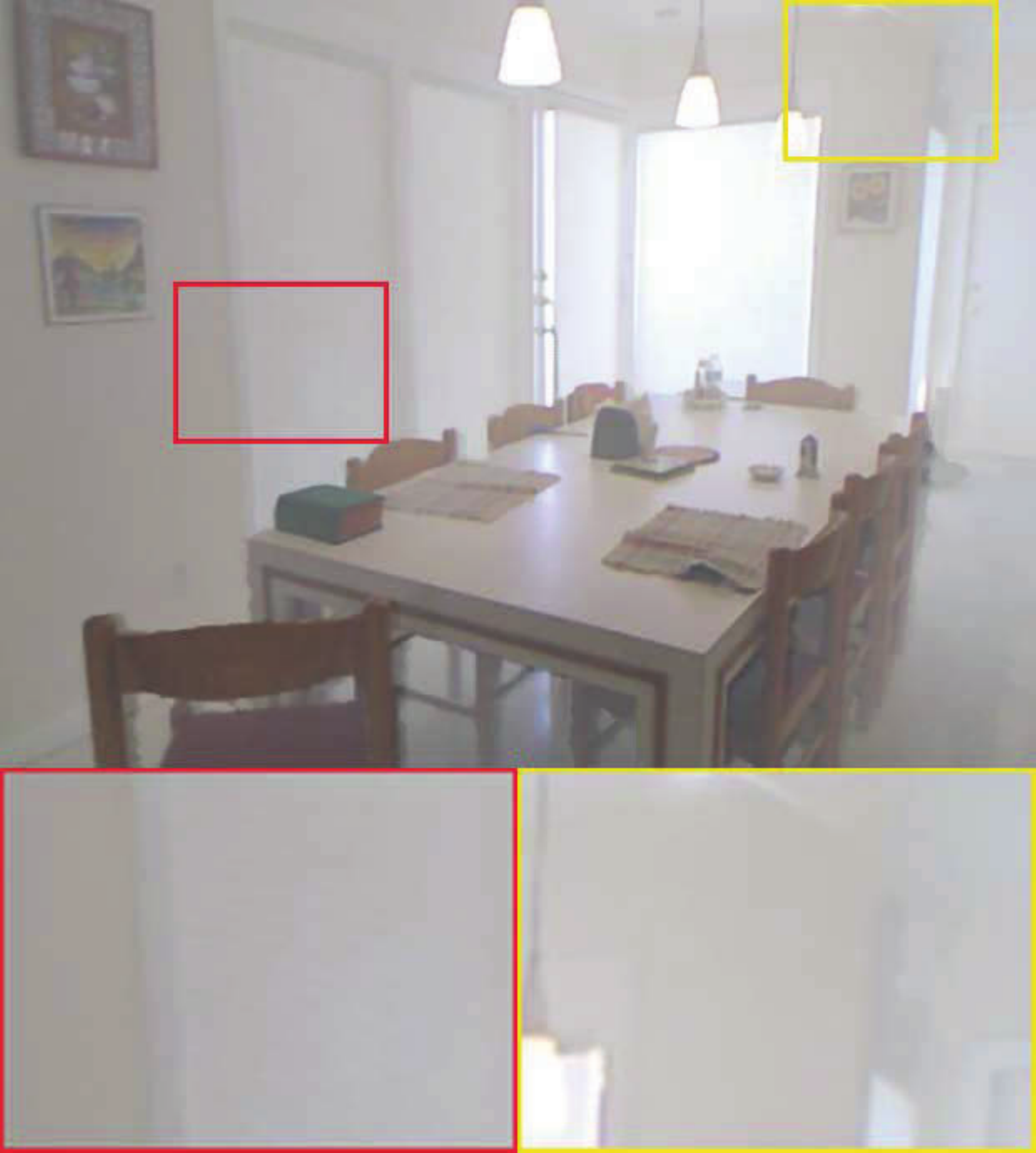}
    \end{minipage}}
    {\begin{minipage}[t]{0.24\textwidth}
    \centering\includegraphics[width=1\textwidth]{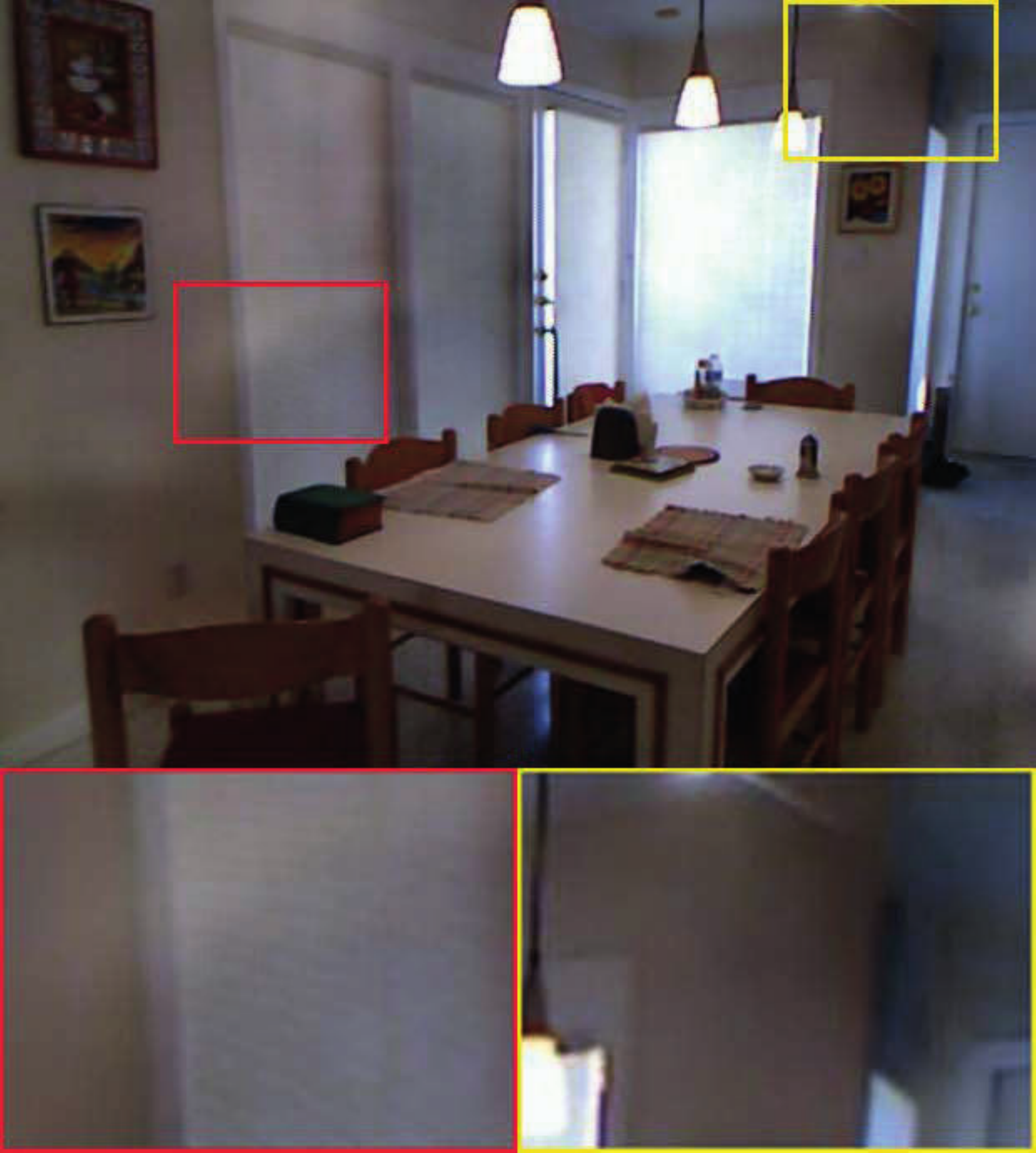}
    \end{minipage}}
    {\begin{minipage}[t]{0.24\textwidth}
    \centering\includegraphics[width=1\textwidth]{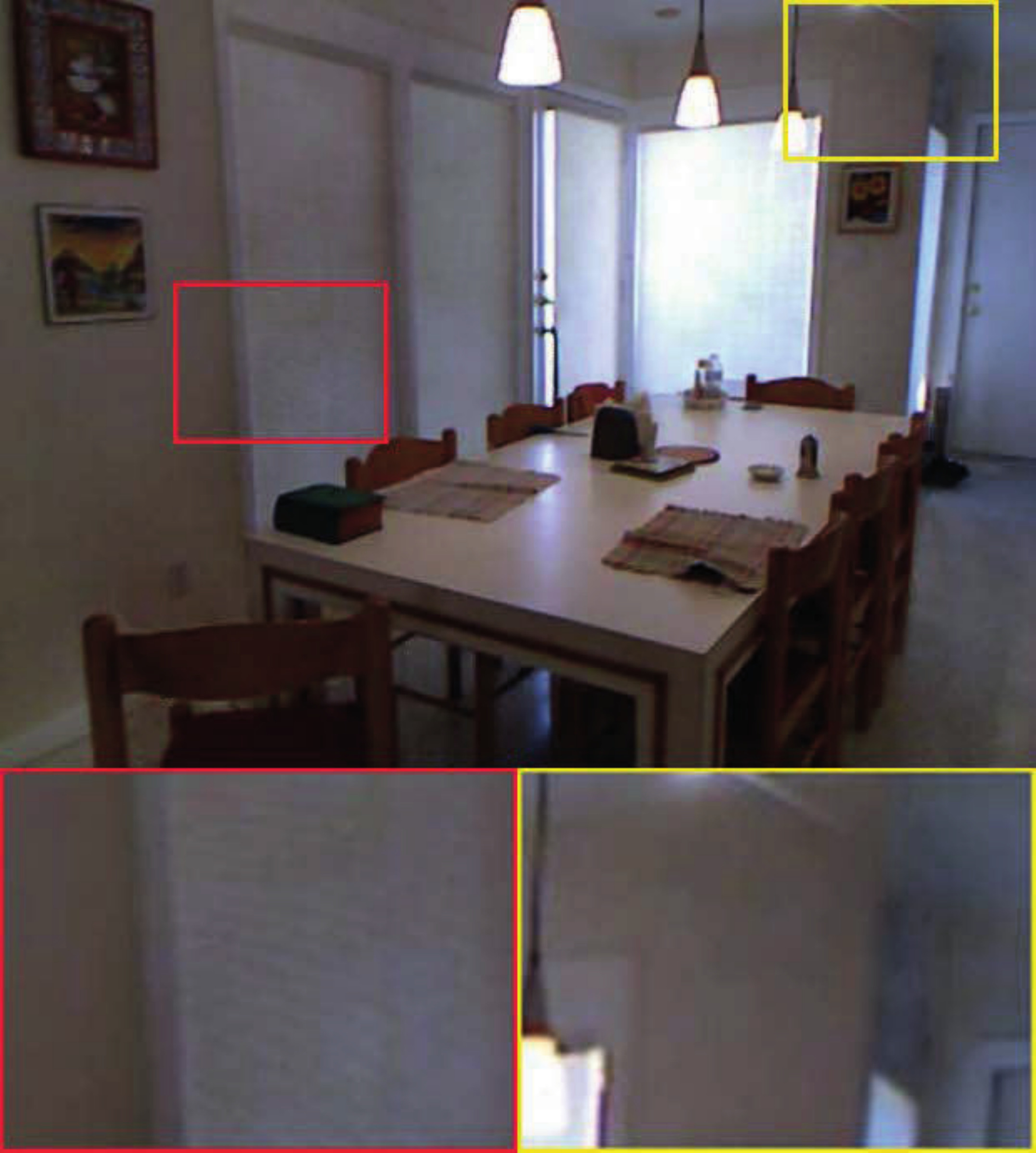}
    \end{minipage}}
    {\begin{minipage}[t]{0.24\textwidth}
    \centering\includegraphics[width=1\textwidth]{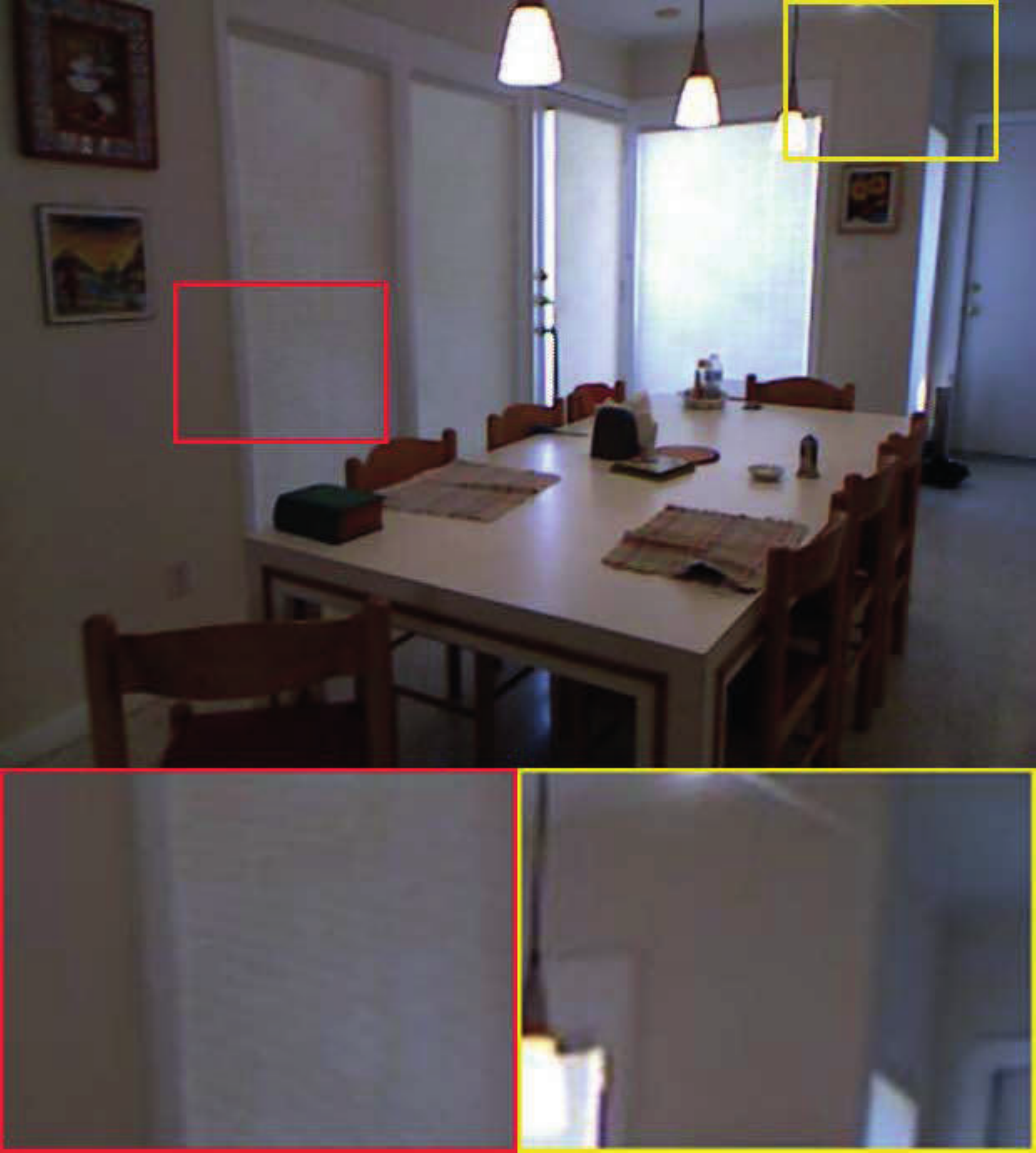}
    \end{minipage}} \\
    \vspace{-0.15cm}
    \subfigure[Hazy input]{\begin{minipage}[t]{0.24\textwidth}
    \centering\includegraphics[width=1\textwidth]{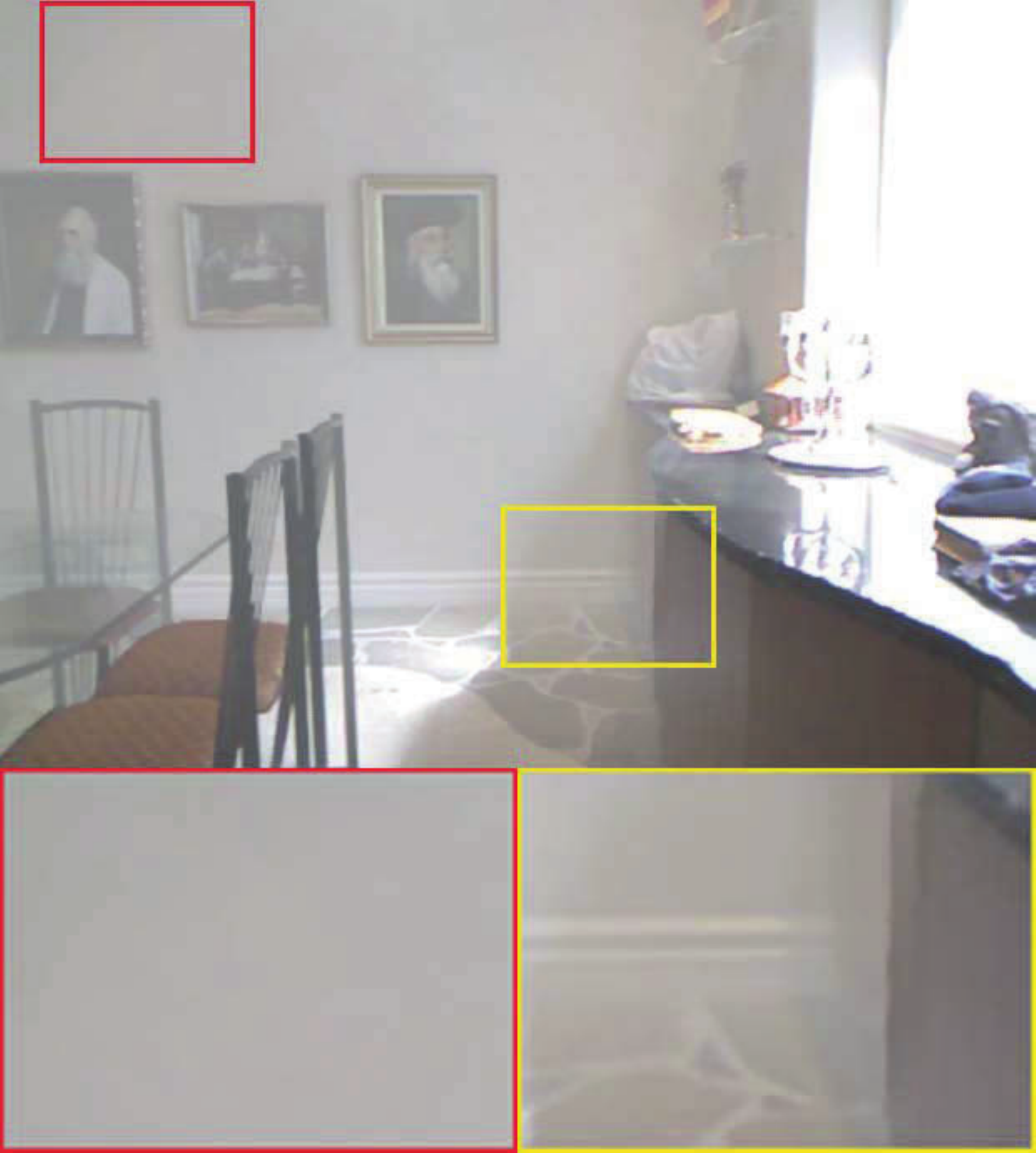}\end{minipage}}
    \subfigure[Grid-DehazeNet~\cite{liu_GridDehazeNet_ICCV_2019}]{\begin{minipage}[t]{0.24\textwidth}
    \centering\includegraphics[width=1\textwidth]{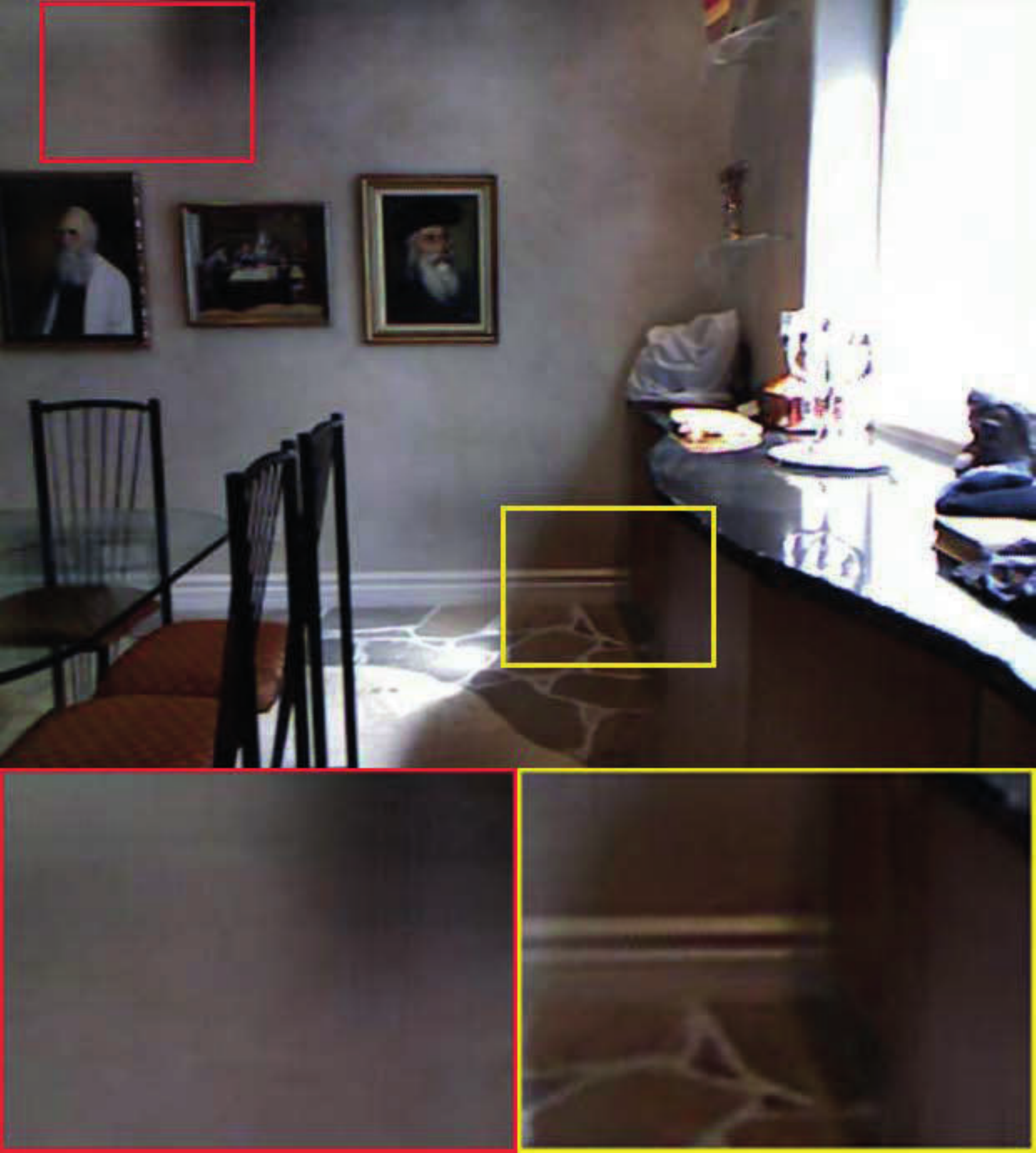}\end{minipage}}
    \subfigure[Ours]{\begin{minipage}[t]{0.24\textwidth}
    \centering\includegraphics[width=1\textwidth]{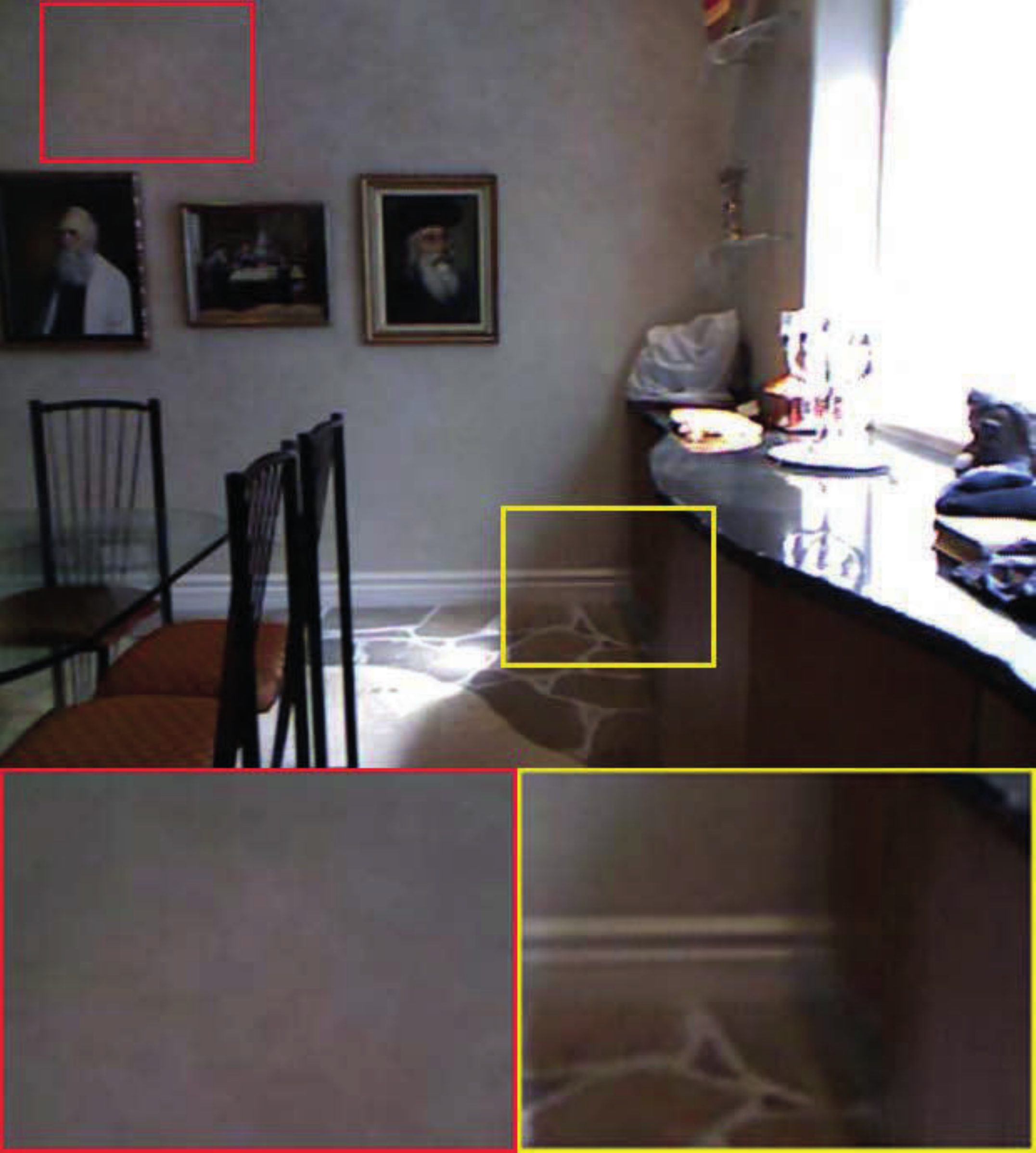}\end{minipage}}
    \subfigure[Ground-Truth]{\begin{minipage}[t]{0.24\textwidth}
    \centering\includegraphics[width=1\textwidth]{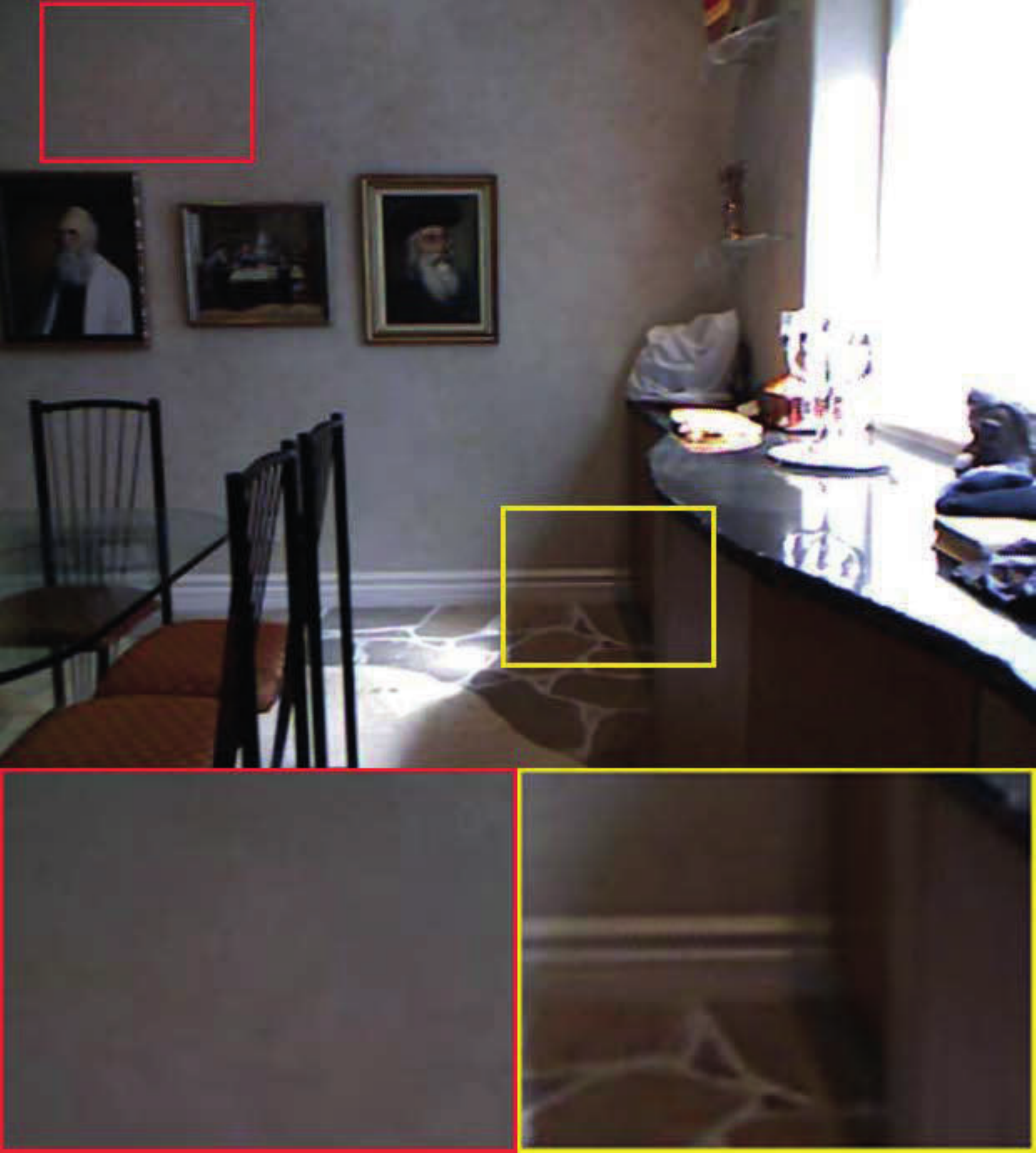}\end{minipage}} \\
    \centering
    \caption{\textbf{Visual comparisons on the SOTS indoor dataset for image dehazing.} 
    Column (a) is the input hazy images. (b), (c) is from Grid-DehazeNet~\cite{liu_GridDehazeNet_ICCV_2019} and our method, respectively. (d) is the ground-truth image. Best Viewed on Screen.}
    \label{fig:fig_rain}
\end{figure*}

\begin{figure*}[htbp]
    \centering
    {\begin{minipage}[t]{0.24\textwidth}
    \centering\includegraphics[width=1\textwidth]{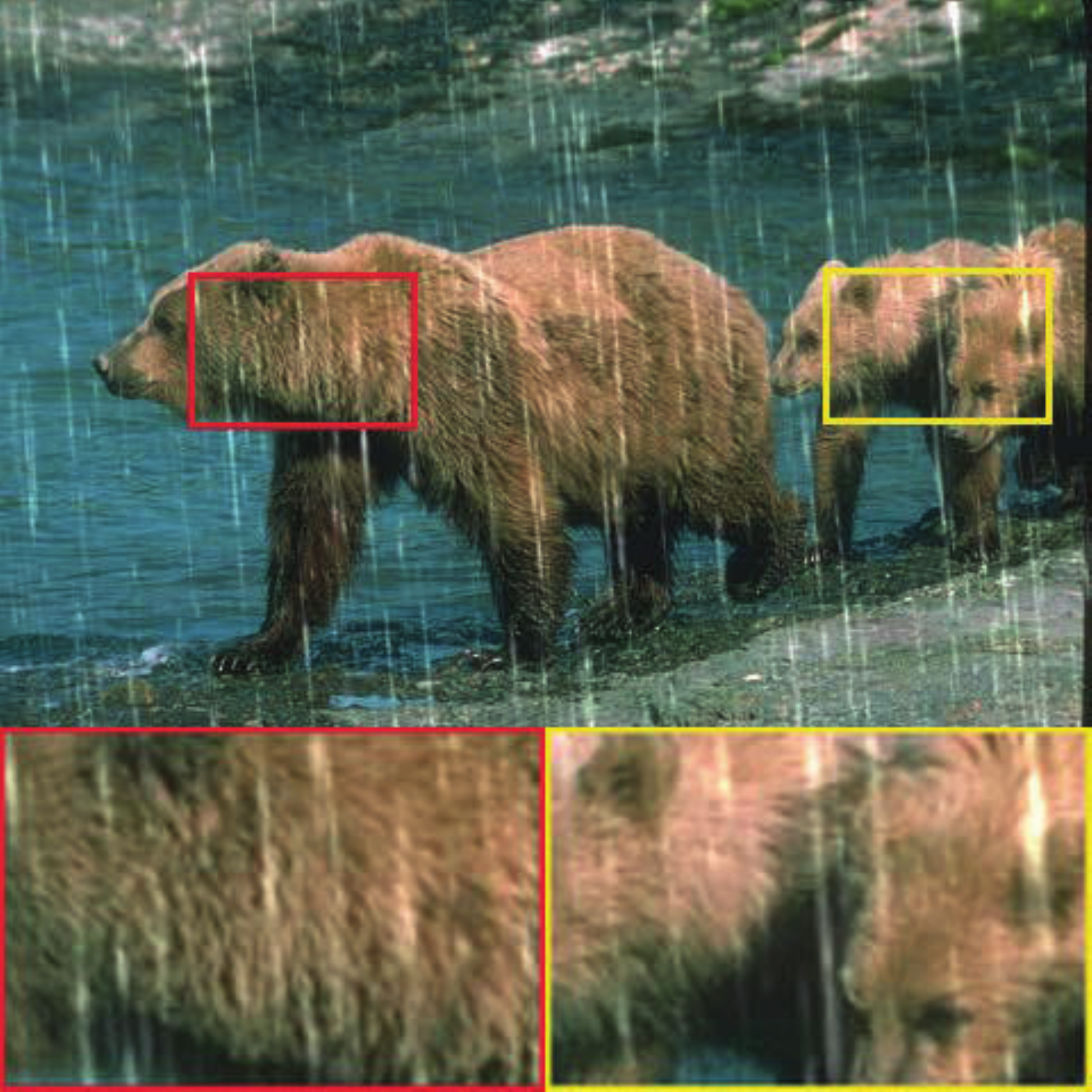}
    \end{minipage}}
    {\begin{minipage}[t]{0.24\textwidth}
    \centering\includegraphics[width=1\textwidth]{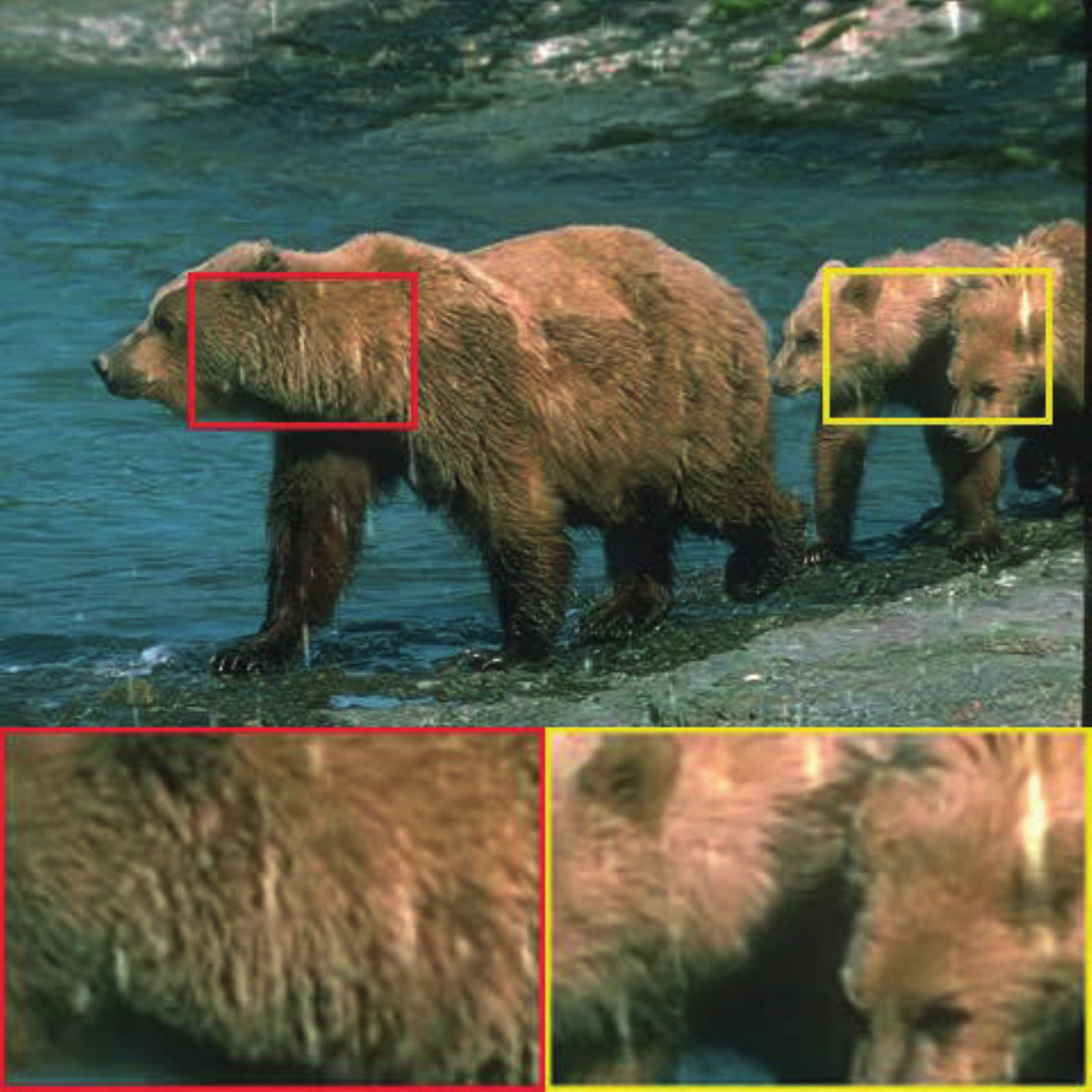}
    \end{minipage}}
    {\begin{minipage}[t]{0.24\textwidth}
    \centering\includegraphics[width=1\textwidth]{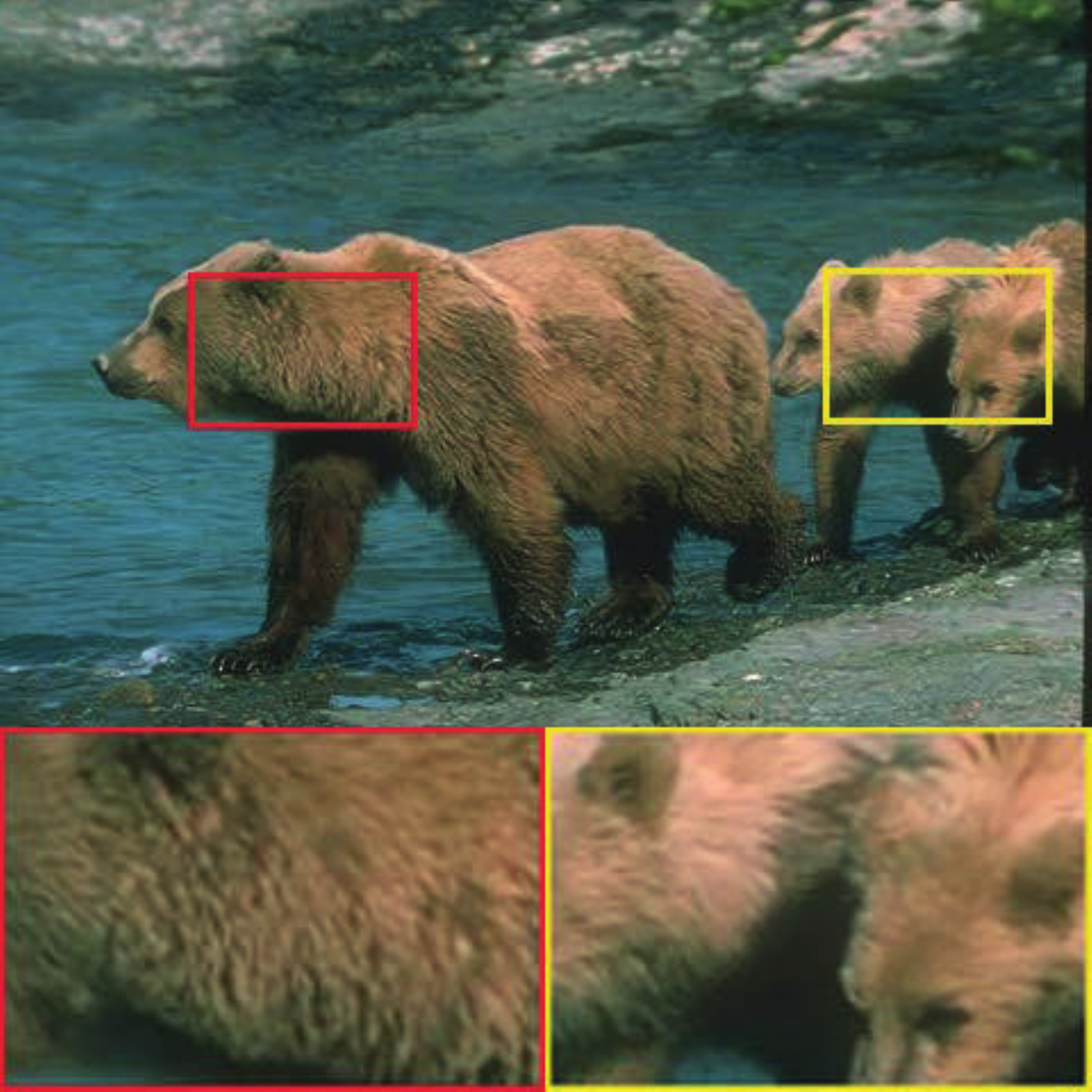}
    \end{minipage}}
    {\begin{minipage}[t]{0.24\textwidth}
    \centering\includegraphics[width=1\textwidth]{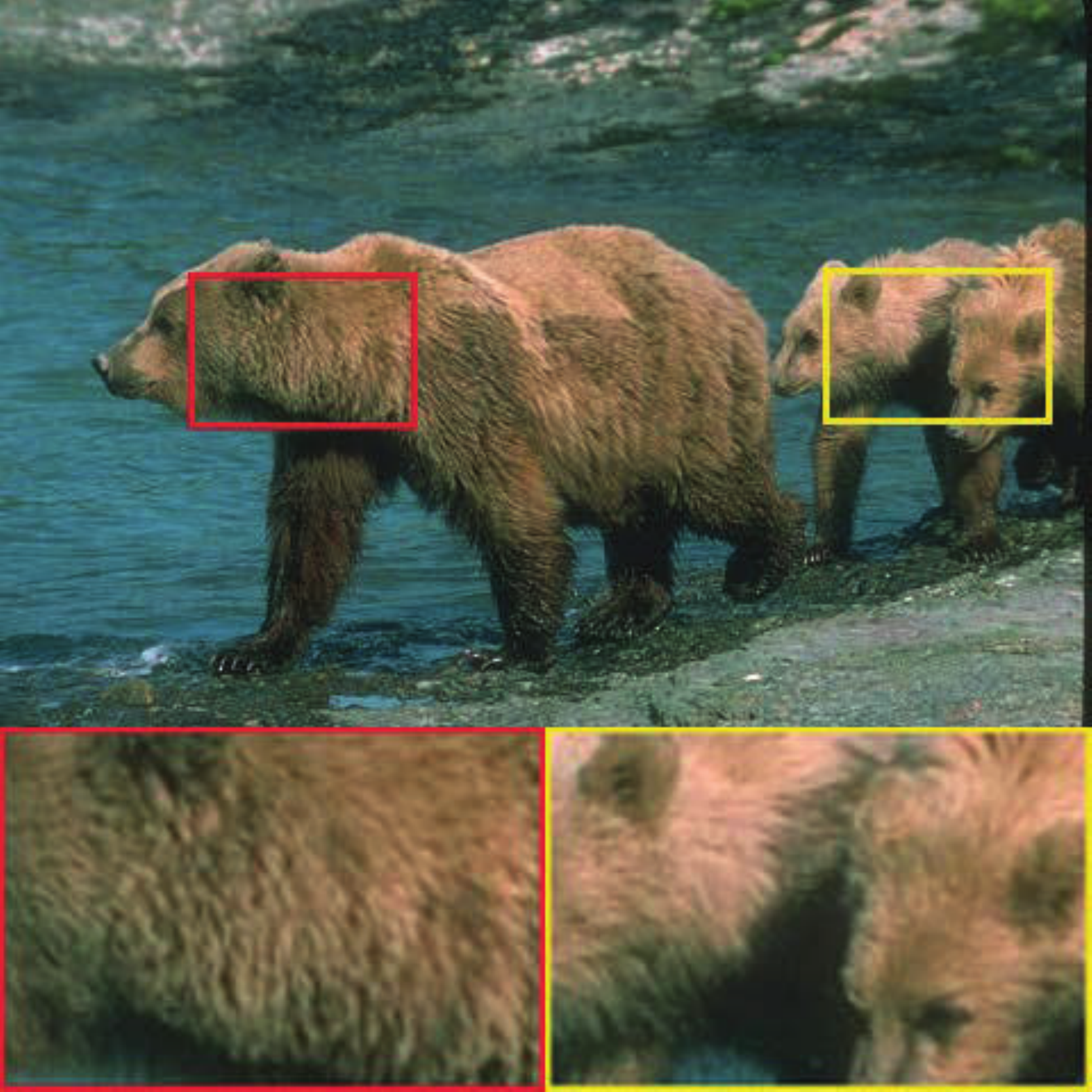}
    \end{minipage}} \\
    \vspace{-0.15cm}
    \subfigure[Rainy input]{\begin{minipage}[t]{0.24\textwidth}
    \centering\includegraphics[width=1\textwidth]{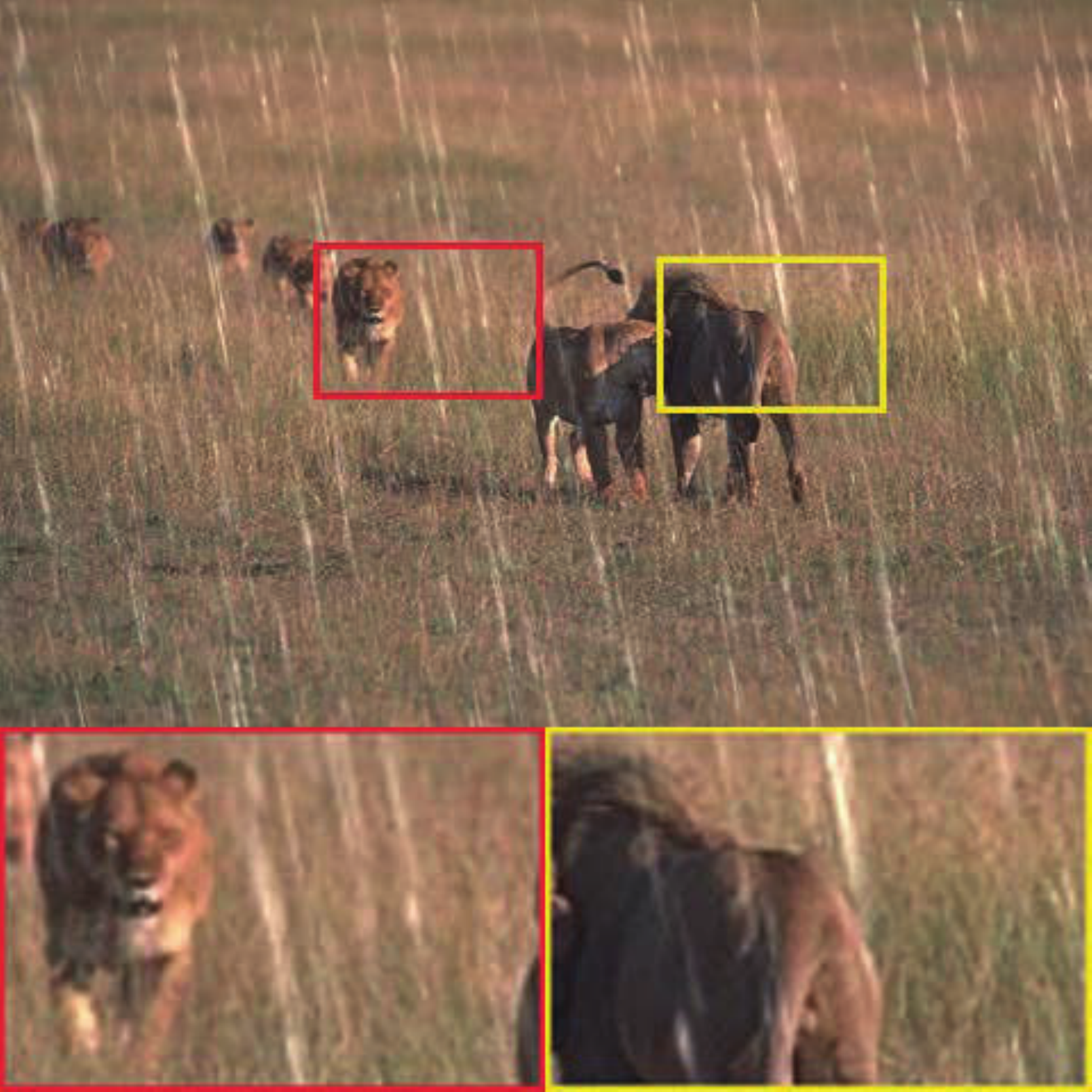}\end{minipage}}
    \subfigure[PreNet~\cite{ren_PreNet_CVPR_2019}]{\begin{minipage}[t]{0.24\textwidth}
    \centering\includegraphics[width=1\textwidth]{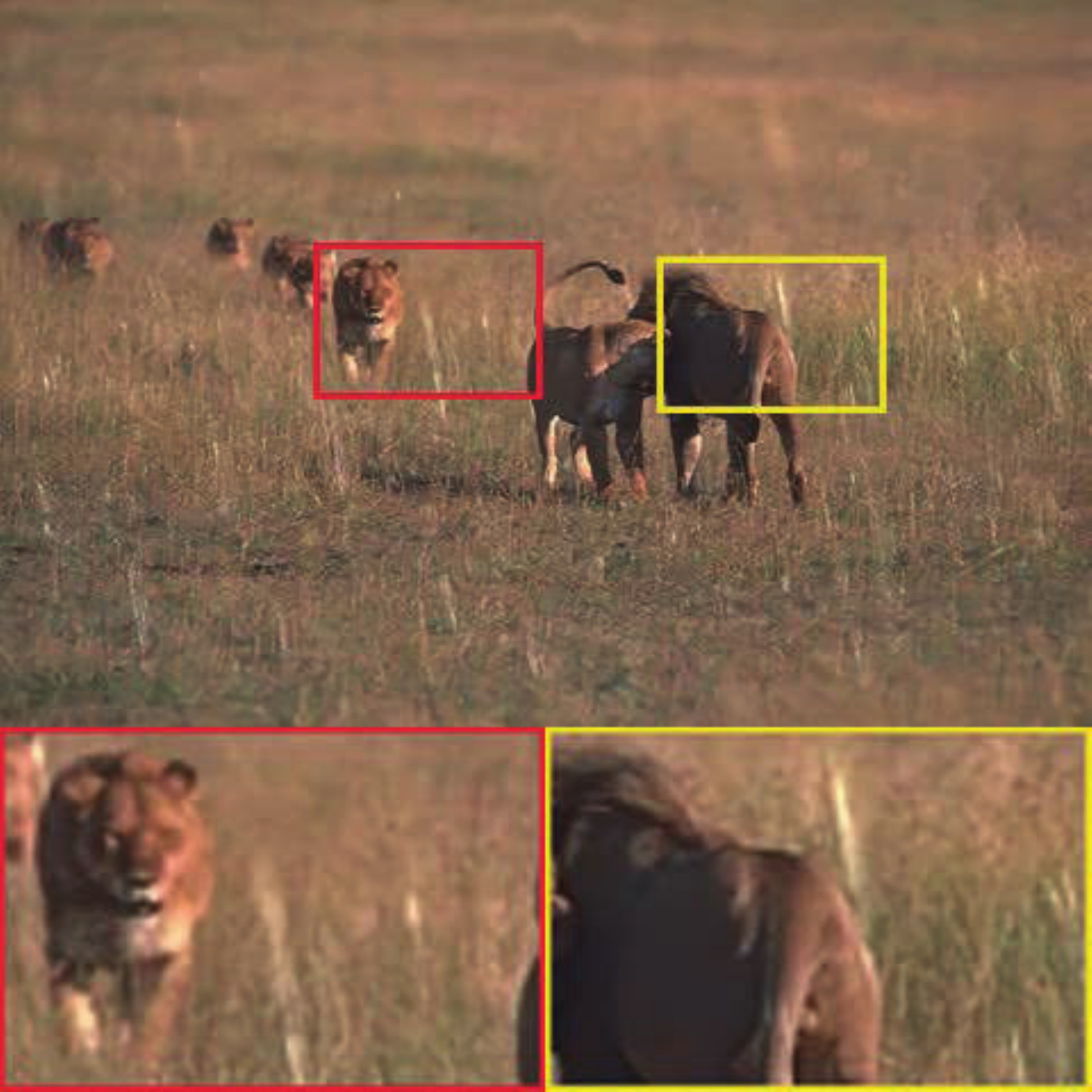}\end{minipage}}
    \subfigure[Ours]{\begin{minipage}[t]{0.24\textwidth}
    \centering\includegraphics[width=1\textwidth]{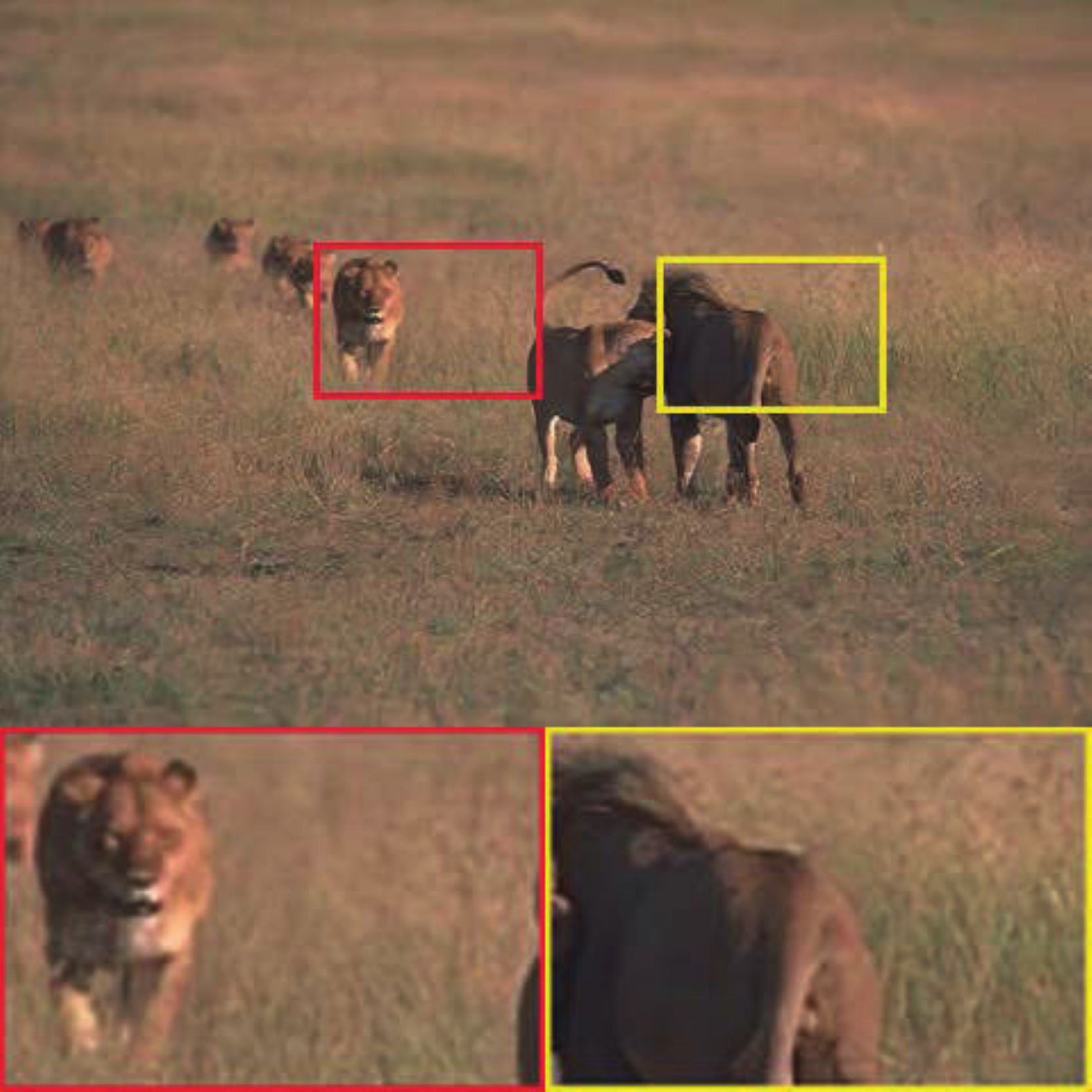}\end{minipage}}
    \subfigure[Ground-Truth]{\begin{minipage}[t]{0.24\textwidth}
    \centering\includegraphics[width=1\textwidth]{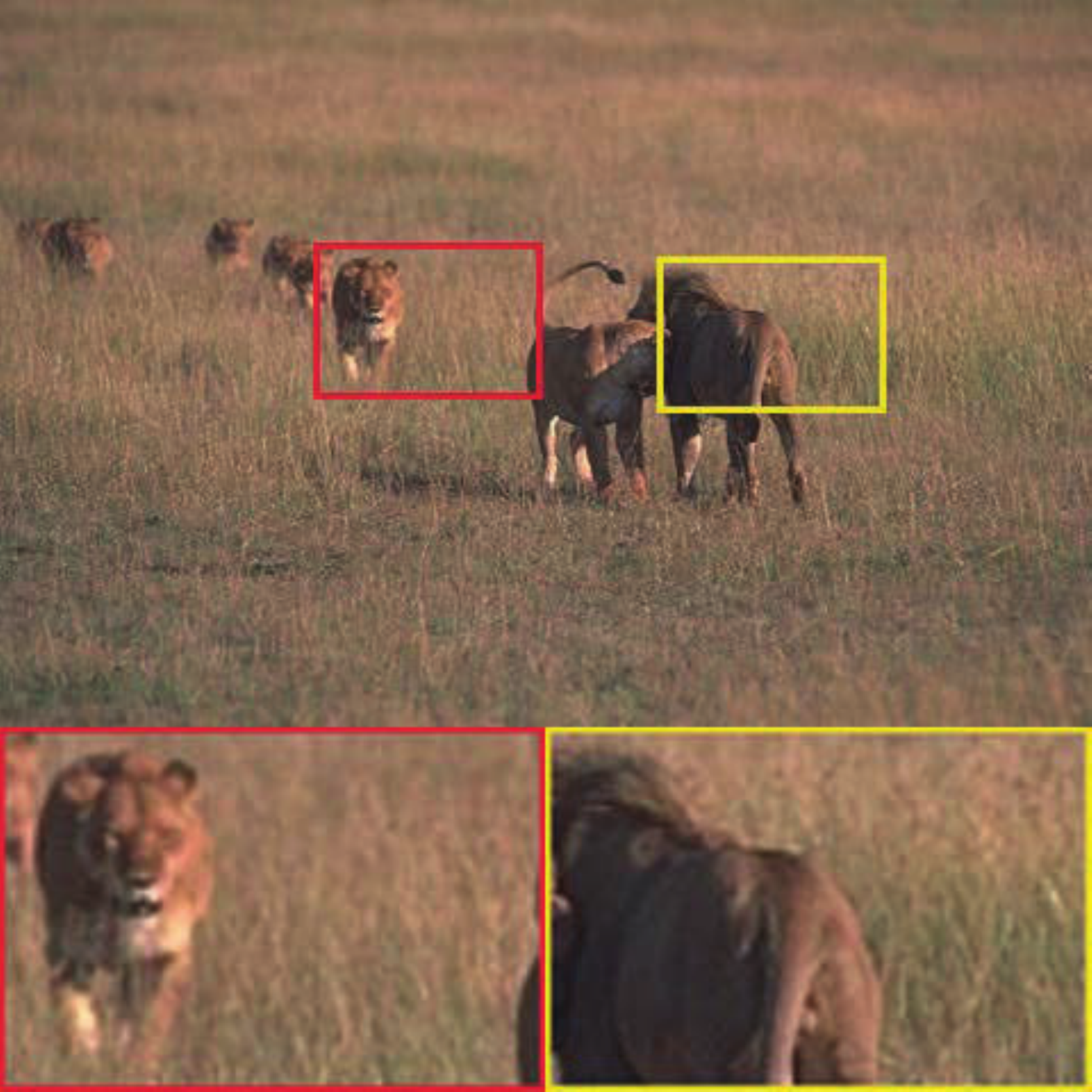}\end{minipage}} \\
    \centering
    \caption{\textbf{Visual comparisons on the R100H dataset for image deraining.} 
    Column (a) is the input rainy images. (b), (c) is from PreNet~\cite{ren_PreNet_CVPR_2019} and our method, respectively. (d) is the ground-truth image. Best Viewed on Screen.}
    \label{fig:fig_haze}
\end{figure*}

\begin{table}[t]
    \caption{\Fix{\textbf{Quantitative analysis on the synthesized GoPro dataset with different levels of Gaussian noise}. The different numbers in the first row mean the magnitude of the Gaussian noise added on the input blurry images.}}
    \label{tab:MultiLevelNoise}
    \centering
    \resizebox{0.48 \textwidth}{!}{
    \begin{tabular}{cccccc}
        \toprule
         Methods $\backslash$ PSNR(dB) &Original & $\sigma = 5$ & $\sigma = 10$ & $\sigma = 15$ & $\sigma = 20$ \cr
        \midrule
        \textbf{Ours}(w/o Idem.) & 31.80 & 30.76 & 29.58 & 28.16 & 26.05\cr
        \textbf{Ours}(w/ Idem.)  & 31.92 & 30.89 & 29.66 & 28.24 & 26.55\cr
        \toprule
    \end{tabular}
    }
\end{table}

\subsection{\Fix{Noise Adaptation of Our Method}}
\Fix{To verify the robustness of our model to noise, we analyze the performance of our pre-trained model by adding different levels of Gaussian noise to the blurry input of the GoPro dataset~\cite{nah_DeepDeblur_CVPR_2017}. The noise levels include 5, 10, 15 and 20, where each number represents the standard deviation of the normal distribution noise within the pixel range of [0, 255]. Then we evaluate the deblurring performance of our pre-trained model with or without our proposed idempotent constraint to analyze the noise adaptability. The deblurring performance under different noise levels is reported in Table~\ref{tab:MultiLevelNoise}. Experimental results show the stability of our model for noisy inputs. 
This observation illustrates that our proposed idempotent constraint can make the model more robust to noise than the model without the constraint.} \par

\subsection{Extension to Image Dehazing}
Our proposed deep idempotent framework is rather general and not limited to image deblurring. We perform image dehazing with our proposed model and the idempotent constraint to investigate the versatility and scalability of different image restoration tasks.
Following the same training pipeline of GridDehazeNet~\cite{liu_GridDehazeNet_ICCV_2019}, our model is trained on Indoor Training Set (ITS) and tested on the Synthetic Objective Testing Set (SOTS) Indoor Subset in RESIDE dataset~\cite{li_RESIDE_TIP_2019}. Table~\ref{tab:DehazeResults} shows that with the idempotent constraint, our model achieves state-of-the-art performance on the SOTS indoor dataset. These results show that our idempotent constraint can boost the dehazing performance by a large margin (1.86dB). 
Fig.~\ref{fig:fig_haze} shows the visual comparison of dehazing results on SOTS indoor dataset. Compared with Grid-DehazeNet~\cite{liu_GridDehazeNet_ICCV_2019}, our model has a clear advantage in visual effects. For instance, previous methods may not fully dehaze some image regions and produce black artifacts in the zoom-in areas.
\begin{table}[ht]
    \caption{\textbf{Quantitative results of applying our idempotent framework on the SOTS indoor dataset for Image Dehazing}. Best and second-best scores are \textbf{highlighted} and \underline{underlined}.}
    \label{tab:DehazeResults}
    \centering
    \resizebox{0.48\textwidth}{!}{
        \Large
        \begin{threeparttable}
        \begin{tabular}{cccccccc}
            \toprule
             Methods  & \cite{cai_dehazenet_TIP_2016} & \cite{chen_GCANet_WACV_2019} & \cite{liu_GridDehazeNet_ICCV_2019} & \cite{Qin_FFANet_AAAI_2020} & \cite{deng_HardGAN_ECCV_2020} & \textbf{Ours}(w/o) & \textbf{Ours}(w/)  \cr
            \midrule
            PSNR~(dB)  & 19.82   & 30.23   & 32.16   & 36.39   & \underline{36.56} & 34.78 & \textbf{36.64} \cr
            SSIM      & .8209   & .9800   & .9836   & .9556   & \textbf{.9905}    & .9823 & \underline{.9851}\cr
            \toprule
        \end{tabular}
        \end{threeparttable}
    }
\end{table}

\begin{table}
\begin{center}
\caption{\textbf{Quantitative results of applying our idempotent framework for image deraining.} Best and second-best scores are \textbf{highlighted} and \underline{underlined}.}
\label{table:deraining}

\resizebox{0.5\textwidth}{!}{
\Large
\begin{tabular}{lcccccccccc}
\toprule
  & \multicolumn{2}{c}{Test100~\cite{zhang_DerainDataset1_TCSVT_2019}}&\multicolumn{2}{c}{Rain100H~\cite{fu_DerainDataset2_CVPR_2017}}&\multicolumn{2}{c}{Rain100L~\cite{fu_DerainDataset2_CVPR_2017}}\\
                                                Methods & PSNR~(dB) & SSIM & PSNR~(dB) & SSIM & PSNR~(dB) & SSIM & \\
    \midrule
    DerainNet~\cite{fu_DerainNet_TIP_2017}      &22.77 &0.810 &14.92 &0.592 &27.03 &0.884 \\
    SEMI~\cite{wei_SemiDerain_CVPR_2019}        &22.35 &0.788 &16.56 &0.486 &25.03 &0.842 \\
    DIDMDN~\cite{zhang_DensityDerain_CVPR_2018} &22.56 &0.818 &17.35 &0.524 &25.23 &0.741 \\
    UMRL~\cite{yasarla_UncerDerain_CVPR_2019}   &24.41 &0.829 &26.01 &0.832 &29.18 &0.923 \\
    RESCAN~\cite{li_RecurrentDerain_ECCV_2018}  &25.00 &0.835 &26.36 &0.786 &29.80 &0.881 \\
    PreNet~\cite{ren_PreNet_CVPR_2019}          &24.81 &0.851 &26.77 &0.858 &32.44 &0.950 \\
    MSPFN~\cite{jiang_MSPFN_CVPR_2020}          &27.50 &0.876 &28.66 &0.860 &32.40 &0.933 \\
    \midrule
    \textbf{Ours}(w/o Idem.)  &\underline{28.94} &\underline{0.889} &\underline{29.97} &\underline{0.881} &\underline{34.14} &\underline{0.942}  \\
    \textbf{Ours}(w/ Idem.)   &\textbf{29.00} &\textbf{0.892} &\textbf{30.10} &\textbf{0.882} &\textbf{34.68} & \textbf{0.954}  \\
    \toprule
\end{tabular}}
\end{center}
\end{table}

\subsection{Extension to Image Deraining}
We also extend our proposed deep idempotent framework to image deraining. Following the same training pipeline of MSPFN~\cite{jiang_MSPFN_CVPR_2020}, we train our model on about 13,700 clean/rain image pairs collected from~\cite{zhang_DerainDataset1_TCSVT_2019, fu_DerainDataset2_CVPR_2017}. We evaluate the performance on the testing datasets, including Test100~\cite{zhang_DerainDataset1_TCSVT_2019}, Rain100H~\cite{fu_DerainDataset2_CVPR_2017} and Rain100L~\cite{fu_DerainDataset2_CVPR_2017}. The results in Table~\ref{table:deraining} show that our designed structure and idempotent constraint boost the deraining performance, respectively. Fig.~\ref{fig:fig_rain} shows the visual comparison of deraining results on R100H dataset. Compared with PreNet~\cite{ren_PreNet_CVPR_2019}, our model has better visual effects.

\section{Conclusion}
In this paper, we have presented a novel deep idempotent network for efficient single image blind deblurring. First, we introduced the idempotent constraint to the deep deblurring network, which improves the non-uniform deblurring performance and achieves stable results \wrt re-deblurring multiple times. Second, we designed a simple yet efficient deblurring network through progressive residual deblurring with recurrent structure. Our model achieves state-of-the-art performance with smaller parameters and faster inference time than the state-of-the-art methods. The introduced idempotent constraint, as a regularization term, plays an important role in reducing the solution search space and thus leads to a more stable solution regardless of the times of re-deblurring. By adopting our idempotent constraint in model training, our model shows great generalization performance on real-world and synthetic datasets. Our framework is not limited to image deblurring, and we have verified the superiority of our framework in image dehazing and image deraining. In the future, we will further extend it to other image restoration tasks such as image denoising~\cite{zhang_Denoising_TIP_2017}. \Fix{It is also promising to explore a tailored multiple-image input idempotent network for video~\cite{zhang_recursive_video_deblur_TCSVT_2021} or light field image deblurring~\cite{srinivasan_lightmotiondeblur_cvpr_2017}.}

\section*{Acknowledgments}
This research was supported in part by National Natural Science Foundation of China (61871325, 61901387, 62001394), National Key Research and Development Program of China (2018AAA0102803), Zhejiang Lab (NO.2021MC0AB05) and Australian Research Council (DP220100800).

\ifCLASSOPTIONcaptionsoff
  \newpage
\fi

\begin{IEEEbiographynophoto}{Yuxin Mao}
is currently a PhD student with School of Electronics and Information, Northwestern Polytechnical University, Xi'an, China. He received his Bachelor of Engineering degree from Southwest Jiaotong University in 2020. He won the best Paper Award Nominee at ICIUS 2019.
\end{IEEEbiographynophoto}

\begin{IEEEbiographynophoto}{Zhexiong Wan}
is currently a PhD student with School of Electronics and Information, Northwestern Polytechnical University, Xi'an, China. He received his Bachelor of Engineering degree from Northwestern Polytechnic University in 2019. 
\end{IEEEbiographynophoto}


\begin{IEEEbiographynophoto}{Yuchao Dai} is currently a Professor with School of Electronics and Information at the Northwestern Polytechnical University (NPU). He received the B.E. degree, M.E degree and Ph.D. degree all in signal and information processing from Northwestern Polytechnical University, Xi'an, China, in 2005, 2008 and 2012, respectively. He was an ARC DECRA Fellow with the Research School of Engineering at the Australian National University, Canberra, Australia. His research interests include structure from motion, multi-view geometry, low-level computer vision, deep learning, compressive sensing and optimization. He won the Best Paper Award in IEEE CVPR 2012, the DSTO Best Fundamental Contribution to Image Processing Paper Prize at DICTA 2014, the Best Algorithm Prize in NRSFM Challenge at CVPR 2017, the Best Student Paper Prize at DICTA 2017, the Best Deep/Machine Learning Paper Prize at APSIPA ASC 2017, the Best Paper Award Nominee at IEEE CVPR 2020. He served as Area Chair in CVPR, ICCV, ACM MM, ACCV, WACV and etc.  \end{IEEEbiographynophoto}

\begin{IEEEbiographynophoto}{Xin Yu}
received his B.S. degree in Electronic Engineering from University of Electronic Science and Technology of China, Chengdu, China, in 2009, and received his Ph.D. degree in the Department of Electronic Engineering, Tsinghua University, Beijing, China, in 2015. He also received a Ph.D. degree in the College of Engineering and Computer Science, Australian National University, Canberra, Australia, in 2019. He is currently a lecturer in University of Technology Sydney. His interests include computer vision and image processing.
\end{IEEEbiographynophoto}




\end{document}